\documentclass[10pt,twocolumn,letterpaper]{article}

\usepackage{cvpr}      
\definecolor{cvprblue}{rgb}{0.21,0.49,0.74}
\usepackage[pagebackref,breaklinks,colorlinks,allcolors=cvprblue]{hyperref}

\def\paperID{} 
\def\confName{CVPR}
\def\confYear{2026}
\usepackage{booktabs}
\usepackage{colortbl}
\usepackage{xcolor}
\usepackage{multirow}
\usepackage{graphicx}
\usepackage[accsupp]{axessibility}
\title{CaptionQA: Is Your Caption as Useful as the Image Itself?}

\author{
Shijia Yang$^{5}$\thanks{Equal Contribution} \quad Yunong Liu$^{2*}$ \quad Bohan Zhai$^{3*}$ \quad Ximeng Sun$^1$ \\
Zicheng Liu$^1$ \quad Emad Barsoum$^1$ \quad Manling Li${^4}$ \quad Chenfeng Xu${^5}$ \\
\textsc{
$^1$ Advanced Micro Devices, Inc. \space $^2$Stanford University}\\
\textsc{$^3$Independent Researcher \space $^4$Northwestern University \space $^5$UT Austin} \\ 
Project page: \url{https://captionqa.github.io/website/}
}

\begin{document}
\maketitle
\begin{abstract}
Image captions serve as efficient surrogates for visual content in multimodal systems such as retrieval, recommendation, multi-step agentic inference pipelines. Yet current evaluation practices miss a fundamental question: \emph{Can captions stand-in for images in real downstream tasks?} We propose a utility-based benchmark, \textbf{CaptionQA}, to evaluate model-generated captions, where caption quality is measured by how well it supports downstream tasks. CaptionQA is an extensible domain-dependent benchmark covering 4 domains: Natural, Document, E-commerce, and Embodied AI, each with fine-grained taxonomies (25 top-level and 69 subcategories) that identify useful information for domain-specific tasks. CaptionQA builds 33,027 densely annotated multiple-choice questions (50.3 per image on average) that explicitly require visual information to answer, providing a comprehensive probe of caption utility. In our evaluation protocol, an LLM answers these questions using captions alone, directly measuring whether captions preserve image-level utility and are utilizable by a downstream LLM. Evaluating state-of-the-art MLLMs reveals substantial gaps between the image and its caption utility. Notably, models nearly identical on traditional image-QA benchmarks lower by up to 32\% in caption utility. We release CaptionQA along with an open-source pipeline for extension to new domains.
\end{abstract}

\begin{figure*}[h!]
\centering
\includegraphics[width=0.95\textwidth]{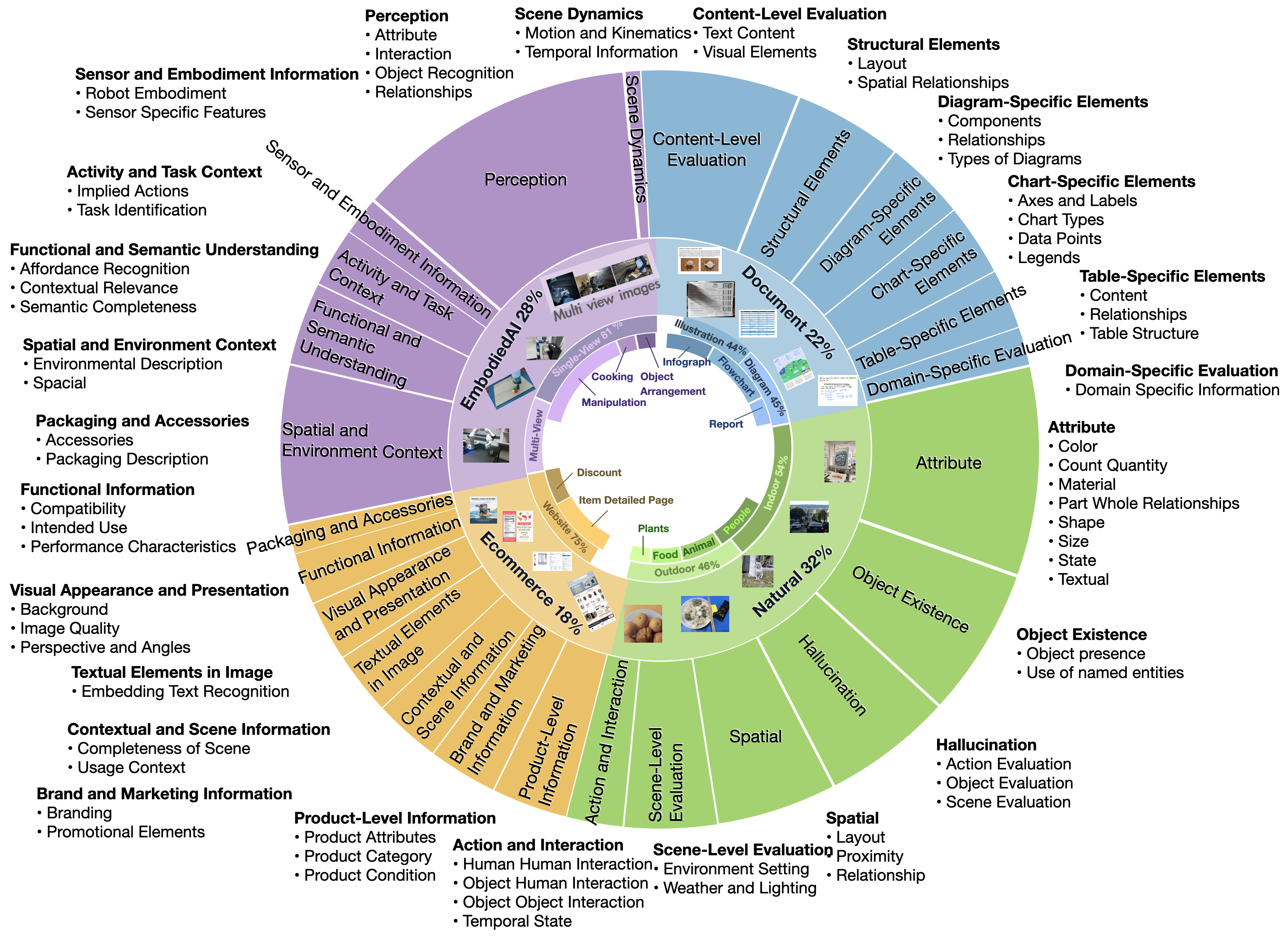}
\vspace{-0.3cm}
\caption{\textbf{CaptionQA taxonomies across four domains}, the visual information that captions must carry to be useful for downstream tasks.
The Natural domain (6 top-level, 22 subcategories) emphasizes object properties, spatial relations, and hallucination; the Document domain (6, 15) targets layout, content, and document-specific structure; the E-commerce domain (7, 16) focuses on product attributes and presentation; and the Embodied AI domain (6, 16) captures perception, spatial understanding, and task-relevant cues for robotics.}
\label{fig:taxonomy_overview}
\vspace{-0.5cm}
\end{figure*}

\vspace{-0.5cm}
\section{Introduction}
\label{sec:intro}
Image captioning is a fundamental task in vision-language research, as it directly probes what an MLLM perceives and how it describes. Beyond scientific value, captions increasingly function as stand-ins for images in real-world systems, offering an efficient representation: they turn unstructured visual inputs into searchable, analyzable signals for document automation~\cite{mindsdb2024unstructured,fireworks2024multimodal,mistral2025ocr,snowflake2025multimodalpdf} and enterprise-scale retrieval~\cite{kolouju2025good4cir,huynh2025collm}, serve as generalized item features for e-commerce recommendation~\cite{deldjoo2021study,wang2025leveraging}, and act as privacy-preserving semantic indices over natural multimodal datasets. In agentic and embodied AI pipelines, captions further function as language-based state representations and long-term memory structures, bridging raw perceptual observations with downstream reasoning and planning~\cite{yang2025magma,yao2025survey}.

Despite their widespread practical use, captions remain under-evaluated in academic research. Current MLLM evaluation focuses mainly on QA-on-image benchmarks, where models answer general recognition or reasoning questions given the image. These benchmarks, however, primarily reflect passive visual question answering rather than directly revealing a model’s captioning ability~\cite{zhai2023halle,liu2025good,ye2025painting}. Dedicated captioning benchmarks exist, but they are either outdated or inadequate~\cite{Papineni:2002,vedantam2015cider}, while newer VLM/LLM-based evaluations tend to be complex, non-deterministic, limited to natural images, and narrowly focused on object names, attributes, or simple relationships.
As a result, the prevailing benchmarks fail to encourage models to behave in line with real-world usage, creating a gap between how MLLMs are assessed and how captions are actually used.

This gap motivates a utility-based view of captioning: \emph{what matters is not how much detail a caption includes, but whether it captures the details that matter for the intended use case.} Under this view, we evaluate captions by how effectively they support downstream tasks--that is, how closely they match the image in \emph{utility}. To this end, we introduce \textbf{CaptionQA}, a benchmark that actively tests caption utility by measuring how well a caption can stand-in for its image.

Since "utility" is domain-dependent, CaptionQA currently spans four major domains--\textbf{Natural}, \textbf{Document}, \textbf{E-commerce}, and \textbf{Embodied AI}--and can be readily extended via our open-source code. For each domain, we work with human experts to construct a \emph{fine-grained taxonomy that identifies useful information for downstream tasks.}. These domain-specific taxonomies specify what captions are expected to support, as shown in Figure~\ref{fig:taxonomy_overview}. Using these taxonomies, CaptionQA constructs dense multiple-choice QA pairs directly from real-world images. Each question is explicitly tied to one or more taxonomy nodes, so answering it requires the specific \emph{image-level utility} that domain-specific downstream tasks would need. 

To test whether captions retain this utility, we prompt an MLLM to generate a caption for each image. A separate, text-only LLM is then asked to answer the corresponding multiple-choice questions using only the caption, without seeing the image (QA-on-caption). High CaptionQA scores indicate that captions approach image-level utility for downstream tasks, thereby answering: \emph{is your caption as useful as the image itself?} Evaluating state-of-the-art MLLMs on CaptionQA, we find that \emph{captions systematically lag behind images in utility}: even strong proprietary models lose 9.2-16.4\% when moving from QA-on-image to QA-on-caption. For intuition, this means that a downstream application that correctly perceives 90\% of information with images would drop to roughly 74\% to 81\% when given the strongest captions, turning about one in every six previously correct decisions into an error. Open-source models fare even worse, with gaps of 11-32.4\%; in the Embodied AI domain, the gap exceeds 40\%, and the Natural and Document domains show similarly large drops, which means that almost half of the useful signal at the image-level is lost when only the caption is kept, and downstream tasks in these domains can become highly inaccurate. Notably, models that are close on standard image-QA benchmarks (e.g., Claude Sonnet 4.5 vs.\ LLaVA-OneVision-7B by just $\sim$1\%) can diverge by up to 32\% on our CaptionQA, showing that \emph{small differences in traditional multimodal evaluation can hide large discrepancies in caption utility}.

In summary, our contributions are as follows:
\begin{itemize}[leftmargin=*]
    \item We introduce \textbf{CaptionQA}, a new utility-oriented caption evaluation benchmark that covers 4 real-world domains, 33{,}027 multiple-choice questions, and an average of 50.3 questions per image, together with domain taxonomies comprising 25 top-level categories and 69 subcategories.
    
    \item We propose a simple and light cost QA-based protocol for caption evaluation, grounded in a utility-based view: a deterministic downstream LLM answers questions using only the caption, allowing us to directly measure how well captions stand-in for images in practical settings.
    
    \item We open-source our carefully designed benchmark construction and quality check pipeline. Extending CaptionQA to a new domain is as quick as writing a new domain-specific taxonomy and running the question generation pipeline, enabling the community to easily extend it to new applications.
\end{itemize}

\section{Related Work}

Evaluating open-ended image captions is a long-standing challenge. The field has evolved through three main paradigms, each presenting limitations that motivate our utility-centric approach.
\vspace{-0.3cm}
\paragraph{Traditional Metrics and Fact-Based Parsers.}
Initial approaches to caption evaluation, built on foundational datasets like MSCOCO~\cite{lin2014microsoft} and Flickr30k~\cite{young2014image}, were adapted from machine translation. These n-gram-based metrics, such as BLEU~\cite{Papineni:2002}, or consensus-based metrics like CIDEr~\cite{vedantam2015cider}, primarily measure textual overlap with reference captions. It is now widely recognized that this approach is insufficient, as these metrics often overlook critical factual errors and can be poorly correlated with human perceptual judgments~\cite{kilickaya2016re,cui2018learningevaluateimagecaptioning}.

To address this factual limitation, a second generation of object-centric evaluation methods emerged. SPICE and CHAIR~\cite{anderson2016spice,rohrbach2019objecthallucinationimagecaptioning} parse captions into semantic tuples, while newer benchmarks~\cite{dong2024benchmarking,zhai2023halle,lu2025benchmarking} attempt to parse a caption to extract "facts" and compute factual precision/recall. While a significant step towards correctness, these methods often rely on complex~\cite{lu2025benchmarking} pipelines, such as LLM-assisted judges~\cite{zhai2023halle,chan2023clairevaluatingimagecaptions} or graph-matching algorithms~\cite{lu2025benchmarking}, which can be non-deterministic and difficult to reproduce. Furthermore, they remain focused on natural images, and their "flat" attribute lists are not readily adapted to the structured, high-value information in specialized domains.
\vspace{-0.3cm}
\paragraph{Detailed Captioning and the VLM-as-Judge Paradigm.}
A recent trend has shifted the focus to "detailed" captioning~\cite{ye2025painting}. To score these long and complex captions, a new "VLM-as-a-Judge" paradigm has become popular, exemplified by benchmarks like CapArena~\cite{cheng2025caparena} and CAPability~\cite{liu2025good}. This approach, while well-correlated with human judgment, introduces new challenges. It outsources evaluation to a non-deterministic "black-box" VLM, making scores dependent on specific prompts and API updates, which hinders reproducibility. Moreover, this trend often conflates utility with verbosity. By optimizing for "thoroughness" (CAPability) or "detail" (DeCapBench~\cite{ye2025painting}), models are encouraged to produce exhaustive, noisy captions, misaligning with the real-world need for efficient, salient information. Thus, our main contribution is not simply switching the evaluation format (QA answer vs.\ statements judge), but proposing a downstream-driven definition of a good caption and a new evaluation objective compare to prior works.
\vspace{-0.3cm}
\paragraph{QA-based Multimodal Evaluation.}
This category itself is split. First, the most common benchmarks reported in SOTA MLLM releases~\cite{antol2015vqa,masry2022chartqa,chen2024we,saikh2022scienceqa,mathew2021docvqa,li2024seed,fu2023mme,zhang2024mme} (e.g., MMBench~\cite{liu2024mmbench}, MMMU~\cite{yue2024mmmu}) evaluate a different capability. These benchmarks test a passive fact-retrieval skill, which is not a proxy for the active compositional skill of generating a comprehensive description. Furthermore, their sparse metric (often 1-2 Qs/image) fails to cover the full scope of an image's content. A line of work~\cite{lian2025describe,lin2024evaluating,yue2025qaeval} aligns with our premise of using QA as an evaluation metric for generative models. This paradigm offers a promising path toward objective, fact-based assessment. However, these methods have not yet provided the deep, structured taxonomy or critical cross-domain coverage necessary to evaluate the specialized caption utility.

In contrast to these approaches, our work introduces a utility-centric paradigm that is deterministic, evaluates the correct active captioning task, and provides deep, taxonomy-driven diagnostics across high-value specialized domains for the first time.

\section{Task and Evaluation Setup}
\label{sec:task}

CaptionQA evaluates image captions by asking whether they can \emph{replace the image} for downstream reasoning.

\subsection{Task Formulation}

Given an image caption $\mathbf{C}$ produced by a target MLLM (the model we want to evaluate) and a domain-specific multiple-choice question $\mathbf{Q}$, a \emph{text-only} LLM produces an answer $\mathbf{A}$ without access to the original image:
\[
    \mathbf{A} = \mathrm{LLM}(\mathbf{Q}, \mathbf{C}).
\]

We first prompt different MLLMs with the caption prompts in Section~\ref{sec:prompt} to obtain captions for every image. 
Then, the QA LLM answers all benchmark questions using only these captions (Section~\ref{sec:llm}). 
Once we have an answer $\mathbf{A}$ for each caption-question pair, we score the captioning MLLM using a multiple-choice scoring scheme with slight modifications (Section~\ref{sec:score}). 
This directly measures how much task-relevant information each caption preserves.

\subsection{Benchmark Scope \& Statistics}

CaptionQA covers four representative domains that reflect diverse real-world applications: natural images, documents, e-commerce product pages, and embodied AI scenes. Each domain is associated with its own aspect schema (taxonomy) that captures the information most relevant to downstream tasks in that domain (e.g., object existence, attributes, layout, actions, affordances). Across all domains, the taxonomy contains 25 top-level categories and 69 subcategories. The final benchmark consists of 33{,}027 multiple-choice questions over 657 images, with an average of 50.3 questions per image. 
Table~\ref{tab:captionqa_stats} summarizes the per-domain statistics.

\begin{table}[t]
\centering
\small
\setlength{\tabcolsep}{4pt}
\caption{Overall statistics of the \textbf{CaptionQA} benchmark across four domains. 
``Top-level'' and ``Sub'' refer to the numbers of taxonomy categories at each level.}
\vspace{-0.3cm}
\label{tab:captionqa_stats}
\begin{tabular}{lrrrr}
\toprule
\textbf{Domain}  & \textbf{\#Questions} & \textbf{\#Top-level} & \textbf{\#Sub} & \textbf{\#Images} \\
\midrule
Natural        & 10{,}445 & 6  & 22 & 158  \\
Document       &  7{,}422 & 6  & 15 & 178 \\
E-commerce     &  5{,}886 & 7  & 16 & 121 \\
Embodied AI    &  9{,}274 & 6  & 16 & 200 \\
\midrule
\textbf{Total} & 33{,}027 & 25 & 69 & 657 \\
\bottomrule
\end{tabular}
\vspace{-.5cm}
\end{table}



\subsection{Caption Prompts}
\label{sec:prompt}

Caption quality is sensitive to the instruction given to the MLLM. 
To study how prompting affects the utility of generated captions, we evaluate each model under four captioning prompts, shared across all domains:

\begin{itemize}[leftmargin=*]
    \item \textbf{Long.} 
    \emph{``Write a very long and detailed caption describing the given image as comprehensively as possible.''}
    
    \item \textbf{Short.} 
    \emph{``Write a very short caption for the given image.''}
    
    \item \textbf{Simple.} 
    \emph{``Describe this image in detail.''}
    
    \item \textbf{Taxonomy-Hinted.}
    We explicitly condition the caption on our domain-specific taxonomy. 
    Concretely, we ask the model to
    \emph{``Describe this image from the following perspectives. 
    Skip any aspect that does not apply.''}
    and then list taxonomy nodes in the form
    \emph{\texttt{Top-category -> Subcategory}}, e.g.,
    \emph{\texttt{Object Existence -> Object presence}}, 
    \emph{\texttt{Attribute -> Color}}, etc.
\end{itemize}

We suggest using \textbf{Simple} prompt as the default evaluation setting, since it achieves a strong balance between performance and prompt length.

\subsection{Benchmark Scoring}
\label{sec:score}

Given a caption and a multiple-choice question, the QA LLM can be in one of three situations:  (i) the caption supports the correct answer, (ii) the caption supports an incorrect answer, or (iii) the caption does not contain enough information to answer reliably.  To capture the third case explicitly, we append an additional option \emph{``Cannot answer from the caption.''} to every non-yes/no question. Then, we have two metrics:

\begin{itemize}[leftmargin=*]
    \item \textbf{Accuracy (Acc).} 
    Fraction of questions where the QA LLM selects the ground-truth option, reflecting correct/incorrect ratio.
    
    \item \textbf{Cannot ratio (Cannot).} 
    Fraction of questions where the QA LLM selects \emph{``Cannot answer''}, reflecting how often the caption fails to provide sufficient information (or the QA LLM judges it so).
\end{itemize}

To summarize both correctness and informative coverage in a single number, we define a per-question score as:
\[
s =
\begin{cases}
1, & \text{if selection is correct}, \\
0, & \text{if selection is incorrect}, \\
\frac{1}{K} + 0.05, & \text{if ``Cannot answer from the caption.''},
\end{cases}
\]
where $K$ is the number of semantic options (excluding the ``Cannot'' choice).  
The final score reported in our tables is the average of $s$ over all questions. We also randomly shuffle options order to avoid bias.

Thus, a caption is rewarded most when it enables the QA model to answer correctly, penalized most when it systematically leads to wrong answers. This design explicitly favors \emph{precision} over hallucinated detail~\cite{kalai2025languagemodelshallucinate}: a caption that says less, but avoids misleading content, is preferable to one that confidently encourages wrong answers.

\subsection{LLM QA Model Selection}
\label{sec:llm}

Because the text-only QA model is used to read captions and answer questions, its behavior directly determines the benchmark scores of all captioning MLLMs.  
We therefore do \emph{not} simply pick an off-the-shelf model (e.g., “just GPT-5”), but instead systematically evaluate candidate QA LLMs along four dimensions:

\begin{itemize}[leftmargin=*]
    \item \textbf{Faithfulness.}  
    When no caption is provided (or an empty string is given), an ideal QA model should almost always select \emph{``Cannot answer from the caption.''}. We measure the maximum attainable Cannot ratio in this setting to verify that the QA model does not hallucinate answers without textual evidence.

    \item \textbf{Efficiency.} We estimate the inference throughput in questions per second (QPS) and prefer models that achieve high QPS under our evaluation setup, since CaptionQA contains many MC questions.

    \item \textbf{Performance.}  
    We assess how well each QA model answers the questions given the captions. For this, we construct a fixed caption pool by randomly sampling one caption per question from all captioning MLLMs considered in this paper, and evaluate all candidate QA models on exactly the same caption-question pairs. We favor the LLM QA model that gives higher accuracy than the lower one.
    
    \item \textbf{Stability.}  
    We run each candidate QA model three times with temperature $0$ and measure the standard deviation of its accuracy.  
    Low variance indicates more deterministic.
\end{itemize}

\begin{figure}[t]
\centering
\includegraphics[width=0.65\columnwidth]{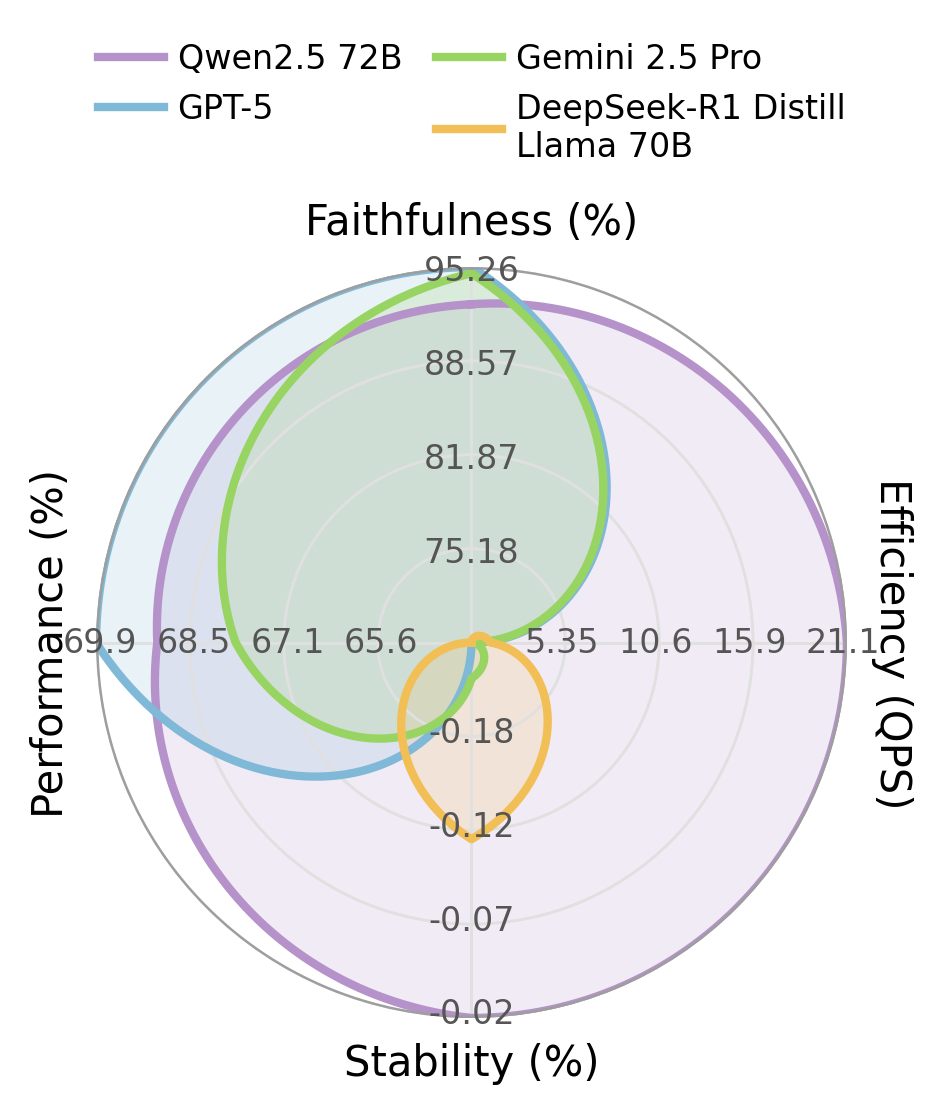}
\vspace{-0.5cm}
\caption{Comparison of text-only QA LLMs (GPT-5, Gemini 2.5 Pro, DeepSeek-R1 Llama 70B, Qwen2.5 72B) along four axes: faithfulness, efficiency (QPS), stability, and performance.}
\vspace{-0.5cm}
\label{fig:qa_llm}
\end{figure}

Based on this analysis on 4 different QA models as shown in Fig.~\ref{fig:qa_llm}, we adopt \textbf{Qwen2.5 72B} as the QA model for all main experiments. It offers a good trade-off: $21.14$ QPS with only $\pm 0.02\%$ variation in accuracy between runs, allowing \emph{the full benchmark to be evaluated in 25 minutes on a single AMD MI325 GPU}. Although its faithfulness (92.61\%) and accuracy (68.98\%) are slightly below GPT-5, it is roughly two orders of magnitude faster, making large-scale caption evaluation practical.

\section{Benchmark Construction}
\begin{figure*}[t]
    \centering
    \includegraphics[width=.97\textwidth]{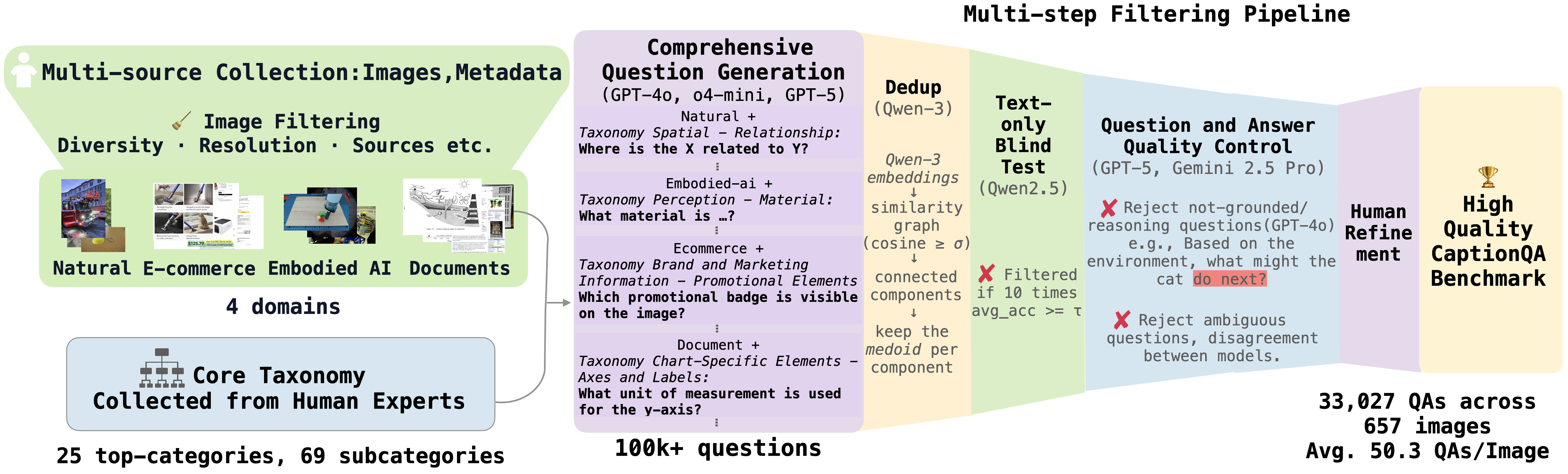} 
    \vspace{-0.5cm}
    \caption{\textbf{Benchmark construction pipeline.} Starting from a human-designed taxonomy and curated images for each domain, we use multiple generators to produce a large pool of taxonomy-guided questions. This pool is then refined by (1) embedding-based deduplication, (2) a text-only blind test to remove questions answerable from priors, (3) dual-VLM quality control to flag ungrounded or reasoning-heavy items, and (4) final human refinement, yielding high-quality, utility-focused QA pairs.}
    \label{fig:pipeline}
\end{figure*}

In this section, we describe how we construct CaptionQA on a given domain $\mathbf{D}$. The pipeline is fully domain-agnostic: the only domain-specific component is the taxonomy. Once a taxonomy is written for a new domain with target downstream needs, the same question generation pipeline can be applied without modification.

\subsection{Taxonomy Design}

For each domain $\mathbf{D}$, we design a two-level taxonomy that enumerates the types of information downstream tasks may require. The first level contains broad top-level categories (e.g., object existence, attributes, spatial relationships, actions, scene-level properties, hallucination in the Natural domain); the second level refines each into subcategories (e.g., \emph{Attribute} $\rightarrow$ color, shape, size, text, material, state).

We adopt a model-assisted, human-driven process. We first prompt GPT-5 with domain descriptions and examples to draft candidate taxonomies. Human experts then iteratively review, merge, and prune categories to ensure coverage of downstream utilities while avoiding redundancy and ill-defined entries. The final taxonomies are compact and domain-specific, yet follow a consistent structure across domains. Full taxonomies for all four domains are provided in the Supplementary.

\subsection{Image Collection}
Images are curated by human experts to cover the taxonomy, include diverse content, and stress challenging edge cases. For the \emph{Natural} domain, we self-collect real-world photographs and phone album images with diverse scenes and text-rich content. For the \emph{Document} domain, we capture screenshots of public-domain documents such as tax forms, financial reports, lecture materials, and office documents. For the \emph{E-commerce} domain, we screenshot desktop and mobile product pages across different platforms and brands. For the \emph{Embodied AI} domain, we sample single-view and multi-view frames from existing datasets (Ego4D, EgoExo4D, ScanNet, Open X-Embodiment) and manually select scenes spanning a range of tasks and environments.

\subsection{Question Generation Pipeline}

\subsubsection{Question generation.}
Given a domain-specific taxonomy, we automatically generate multiple-choice questions that target each utility dimension. For every domain (Natural, Document, E-commerce, Embodied AI), we expand each taxonomy node into a focused prompt (e.g., ``Object Existence $\rightarrow$ Object presence'', ``Table-Specific Elements $\rightarrow$ Table structure: presence of headers'', ``Functional and Semantic Understanding $\rightarrow$ Affordance recognition''). For each image and each taxonomy node, we query a multimodal generator model and ask it to produce one or more multiple-choice questions with options and a single correct answer, explicitly constrained to that aspect of the image; if no valid question can be formed, the generator returns nothing. We instantiate this step with three generator models (GPT-5, 4o, and o4-mini), and aggregate all outputs into a pool of candidate questions. Each question is stored together with its originating taxonomy node, so that downstream evaluation can be organized by domain, category, and subcategory.

\subsubsection{Filtering text-answerable questions.}
Many generated questions do not truly require visual grounding and can be answered from world knowledge alone. To remove such cases, we run a text-only filter using Qwen2.5-72B. For each multiple-choice question, we hide the image and only provide the question and options, explicitly instructing the model that it \emph{cannot see the image} and must answer with a single option letter. We repeat this process 10 times per question with randomly shuffled option orders and count how often the text-only model selects the correct answer. If its accuracy exceeds an option-dependent threshold (slightly above random chance), we treat the question as text-answerable and discard it; otherwise, we keep it. This stage prunes questions solvable without vision so that the remaining benchmark focuses on utilities that genuinely depend on the visual signal.

\subsubsection{Question deduplication.}
Different generators and taxonomy nodes can still produce many near-duplicate questions for the same image. We collapse these by per-image deduplication in embedding space. We encode each question plus its options using a text-embedding model (Qwen3-Embedding-8B), obtaining a $\ell_2$-normalized vector for every question. For each image, we compute cosine similarities between all embeddings and build a mutual $k$-NN graph with similarity threshold $\tau$; connected components define groups of semantically similar questions.We refine groups for cohesion, merge any remaining highly similar groups, and select a medoid (the most central question in embedding space) as the representative of each group. This yields, for every image, a compact set of canonical questions with their choices and answers and preserves links to all source models and taxonomy categories.

\subsection{Quality Control \& Human Verification}


\begin{table}[t]
\centering
\small
\caption{Question accuracy (\%) in the pre-quality and post-quality stages across four domains.}
\vspace{-0.3cm}
\begin{tabular}{lcccc}
\toprule
\textbf{Stage} & \textbf{Nat.} & \textbf{Doc.} & \textbf{E-com.} & \textbf{Emb. AI} \\
\midrule
Pre-quality  & 86 & 88 & 88 & 86 \\
Post-quality & 97 & 99 & 100 & 95 \\
\bottomrule
\end{tabular}
\vspace{-0.4cm}
\label{tab:pre_post_accuracy}
\end{table}

We combine automatic model-based validation with targeted human review to scale quality control to a large amount of questions. For each multiple-choice question, we show the image, question, and options to a vision–language model, but extend the option set with four meta-flags:
\textsc{Ambiguous\_Question}, 
\textsc{Unanswerable\_From\_Image}, 
\textsc{Not\_Suitable\_For\_Caption\_Eval}, 
and \textsc{None\_Of\_} \textsc{The\_Above}. 
The model is instructed to select one option.

We run this validation with GPT-5 and Gemini 2.5 Pro, and interpret their selections both as content flags and as independent answers. We keep only questions where \emph{both} validators choose the same option and that option matches the original answer; these high-confidence questions enter the benchmark directly. Any disagreement between the two models or with the ground truth, or any meta-flag, routes the question to human refinement. Annotators inspect the image and question, fix the answer or rephrase the question when possible, and discard items that cannot be made unambiguous and image-grounded. This workflow greatly reduces human effort by more than 90\% in question reviewing amount, while ensuring that all retained questions are visually answerable, well-posed, and suitable for caption evaluation. We sampled 50 questions per domain before and after this stage and report the question accuracy in Table ~\ref{tab:pre_post_accuracy}, showing accuracy significantly increased after careful human refinement. 

\begin{table*}[t]
\centering
\footnotesize
\caption{Performance comparison on the \textbf{CaptionQA }benchmark across four domains. All results are reported as score (\%). \textbf{Overall}
indicates average performance across all domains (gray column).}
\vspace{-0.3cm}
\label{tab:main_results}
\setlength{\tabcolsep}{2pt}
\begin{tabular*}{\linewidth}{@{\extracolsep{\fill}} c|c l c >{\columncolor[gray]{0.9}}c c c c c}
\toprule
\textbf{Prompt} & \textbf{Type} & \textbf{Model} & \textbf{Size} & \textbf{Overall} & \textbf{Natural} & \textbf{Document} & \textbf{E-commerce} & \textbf{Embodied AI} \\
\midrule
\multirow{10}{*}{\rotatebox[origin=c]{90}{Long}}
  & \multirow{7}{*}{\scriptsize \shortstack{Open-Source\\VLMs}} & Qwen3-VL ~\cite{yang2025qwen3} & 30B-A3B & \textbf{87.12} & \textbf{85.62} & 86.05 & \textbf{94.27} & \textbf{82.55} \\
  & & GLM-4.1V ~\cite{hong2025glm} & 9B & 85.83 & 82.92 & \textbf{88.34} & 93.03 & 79.01 \\
  & & Qwen2.5-VL ~\cite{bai2025qwen25vltechnicalreport} & 32B & 81.40 & 78.95 & 82.14 & 90.76 & 73.73 \\
  & & InternVL3.5 ~\cite{wang2025internvl3} & 38B & 80.11 & 78.70 & 78.78 & 87.53 & 75.41 \\
  & & InternVL3 ~\cite{zhu2025internvl3} & 14B & 73.81 & 72.41 & 71.88 & 84.05 & 66.89 \\
  & & NVLM-D ~\cite{dai2024nvlm} & 72B & 71.06 & 72.66 & 65.39 & 79.05 & 67.15 \\
  & & LLaVA-OneVision ~\cite{li2024llava} & 7B & 68.76 & 70.20 & 61.93 & 77.24 & 65.67 \\
\cmidrule{2-9}
  & \multirow{3}{*}[0.1em]{\scriptsize \shortstack{Proprietary\\VLMs}} & Gemini 2.5 Pro ~\cite{huang2025gemini} & -- & \textbf{90.12} & 89.44 & 88.67 & 95.60 & \textbf{86.78} \\
  & & GPT-5 & -- & 89.82 & \textbf{90.34} & \textbf{90.01} & \textbf{96.11} & 82.83 \\
  & & Claude Sonnet 4.5 & -- & 80.97 & 77.78 & 85.08 & 91.11 & 69.90 \\
\midrule
\multirow{10}{*}{\rotatebox[origin=c]{90}{Short}}
  & \multirow{7}{*}{\scriptsize \shortstack{Open-Source\\VLMs}} & GLM-4.1V & 9B & \textbf{60.87} & \textbf{62.43} & \textbf{55.69} & \textbf{64.05} & \textbf{61.29} \\
  & & Qwen3-VL & 4B & 57.49 & 60.52 & 48.12 & 61.69 & 59.63 \\
  & & InternVL3.5 & 38B & 54.27 & 56.52 & 44.24 & 59.16 & 57.16 \\
  & & InternVL3 & 8B & 53.31 & 56.50 & 44.58 & 56.99 & 55.17 \\
  & & Qwen2.5-VL & 32B & 52.88 & 54.74 & 45.32 & 57.89 & 53.58 \\
  & & NVLM-D & 72B & 48.67 & 51.29 & 40.99 & 50.88 & 51.53 \\
  & & LLaVA-OneVision & 7B & 47.13 & 51.01 & 38.28 & 49.80 & 49.41 \\
\cmidrule{2-9}
  & \multirow{3}{*}[0.1em]{\scriptsize \shortstack{Proprietary\\VLMs}} & Gemini 2.5 Pro & -- & \textbf{54.49} & \textbf{55.33} & 46.25 & \textbf{58.98} & 57.40 \\
  & & GPT-o4-mini & -- & 53.94 & 54.76 & \textbf{46.40} & 56.84 & \textbf{57.76} \\
  & & Claude Sonnet 4.5 & -- & 53.82 & 56.77 & 46.04 & 59.31 & 53.16 \\
\midrule
\multirow{10}{*}{\rotatebox[origin=c]{90}{Simple}}
  & \multirow{7}{*}{\scriptsize \shortstack{Open-Source\\VLMs}} & Qwen3-VL & 30B-A3B & \textbf{87.02} & \textbf{86.14} & 85.89 & \textbf{93.90} & \textbf{82.15} \\
  & & GLM-4.1V & 9B & 84.28 & 81.67 & \textbf{87.86} & 92.04 & 75.56 \\
  & & Qwen2.5-VL & 32B & 81.20 & 78.35 & 82.67 & 90.81 & 72.98 \\
  & & InternVL3.5 & 38B & 79.58 & 78.26 & 78.91 & 86.47 & 74.68 \\
  & & InternVL3 & 8B & 77.84 & 76.46 & 75.83 & 87.01 & 72.07 \\
  & & NVLM-D & 72B & 71.79 & 73.13 & 65.25 & 78.46 & 70.31 \\
  & & LLaVA-OneVision & 7B & 66.03 & 66.56 & 61.45 & 75.09 & 61.01 \\
\cmidrule{2-9}
  & \multirow{3}{*}[0.1em]{\scriptsize \shortstack{Proprietary\\VLMs}} & GPT-5 & -- & \textbf{90.29} & 88.78 & \textbf{90.81} & 94.73 & \textbf{86.82} \\
  & & Gemini 2.5 Flash & -- & 89.64 & \textbf{88.95} & 88.97 & \textbf{95.73} & 84.89 \\
  & & Claude Sonnet 4.5 & -- & 78.95 & 76.56 & 83.09 & 88.86 & 67.27 \\
\midrule
\multirow{10}{*}{\rotatebox[origin=c]{90}{Taxonomy-Hinted}}
  & \multirow{7}{*}{\scriptsize \shortstack{Open-Source\\VLMs}} & Qwen3-VL & 8B & \textbf{77.21} & \textbf{78.99} & 70.82 & \textbf{84.19} & \textbf{74.85} \\
  & & GLM-4.1V & 9B & 76.18 & 75.87 & \textbf{72.40} & 83.87 & 72.59 \\
  & & Qwen2.5-VL & 32B & 74.91 & 74.64 & 71.44 & 83.55 & 70.02 \\
  & & NVLM-D & 72B & 68.20 & 72.63 & 59.61 & 76.12 & 64.42 \\
  & & InternVL3 & 8B & 67.48 & 71.57 & 55.05 & 75.82 & 67.46 \\
  & & InternVL3.5 & 38B & 65.26 & 69.30 & 51.94 & 73.46 & 66.35 \\
  & & LLaVA-OneVision & 7B & 62.91 & 63.24 & 56.57 & 69.78 & 62.03 \\
\cmidrule{2-9}
  & \multirow{3}{*}[0.1em]{\scriptsize \shortstack{Proprietary\\VLMs}} & Gemini 2.5 Pro & -- & \textbf{86.42} & \textbf{87.47} & 81.34 & \textbf{91.50} & \textbf{85.35} \\
  & & GPT-5 & -- & 83.24 & 86.87 & 72.36 & 88.63 & 85.10 \\
  & & Claude Sonnet 4.5 & -- & 80.19 & 77.36 & \textbf{83.23} & 89.74 & 70.44 \\
\bottomrule
\end{tabular*}
\vspace{-0.1cm}
\end{table*}

\vspace{-0.3cm}
\section{A Gap between Caption and Image Utility}
\vspace{-0.5cm}
\begin{figure}[htb!]
\centering
\includegraphics[width=0.95\columnwidth]{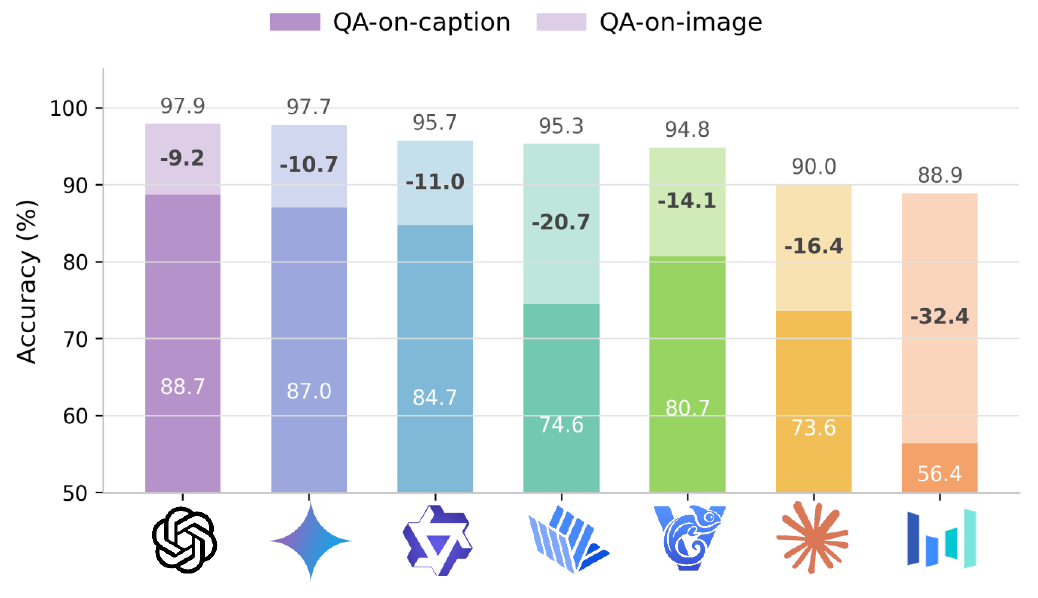}
\vspace{-0.5cm}
\caption{Overall gap between QA-on-image and QA-on-caption for GPT-5, Gemini-2.5-Pro,
Qwen3-VL-30B-A3B, GLM-4.1V-9B, InternVL3.5-38B, Claude-Sonnet-4.5, and LLaVA-OV-7B.
Each bar shows the difference in CaptionQA Acc., averaged over the four domains.}
\label{fig:overall_gap}
\end{figure}
\vspace{-0.5cm}

\begin{table}[t]
\centering
\footnotesize
\setlength{\tabcolsep}{4pt}
\caption{Domain-wise gap (\%) between QA-on-caption and QA-on-image. 
Values are absolute score differences (lower is better).}
\label{tab:caption_image_gap}
\vspace{-0.3cm}
\begin{tabular}{lcccc}
\toprule
\textbf{Model} & \textbf{Nat.} & \textbf{Doc.} & \textbf{E-com.} & \textbf{Emb. AI} \\
\midrule
\multicolumn{5}{c}{\textbf{Open-Source VLMs}} \\
\midrule
Qwen3-VL-30B-A3B        & 12.09 & 10.06 &  4.87 & 16.96 \\
GLM-4.1V-9B             & 17.12 &  7.79 &  6.62 & 24.86 \\
InternVL3.5-38B         & 22.41 & 18.49 & 14.51 & 27.49 \\
LLaVA-OV-7B             & 34.14 & 28.73 & 24.97 & 41.81 \\
\midrule
\multicolumn{5}{c}{\textbf{Proprietary VLMs}} \\
\midrule
GPT-5                   & 11.30 &  6.72 &  4.96 & 13.81 \\
Gemini-2.5-Pro          & 12.02 & 10.11 &  5.03 & 15.78 \\
Claude-Sonnet-4.5       & 19.39 &  8.47 &  8.80 & 29.06 \\
\bottomrule
\end{tabular}
\vspace{-0.3cm}
\end{table}
\vspace{-0.3cm}

\paragraph{Proprietary models lead in CaptionQA.} We evaluate 24 MLLMs under four caption prompts, with full results in the Supplementary. As summarized in Table~\ref{tab:main_results}, the proprietary models lead in the overall CaptionQA score: GPT-5 (90.3\% on the Simple prompt) and Gemini 2.5 Pro (around 90.1\% on the Long prompt) are the strongest overall, while Qwen3-VL-30B-A3B (87.1\%) and GLM-4.1V-9B (85.8\%) on the Simple prompt are the main open-source models. Scores are also highly domain-dependent: E-commerce is the easiest setting (typically 77--96\%), whereas Embodied AI is the most challenging (around 66--87\%).
\vspace{-0.3cm}
\paragraph{A substantial gap between caption utility and image utility.} 
To reveal the gap, we run CaptionQA in a QA-on-image setting, where models answer the same multiple-choice questions directly from images instead of captions. As shown in Figure~\ref{fig:overall_gap}, the questions themselves are not particularly hard when the image is available: GPT-5, for example, reaches close to 98\% accuracy on QA-on-image. However, when the same model must answer via its own captions (QA-on-caption), performance drops by roughly 9\%. Across strong proprietary models, this utility gap is consistently around 9.2-16.4\%; for open-source models, the gap is substantially larger, ranging from \(\sim\)11\% (Qwen3-VL-30B-A3B) to 32.4\% (LLaVA-OneVision-7B).

The gap is not simply a reflection of baseline multimodal ability. Models that are relatively close on QA-on-image can have a huge difference on QA-on-caption performance. Even with slight QA-on-image ability decreases, the utility gap increases significantly. For example, GPT-5 and LLaVA-OneVision-7B differ by only about 9\% in QA-on-image accuracy, yet their QA-on-caption scores differ by 32.3\%. Even more notably, Claude Sonnet 4.5 and LLaVA-OneVision-7B are separated by only 1.1\% on QA-on-image, but their caption-based scores differ by 17.2\%. In other words, standard QA-on-image benchmarks like MMBench or MME would consider these models “close,” while CaptionQA reveals that their captions carry very different amounts of usable information for downstream reasoning.
\vspace{-0.5cm}
\paragraph{Non-uniform gap across domains.}
Domain-wise patterns in Table~\ref{tab:caption_image_gap} show that this utility gap is not uniform. E-commerce has the smallest gaps for all models (around 4.87-24.97\%), suggesting that product-centric information is relatively easy to preserve in text. In contrast, Embodied AI exhibits the largest gaps: even the best open-source model (Qwen3-VL-30B-A3B) loses 16.96\% when switching from images to captions, and LLaVA-OneVision-7B loses over 40\%. Proprietary models also struggle here, with gaps of 13.81\% (GPT-5) and 29.06\% (Claude Sonnet 4.5), suggesting that the utility of robotic-related expression for image caption needs serious attention in the field of MLLM. Natural and Document domains sit in between, with moderate gaps driven by missed spatial relations, fine-grained attributes, or document-structure details.

\subsection{Can complex prompt mitigate the gap?}
\vspace{-0.5cm}
\begin{figure}[htb]
\centering
\includegraphics[width=\columnwidth]{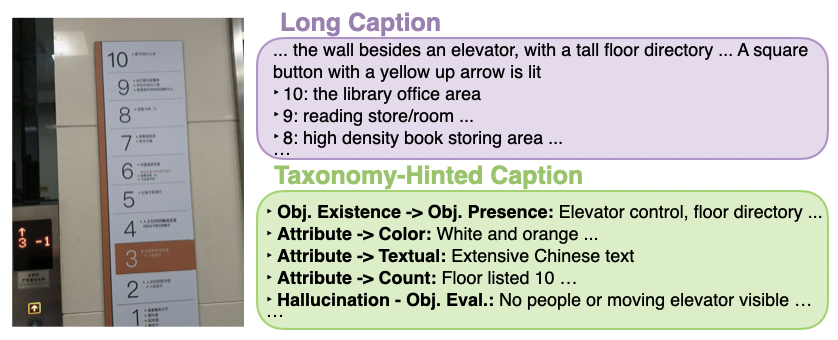}
\vspace{-0.5cm}
\caption{Qualitative example of caption under a complex prompt, Taxonomy-Hinted. Although GPT-5 is instructed to describe the image and focus on provided aspects, it outputs in a fill-in-the-blank style and provide much less information than Long prompt.}
\label{fig:example}
\vspace{-0.5cm}
\end{figure}
\vspace{-0.3cm}
\paragraph{More complex prompts can backfire.} We first examine whether providing explicit captioning-focus through Taxonomy-Hinted prompts improves caption utility. Contrary to expectation, moving from Long to Taxonomy-Hinted prompts degrades performance: 23 of 25 categories show decline (mean $-10.8\%$), with particularly large drops in Document Domain-Specific Evaluation ($-33.1\%$) and Embodied AI Perception ($-7.8\%$). Only two categories show minor gains ($+0.4\%$ and $+2.0\%$).

In our example as shown in Figure~\ref{fig:example}, models frequently fail to interpret complex instructions as conceptual guidance for coverage; instead, they imitate the instruction format literally, producing enumerated or template-like sentences that superficially mention required concepts but fail to describe them adequately. This behavior indicates a \textbf{shift from content grounding to format imitation}, revealing a tension between instruction-following and visual understanding. Thus, the additional prompt complexity \textbf{hurts rather than helps.}

\subsection{Can longer caption mitigate the gap?}

\paragraph{Length shows diminishing returns.} Analyzing the progression from Short to Long reveals a clear pattern of diminishing returns. Moving from Short to Simple (average 21 to 317 words caption) yields mean gains of +33.8\% across all categories. This single transition captures 99\% of the benefit from going all the way to Long (+34.2\%), despite using only 67\% of the caption length. The Simple prompt already elicits most useful information from models.

Further increasing length from Simple to Long (317 to 471 words, a 1.5$\times$ increase) adds minimal value: only +0.35\% mean improvement, with all 25 categories showing less than 2\% change. Additional verbosity does not translate to improved utility.

Domain-specific patterns emerge. Document structural elements and E-commerce textual content gain +41-56\% from Short to Simple, indicating \emph{verbalization bottlenecks}--information is present but under-described. Embodied AI categories gain only +6-34\%, indicating \emph{information bottlenecks}--required details (activity and task context) are still missing in longer caption. These patterns suggest different domains require different caption strategies: verbose descriptions benefit Document and E-commerce, while Embodied AI requires richer visual understanding beyond current model capabilities.

Simple prompts achieve near-optimal performance (99\% of Long's benefits) at reasonable length, avoiding both the inadequacy of Short captions and the potential quality degradation of Long/Taxonomy-Hinted prompts. Table~\ref{tab:main_results} reports all prompt scores for completeness; we recommend Simple as the default for practical use.
\section{Conclusion}

We introduce CaptionQA, a new benchmark that evaluates captions from a utility-first perspective, which measures how well captions support downstream tasks. Our protocol combines a deterministic QA evaluator with a carefully constructed benchmark that avoids shortcut behaviors and enforces quality checks, yielding stable and interpretable scores across models and domains. By open-sourcing the full benchmark construction pipeline, we make it easy to extend CaptionQA: researchers can target new applications by specifying a domain-specific taxonomy and running our question-generation pipeline, immediately obtaining utility-centric caption evaluations. We hope CaptionQA will serve as a practical foundation for building, selecting, and deploying captioning systems that are aligned with real-world uses of MLLMs.
{
    \small
    \bibliographystyle{ieeenat_fullname}
    \bibliography{main}
}

\clearpage
\setcounter{page}{1}
\maketitlesupplementary

\section{Motivation and Overview}

\begin{figure*}[t]
    \centering
    \includegraphics[width=\textwidth]{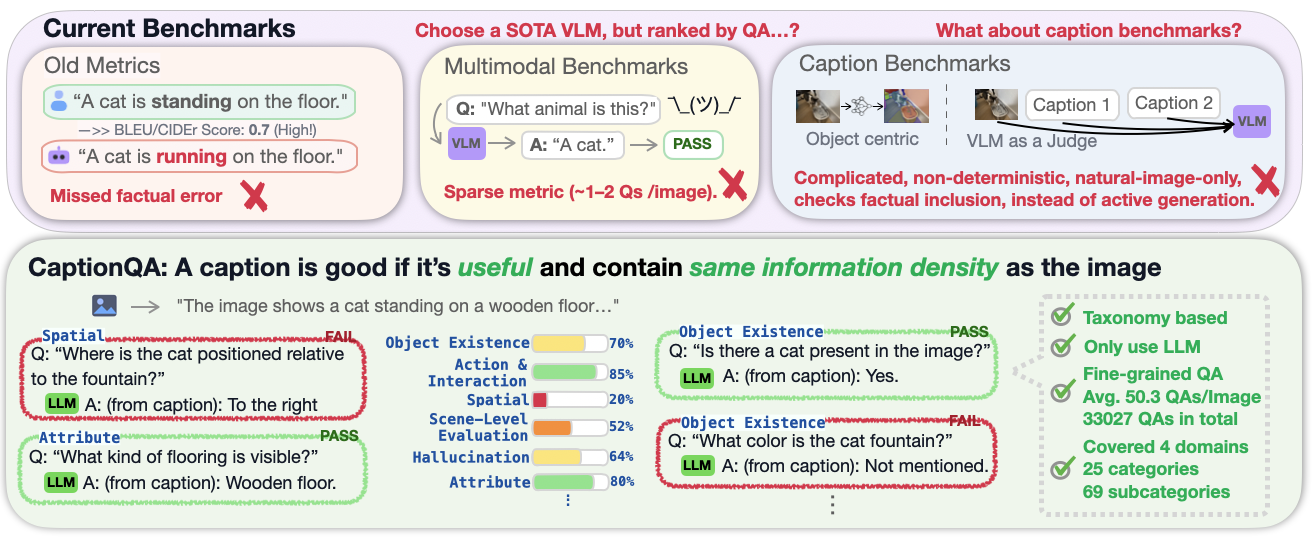}
    \caption{\textbf{Defining and evaluating ``useful'' captions.} Existing practices are either fact-blind (text-similarity metrics) or test a different task with sparse supervision (multimodal benchmarks), or rely on complex, non-deterministic pipelines (caption benchmarks). CaptionQA instead measures how ``useful'' a caption is by testing whether it can stand in for the image on dense, taxonomy-driven QA, and yields fine-grained diagnostics across domains and aspects.}
    \label{fig:motivation}
\end{figure*}

Figure~\ref{fig:motivation} provides a conceptual overview of CaptionQA's evaluation approach. Unlike traditional text-similarity metrics that are fact-blind, multimodal benchmarks that test a different task with sparse supervision, or complex non-deterministic caption evaluation pipelines, CaptionQA measures how ``useful'' a caption is by testing whether it can stand in for the image on dense, taxonomy-driven question answering. This yields fine-grained diagnostics across domains and aspects, directly measuring the task-relevant information preserved in captions.

\section{Question Characteristics}

\begin{figure}[htb!]
\centering
\includegraphics[width=\columnwidth]{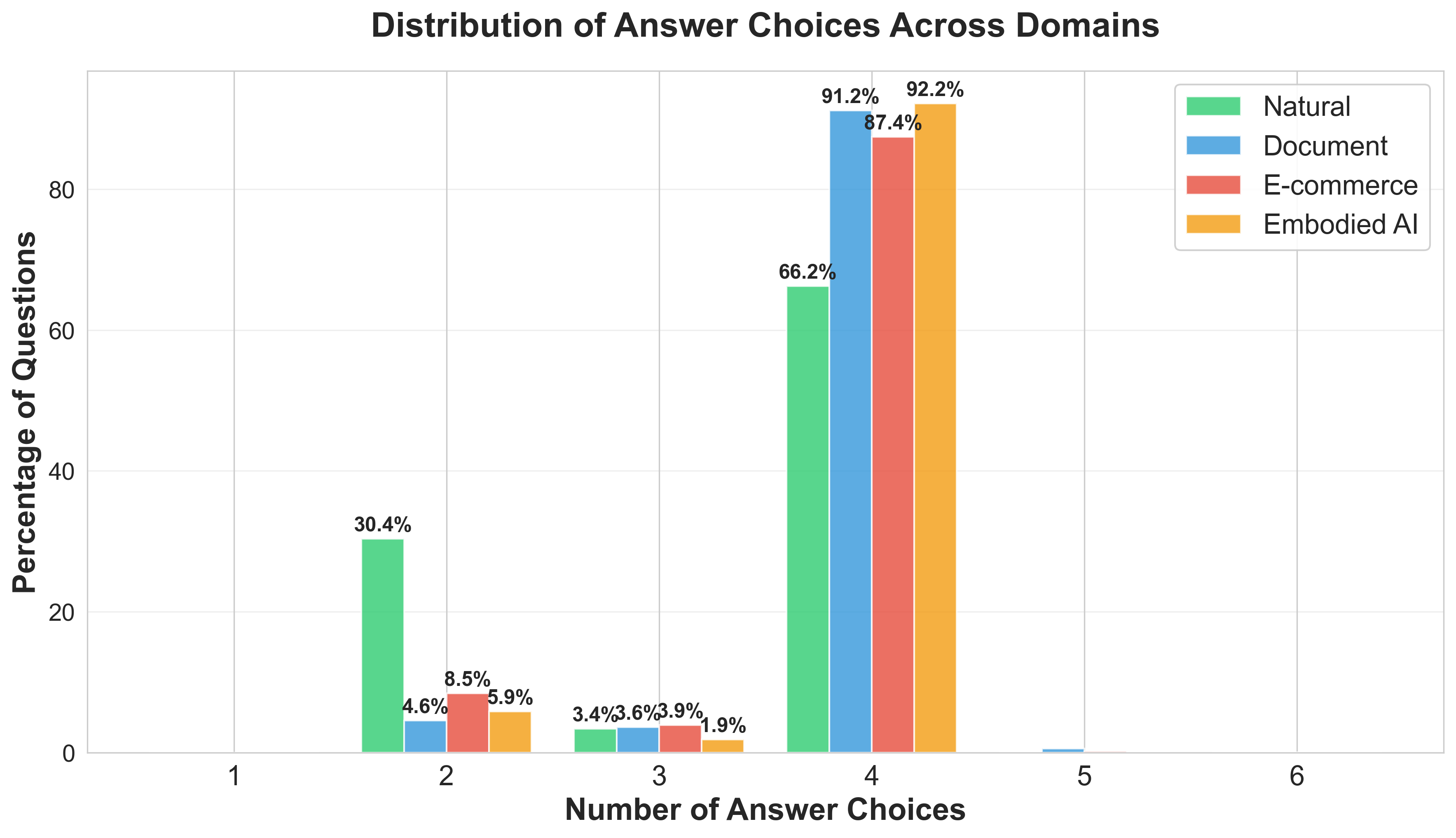}
\caption{Distribution of answer choices across domains. Most questions are 4-choice (87--92\%), making the benchmark more challenging than binary VQA. Natural domain has more binary questions due to attribute verification.}
\label{fig:choices_dist}
\end{figure}

Our pipeline generates predominantly 4-choice multiple-choice questions, which are more challenging than binary yes/no questions. As shown in Figure~\ref{fig:choices_dist}, 87--92\% of questions across domains are 4-choice, with the remaining split between 2-choice and 3-choice questions. The Natural domain has a higher proportion of binary questions (30.4\%) due to yes/no attribute verification queries (e.g., ``Is there a cat in the image?'', ``Is the door open?''). This distribution reflects the taxonomy-driven generation process, where attribute and hallucination categories naturally lend themselves to binary verification, while other categories require distinguishing between multiple plausible options.

\subsection{Question Density Across Domains}

\begin{figure}[htb!]
\centering
\includegraphics[width=0.95\columnwidth]{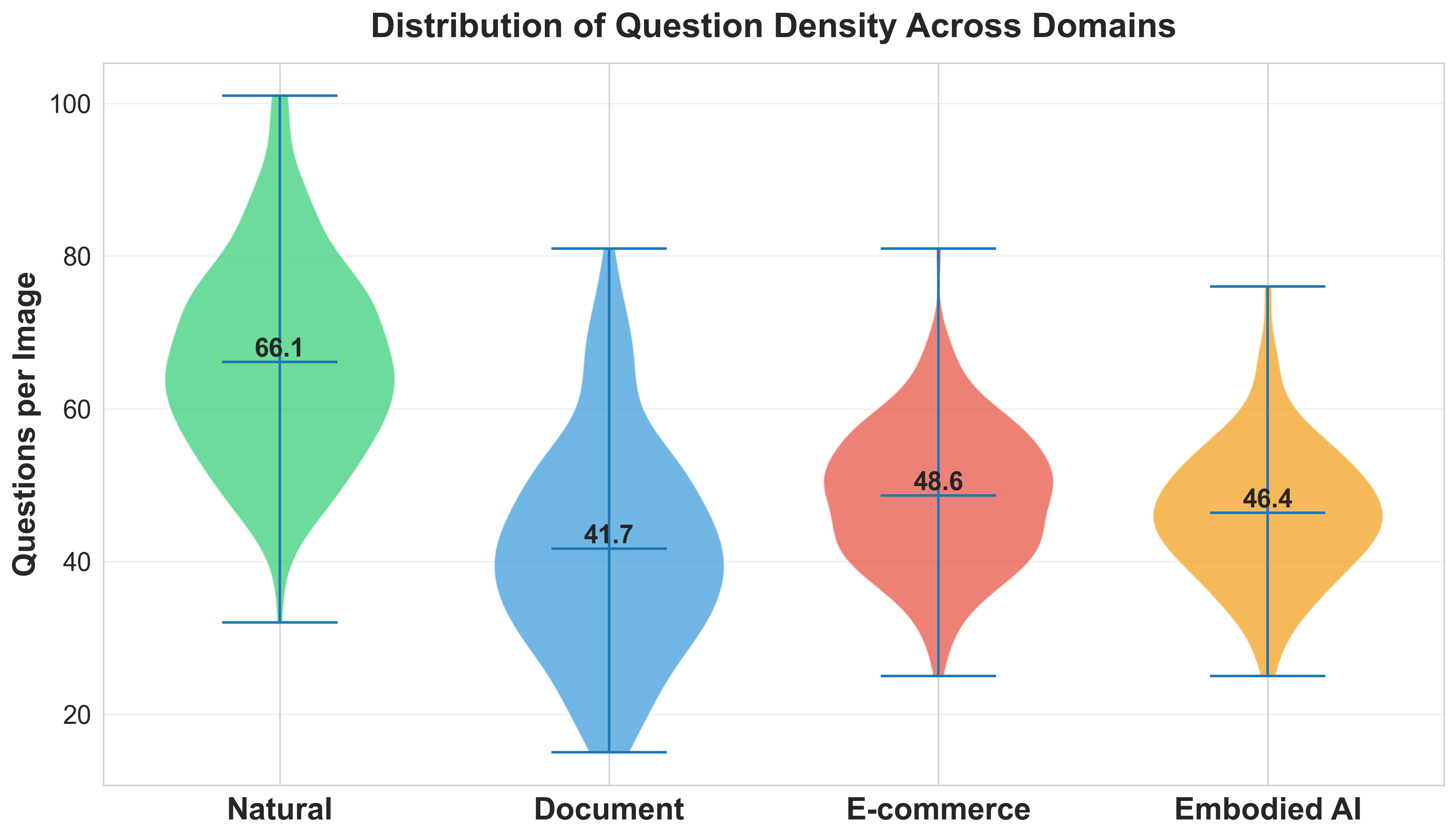}
\caption{Distribution of question density across domains. The violin plots show the distribution of questions per image, with Natural images supporting the highest density of diverse questions and Document images focusing on specific structural and content elements. The consistent distributions within each domain (low variance) demonstrate systematic annotation quality.}
\label{fig:question_density}
\end{figure}

The benchmark exhibits consistent question density across images within each domain, with low variance indicating systematic annotation quality. As shown in Figure~\ref{fig:question_density}, the Natural domain supports the highest density (66.1 questions per image on average) due to rich visual content including objects, attributes, spatial relationships, and potential hallucinations. The Document domain, while having the lowest average (41.7 questions per image), still provides comprehensive coverage of structural elements, content evaluation, and domain-specific aspects. E-commerce and Embodied AI domains show similar densities (48.6 and 46.4 respectively), reflecting their focus on product attributes and task-relevant perception.

\begin{figure}[htb!]
\centering
\includegraphics[width=\columnwidth]{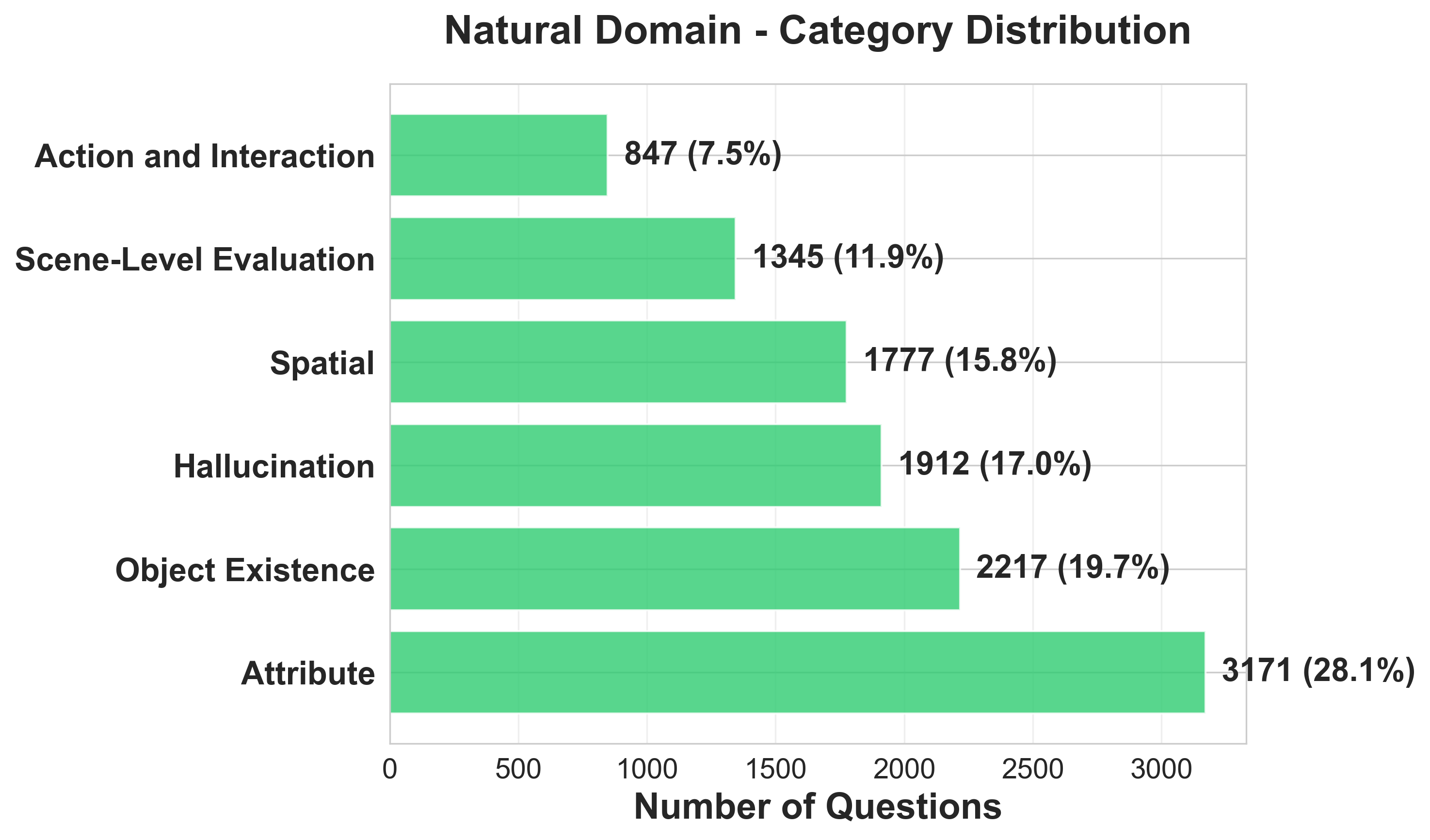}
\caption{\textbf{Natural Domain:} Question distribution across top-level taxonomy categories. Attribute questions dominate, followed by Object Existence and Hallucination detection.}
\label{fig:category_natural}
\end{figure}

\begin{figure}[htb!]
\centering
\includegraphics[width=\columnwidth]{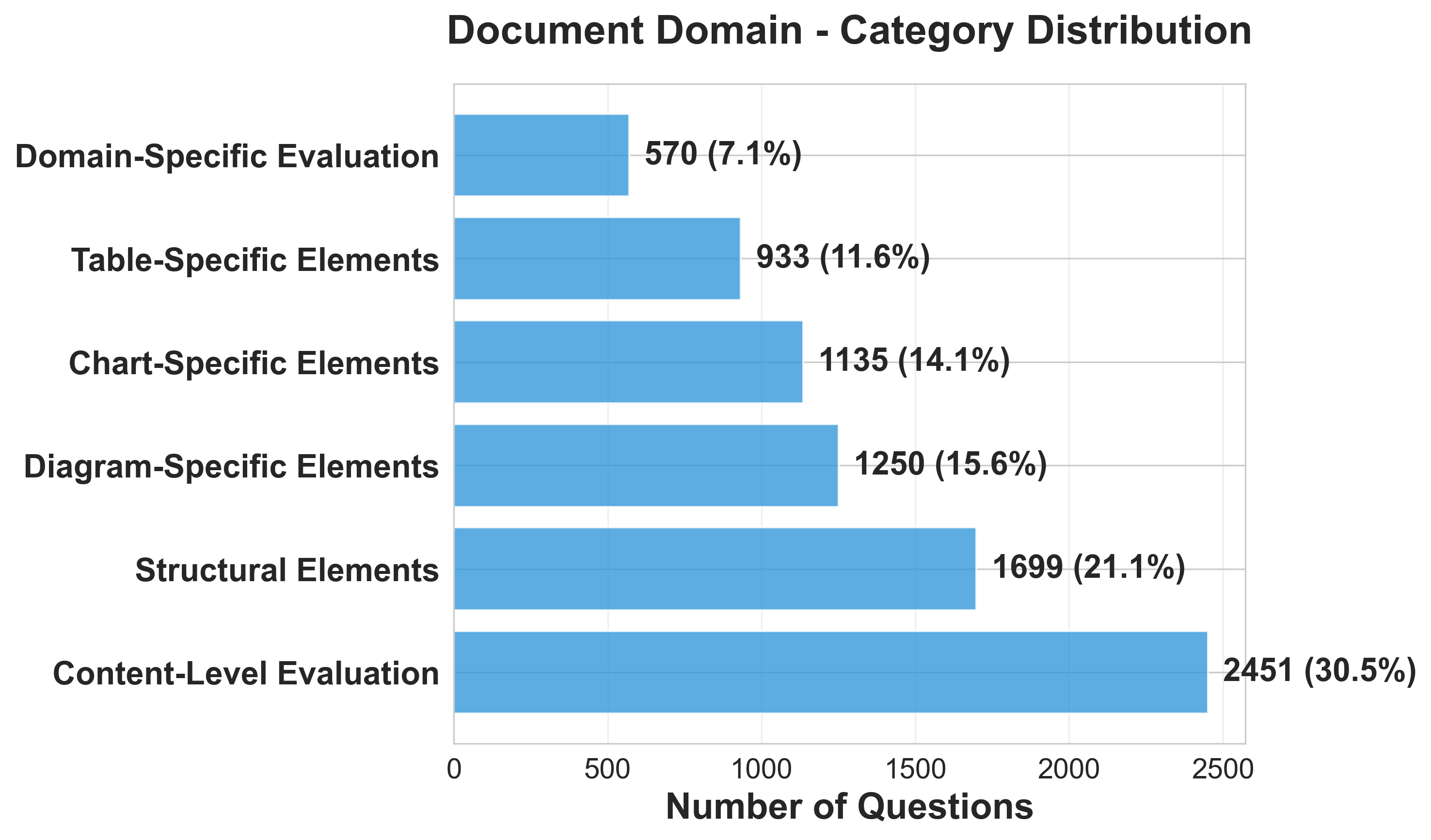}
\caption{\textbf{Document Domain:} Question distribution across top-level taxonomy categories. Content-Level Evaluation and Structural Elements are the primary focus.}
\label{fig:category_document}
\end{figure}

\begin{figure}[htb!]
\centering
\includegraphics[width=\columnwidth]{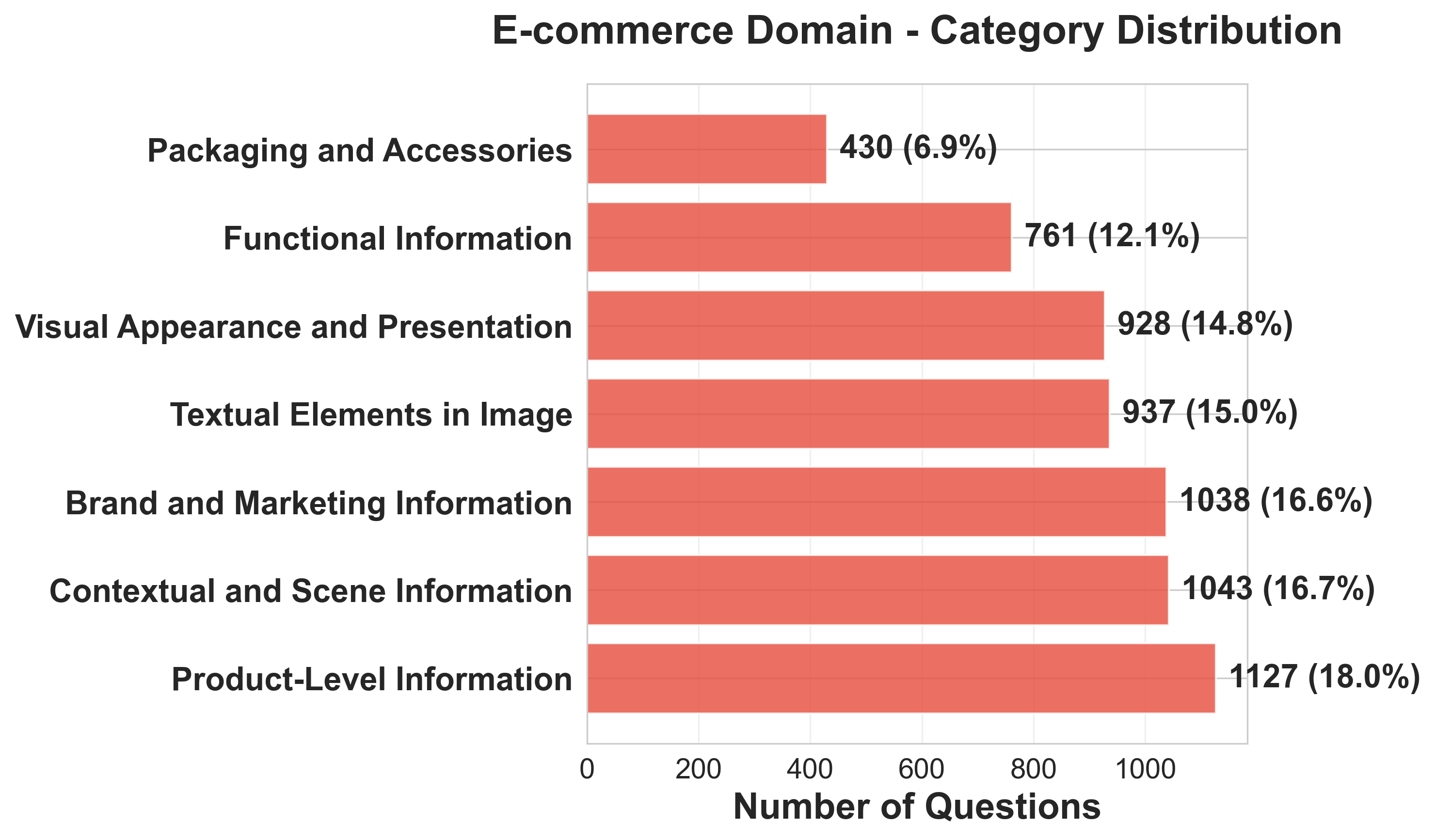}
\caption{\textbf{E-commerce Domain:} Question distribution across top-level taxonomy categories. Questions are evenly distributed across product information, context, and marketing aspects.}
\label{fig:category_ecommerce}
\end{figure}

\begin{figure}[htb!]
\centering
\includegraphics[width=\columnwidth]{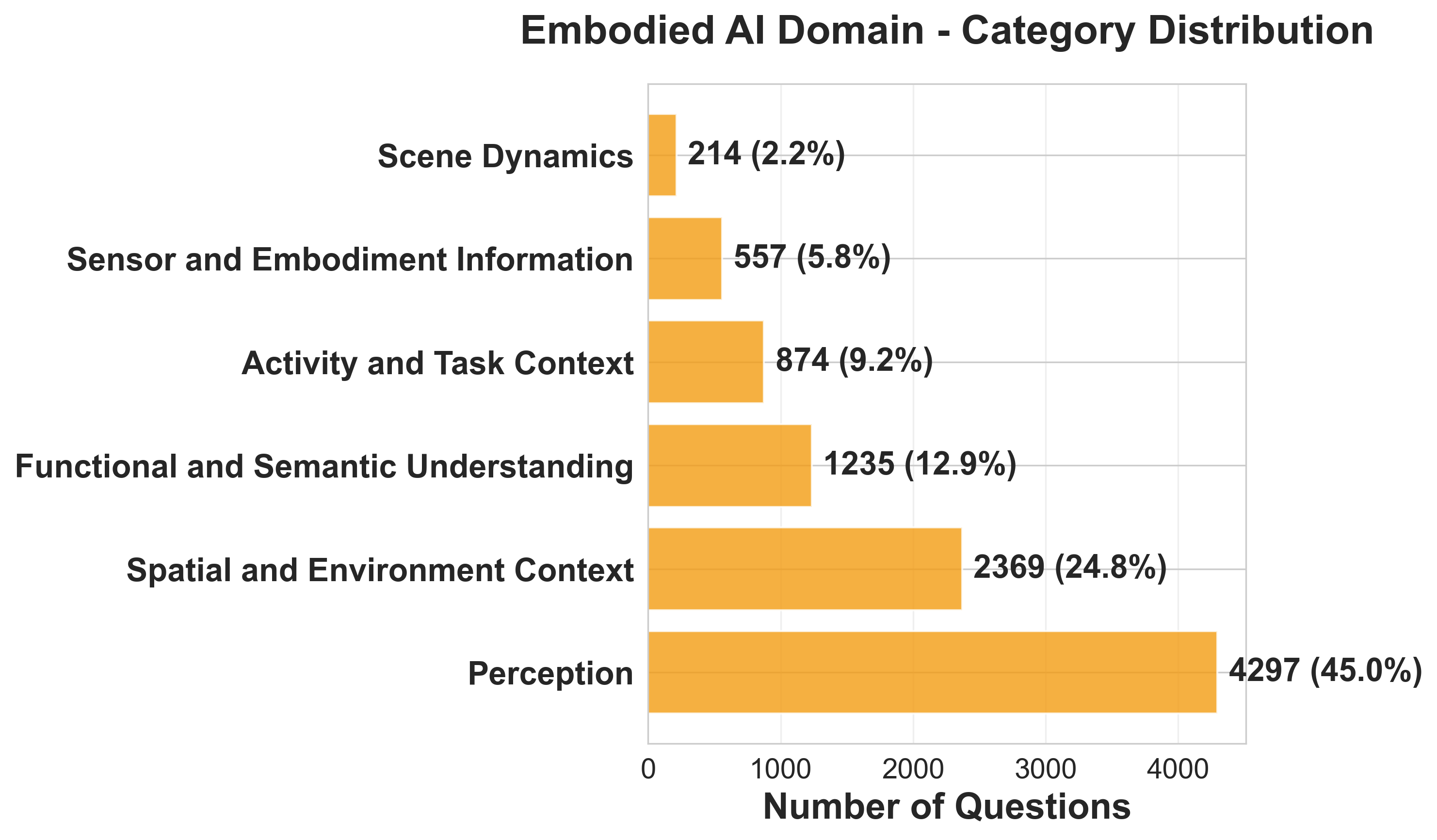}
\caption{\textbf{Embodied AI Domain:} Question distribution across top-level taxonomy categories. Perception and Spatial Context dominate, reflecting robotics task requirements.}
\label{fig:category_embodiedai}
\end{figure}

Figures~\ref{fig:category_natural}--\ref{fig:category_embodiedai} show the distribution of questions across top-level taxonomy categories for each domain. In the Natural domain (Figure~\ref{fig:category_natural}), Attribute questions are most prevalent (28.1\%), followed by Object Existence (19.7\%) and Hallucination (17.0\%). Document questions (Figure~\ref{fig:category_document}) focus heavily on Content-Level Evaluation (30.5\%) and Structural Elements (21.1\%). E-commerce questions (Figure~\ref{fig:category_ecommerce}) are more evenly distributed across Product Information (18.0\%), Contextual Information (16.7\%), and Brand/Marketing (16.6\%). Embodied AI questions (Figure~\ref{fig:category_embodiedai}) emphasize Perception (45.0\%) and Spatial Context (24.8\%), reflecting the task-oriented nature of robotics applications.

\section{Caption Prompts}
\label{sec:caption_prompts}
Caption quality is highly sensitive to the instruction given to the MLLM. To study how prompting affects the utility of generated captions, we evaluate each model under four captioning prompts, shared across all domains:

\begin{itemize}[leftmargin=*,noitemsep]
\item \textbf{Long.} ``Write a very long and detailed caption describing the given image as comprehensively as possible.''
\item \textbf{Short.} ``Write a very short caption for the given image.''
\item \textbf{Simple.} ``Describe this image in detail.''
\item \textbf{Taxonomy-Hinted.} We explicitly condition the caption on our domain-specific taxonomy. Concretely, we ask the model to ``Describe this image from the following perspectives. Skip any aspect that does not apply.'' and then list taxonomy nodes in the form \texttt{Top-category -> Subcategory}, e.g., \texttt{Object Existence -> Object presence}, \texttt{Attribute -> Color}, etc.
\end{itemize}

\begin{figure}[h!]
\centering
\includegraphics[width=\columnwidth]{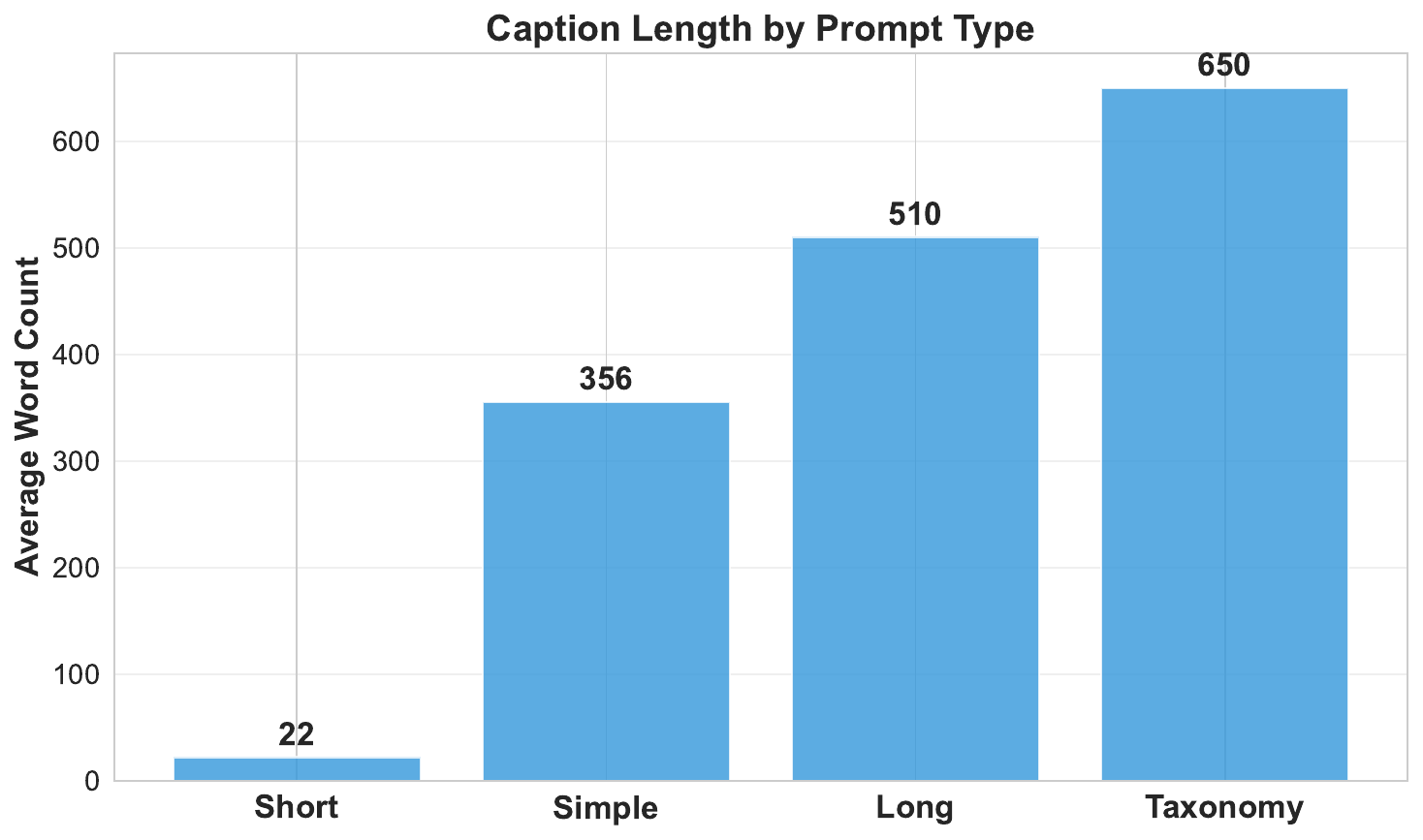}
\caption{Average caption length (word count) by prompt type, averaged across all models and domains. The Taxonomy-Hinted prompt produces the longest captions (650 words on average), followed by Long (510 words), Simple (356 words), and Short (22 words).}
\label{fig:caption_length}
\end{figure}

These prompts yield captions of substantially different lengths, as shown in Figure~\ref{fig:caption_length}. On average across all models and domains, the \textbf{Short} prompt produces captions of 22 words, \textbf{Simple} produces 356 words, \textbf{Long} produces 510 words, and \textbf{Taxonomy-Hinted} produces 650 words. The Taxonomy-Hinted prompt produces the longest captions, as explicitly providing the taxonomy encourages models to address all categories systematically. For the standard CaptionQA evaluation, we suggest adopting \textbf{Simple}, as it demonstrates good performance across models while maintaining reasonable caption length.

\section{Taxonomy Structure}
  \label{sec:appendix_taxonomy}

\begin{figure*}[htb!]
  \centering
  \includegraphics[width=0.95\textwidth]{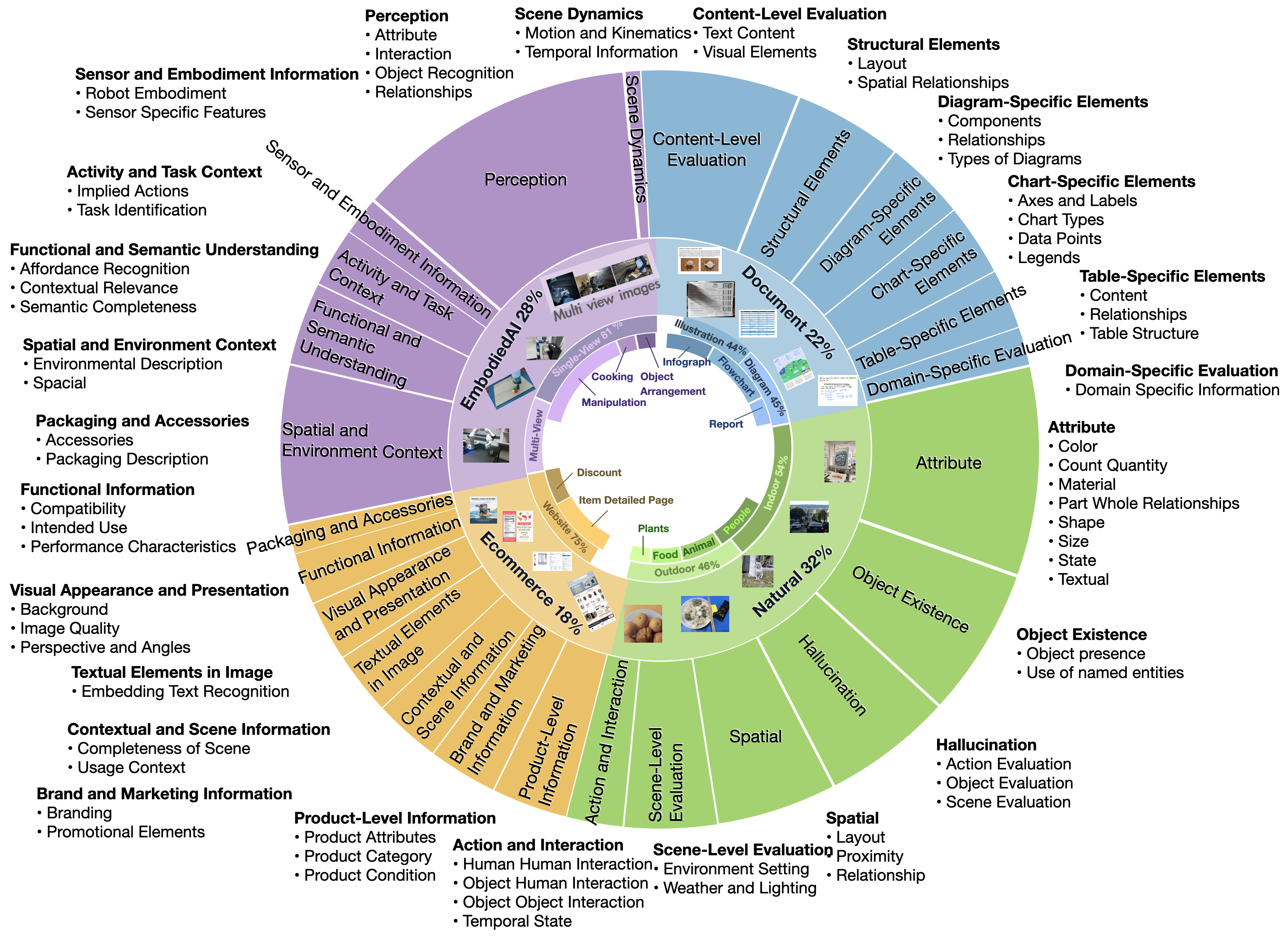}
  \caption{\textbf{Taxonomy structure across all four CaptionQA domains}.
  \textbf{(1) Natural domain} contains 6 top-level and 22 subcategories,
  emphasizing object properties, spatial relationships, and hallucination
  detection. \textbf{(2) Document domain}  contains 6 top-level and 15
  subcategories, focusing on structural elements, content evaluation, and
  document-specific features. \textbf{(3) E-commerce domain}contains
  7 top-level and 16 subcategories, covering product attributes, visual
  presentation, and marketing information. \textbf{(4) Embodied AI domain}  contains 6 top-level and 16 subcategories, prioritizing
  perception, spatial understanding, and task-relevant features for robotics
  applications.}
  \label{fig:taxonomy_overview_supp}
  \end{figure*}

  Figure~\ref{fig:taxonomy_overview_supp} presents the complete hierarchical taxonomy structure across all four CaptionQA domains. The taxonomy guides question generation and ensures comprehensive coverage of domain-specific aspects that captions should capture.

  Each domain is organized into top-level categories (6--7 per domain) and their corresponding subcategories (15--22 per domain, 69 total across all domains). The hierarchical structure balances comprehensive coverage with manageable annotation complexity, where top-level categories capture broad semantic areas while subcategories provide specific evaluation dimensions.

  \subsection{Design Rationale}
   Although CaptionQA uses QA-style proxy, the QAs are derived from carefully designed taxonomy of each domain, reflecting the actual information needed in real downstream applications of each domain, such as retrieval, ranking, matching, classification, form filling, agentic pipelines. Each domain's taxonomy was designed through an iterative human-in-the-loop process. Domain experts from industry partners drafted initial categories based on downstream task requirements, which were then refined through several rounds of discussion and generative model refinement. The resulting taxonomies are domain-specific (emphasizing aspects relevant to each domain's applications), comprehensive (covering all salient visual information without redundancy), balanced (no single category dominates, with a maximum of 46.3\% for Perception in Embodied AI), and practical (guiding question generation while remaining manageable for annotation). This
  taxonomy-driven approach ensures CaptionQA evaluates the \emph{right} information--what downstream tasks actually need---rather than arbitrary caption properties. 

\section{Image Amount Justification}
Instead of collecting tens of thousands of loosely annotated images as in most multimodal benchmarks, CaptionQA adopts a \emph{high-density} design: each image is paired with an average of 50 carefully curated, taxonomy-grounded questions. This dense annotation strategy makes each image substantially more informative than those in traditional VQA benchmarks, where a single image typically supports only 1--3 questions. Moreover, unlike benchmarks that evaluate short multiple-choice answers, caption evaluation requires generating full-sentence outputs for each image. \emph{Increasing the number of images across domains would therefore linearly inflate the total evaluation time}, as caption generation latency grows with both model size and output length. Our design thus strikes a deliberate balance between semantic coverage and evaluation efficiency, enabling comprehensive yet tractable assessment of multimodal understanding.

To validate that our image amount design provides sufficient data for reliable model evaluation, we analyze ranking stability as a function of dataset size. For each domain, we randomly sample subsets of size $k \in [1, N]$ (where $N$ is the total number of images), compute model performance on each subset, and repeat this 10 times per sample size to account for selection variance.

\begin{figure*}[htb!]
\centering
\includegraphics[width=0.95\textwidth]{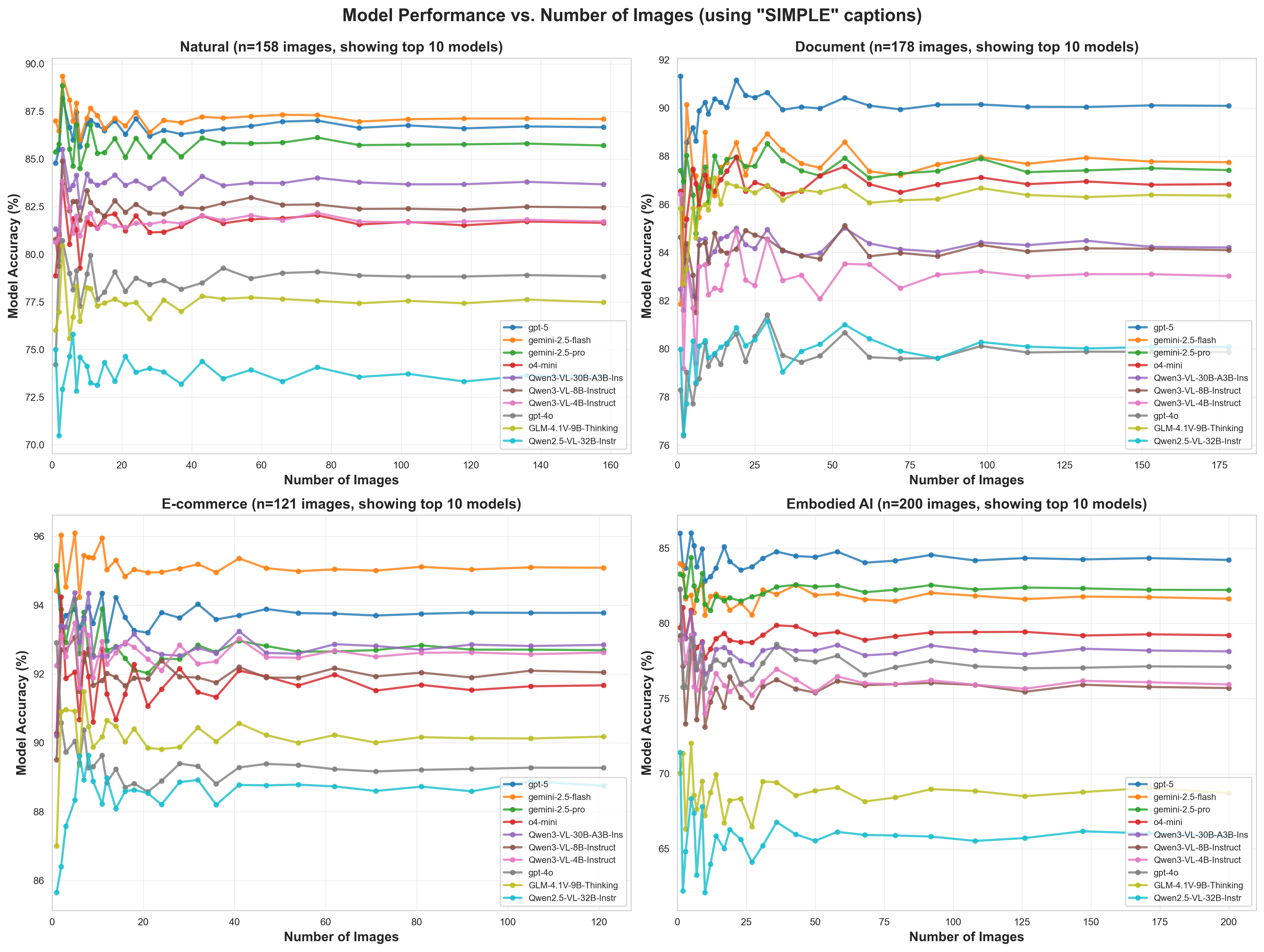}
\caption{\textbf{Model ranking stability vs. number of images (accuracy-based).}
Each line represents one model's accuracy trajectory as more images are randomly sampled (10 trials per sample size). Same color indicates the same model across domains. Performance curves plateau rapidly and maintain relative positions, validating data sufficiency. Top 10 models shown (ranked by average performance across domains).}
\label{fig:ranking_stability_accuracy}
\end{figure*}

\begin{figure*}[htb!]
\centering
\includegraphics[width=0.95\textwidth]{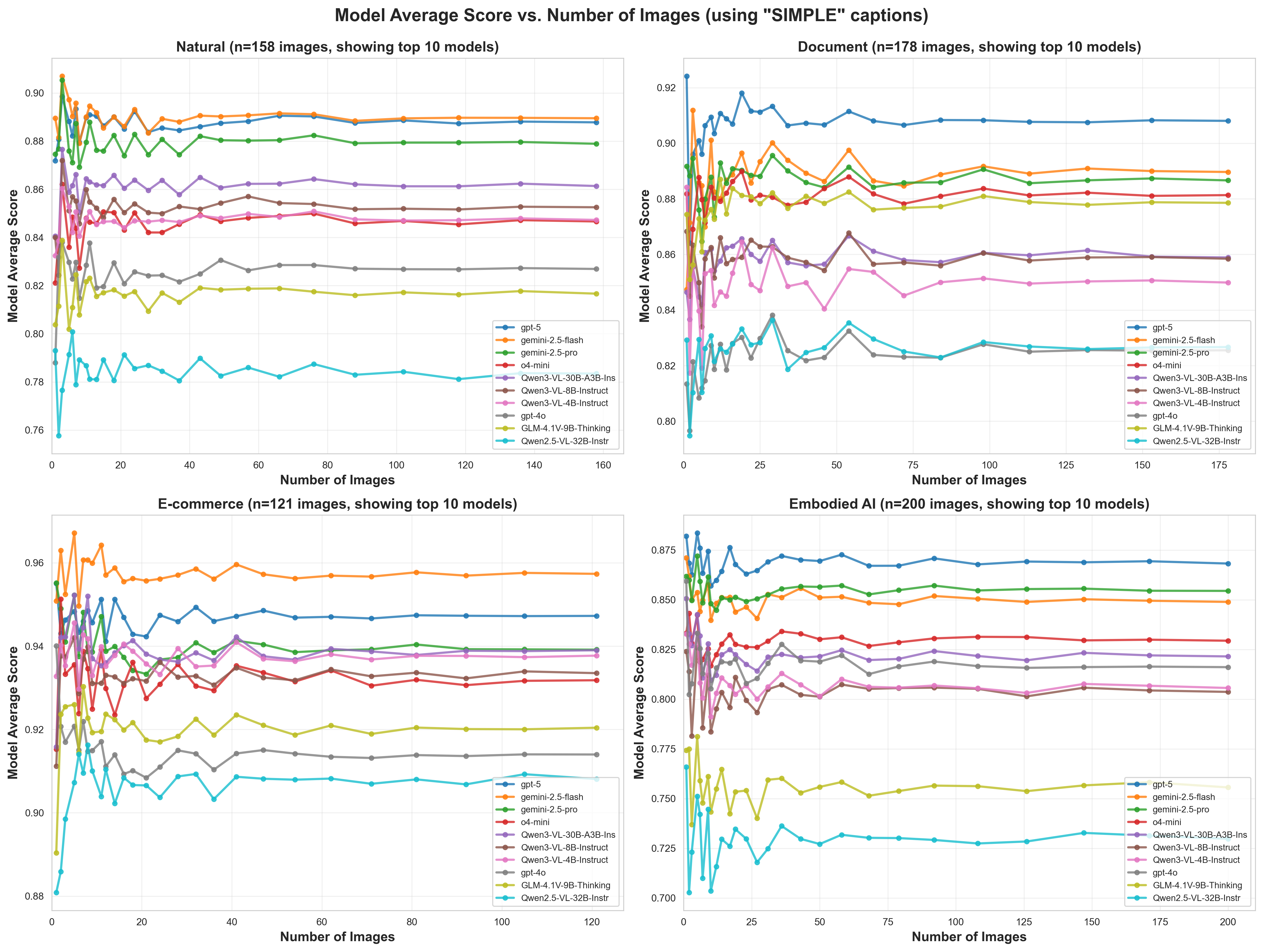}
\caption{\textbf{Model ranking stability vs. number of images (average score-based).}
Same analysis using average score (with partial credit for ``Cannot answer'': $1/n_{\text{choices}} + 0.05$). Patterns mirror Figure~\ref{fig:ranking_stability_accuracy}, confirming data sufficiency holds across evaluation metrics.}
\label{fig:ranking_stability_score}
\end{figure*}

Figures~\ref{fig:ranking_stability_accuracy} and~\ref{fig:ranking_stability_score} show performance trajectories for the top 10 models across all four domains. Three key findings emerge: (1)~\textbf{Rapid stabilization}: Rankings stabilize within 20-40 images ($\sim$10-20\% of full dataset) across all domains. (2)~\textbf{Stable ordering}: After initial stabilization, model rankings remain consistent---performance curves maintain their relative positions without crossing. (3)~\textbf{10$\times$ overcapacity}: Quantitative analysis shows rankings achieve Spearman correlation $\rho > 0.95$ with full rankings using only 10\% of images, indicating CaptionQA contains approximately 10$\times$ more data than necessary for reliable evaluation.

This analysis empirically validates our high-density design: CaptionQA's dense annotation strategy provides reliable model rankings while maintaining evaluation efficiency. Users can obtain $\rho > 0.95$ correlated rankings using only 30-50 images per domain for preliminary evaluation, confirming that semantic coverage through dense questions is more effective than scale through numerous loosely-annotated images.



\begin{table*}[t]
\centering
\small
\caption{Natural-domain caption-utility taxonomy in \textbf{CaptionQA}..}
\label{tab:taxonomy_natural}
\setlength{\tabcolsep}{6pt}
\begin{tabular}{p{0.20\linewidth} p{0.72\linewidth}}
\toprule
\textbf{Level-1 top-level category} & \textbf{Level-2 subcategories (examples in parentheses)} \\
\midrule

\multirow{2}{*}{Object Existence} 
  & Object presence \\
  & Use of named entities \\

\midrule
\multirow{8}{*}{Attribute}
  & Color \\
  & Shape \\
  & Size \\
  & Textual (written content on objects) \\
  & Material \\
  & State (e.g., open, closed, whole, broken) \\
  & Count / Quantity (singular, plural, exact count, range estimate) \\
  & Part–whole relationships (e.g., wheels of a car, branches of a tree) \\

\midrule
\multirow{9}{*}{Spatial}
  & Positional relationships (above, below, beside) \\
  & Orientation (facing forward, tilted) \\
  & Containment (inside, outside) \\
  & Attachment (connected, detached) \\
  & Distance between objects (near, far) \\
  & Clustering (grouped, scattered) \\
  & Scene composition (foreground, background) \\
  & Symmetry / asymmetry \\
  & Overlapping / occlusion \\

\midrule
\multirow{9}{*}{Action and Interaction}
  & Object–object interaction: contact (e.g., collision, gears rotation) \\
  & Object–object interaction: functional (e.g., key opening the door) \\
  & Object–human interaction: human activity (running, sitting) \\
  & Object–human interaction: interaction context (playing sports, using tools) \\
  & Human–human interaction: nonverbal communication (gestures, body language) \\
  & Human–human interaction: collaborative / social dynamics (teamwork, group interaction) \\
  & Human–human interaction: physical / affective interaction (hugs, handshakes) \\
  & Temporal state: indication of motion (blurred objects, motion trails) \\

\midrule
\multirow{6}{*}{Scene-Level Evaluation}
  & Environment / setting: environment type (natural, urban) \\
  & Environment / setting: location type (indoor, outdoor, semi-indoor, semi-outdoor) \\
  & Weather conditions (sunny, rainy, foggy) \\
  & Time of day (daytime, night, dusk) \\
  & Shadows and reflections \\
  & Light source directionality \\

\midrule
\multirow{7}{*}{Hallucination}
  & Object evaluation: object absence (likely but absent objects appropriately omitted) \\
  & Object evaluation: object ambiguity (e.g., wolf vs. husky) \\
  & Object evaluation: occluded objects misinterpretation \\
  & Scene evaluation: scene misinterpretation (snow-covered vs. white sand beach) \\
  & Scene evaluation: confounding elements (handles, reflections, shadows, clutter) \\
  & Action evaluation: implied actions (e.g., “running” when the person is stationary) \\

\bottomrule
\end{tabular}
\end{table*}


\begin{table*}[t]
\centering
\small
\caption{Document-domain caption-utility taxonomy in \textbf{CaptionQA}.}
\label{tab:taxonomy_document}
\setlength{\tabcolsep}{6pt}
\begin{tabular}{p{0.22\linewidth} p{0.70\linewidth}}
\toprule
\textbf{Level-1 top-level category} & \textbf{Level-2 subcategories (examples in parentheses)} \\
\midrule

\multirow{6}{*}{Structural Elements}
  & Layout: key structural elements (title, headers, footnotes, page number) \\
  & Layout: hierarchical structure (section, subsection) \\
  & Layout: columns (single-column, multi-column layout) \\
  & Spatial relationships: alignment (centered, left-aligned, right-aligned) \\
  & Spatial relationships: overlapping elements (text over images, legends over charts) \\
  & Spatial relationships: relative positioning (proximity of labels to corresponding elements) \\

\midrule
\multirow{9}{*}{Content-Level Evaluation}
  & Text-content: textual information (extracted text matches the image) \\
  & Text-content: completeness (missing text, partial text, missing values) \\
  & Text-content: formatting (bold, italic, underline, font size) \\
  & Text-content: style differentiation (captions vs.\ main body text) \\
  & Text-content: accurate recognition of numbers (percentages, decimal points) \\
  & Text-content: units and scales (e.g., 10 km vs.\ 10k) \\
  & Text-content: symbols and special characters (currency symbols, math symbols, emoji) \\
  & Visual elements: presence and identification of figures (charts, diagrams, icons) \\
  & Visual elements: metadata of figures (page number of figures, continuation of figures) \\

\midrule
\multirow{8}{*}{Chart-Specific Elements}
  & Chart types (bar chart, line chart, pie chart, scatter plot) \\
  & Axes and labels: presence of x-axis and y-axis labels \\
  & Axes and labels: axis scale (linear, logarithmic) \\
  & Axes and labels: units of measurement (time, percentage) \\
  & Legends \\
  & Data points: correctness of data points \\
  & Data points: completeness of data point descriptions \\
  & Data points: trend identification (upward trend, downward trend) \\

\midrule
\multirow{8}{*}{Table-Specific Elements}
  & Table structure: presence of headers (column headers, row headers) \\
  & Table structure: merged cells (multi-row cells, multi-column cells) \\
  & Table structure: gridlines (presence of borders, absence of borders) \\
  & Content: completeness of table content (missing cells, missing rows/columns) \\
  & Content: correctness of textual and numeric content in cells \\
  & Content: formatting (bold headers, colored cells for emphasis) \\
  & Content: units in cells (USD, kg) \\
  & Relationships: cross-references (footnotes or notes referring to specific cells) \\

\midrule
\multirow{5}{*}{Diagram-Specific Elements}
  & Types of diagrams (flowcharts, network diagrams, UML diagrams, Venn diagrams) \\
  & Components: nodes (shapes, labels) \\
  & Components: connections (nodes, type of connections, labels, symbols) \\
  & Relationships: directionality of connections (one-way, bidirectional) \\
  & Relationships: hierarchical structure (parent–child relationships in tree diagrams) \\

\midrule
\multirow{3}{*}{Domain-Specific Evaluation}
  & Domain-specific information: financial reports (key metrics such as revenue, profit) \\
  & Domain-specific information: scientific papers (recognition of equations, symbols) \\
  & Domain-specific information: legal documents (extraction of clauses, dates) \\

\bottomrule
\end{tabular}
\end{table*}

\begin{table*}[t]
\centering
\small
\caption{E-commerce-domain caption-utility taxonomy in \textbf{CaptionQA}.}
\label{tab:taxonomy_ecommerce}
\setlength{\tabcolsep}{6pt}
\begin{tabular}{p{0.25\linewidth} p{0.70\linewidth}}
\toprule
\textbf{Level-1 top-level category} & \textbf{Level-2 subcategories (examples in parentheses)} \\
\midrule

\multirow{9}{*}{Product-Level Information}
  & Product category (e.g., laptop, running shoes) \\
  & Product attributes: color (red, matte black) \\
  & Product attributes: dimensions (12 inches tall) \\
  & Product attributes: size (set of 3 mugs, a dozen) \\
  & Product attributes: material (leather, stainless steel) \\
  & Product attributes: shape (rectangular, oval) \\
  & Product attributes: texture (smooth finish, rough surface) \\
  & Product attributes: weight (lightweight backpack) \\
  & Product condition: new vs.\ used appearance (brand new, slightly worn); defects or wear (minor scratches) \\

\midrule
\multirow{7}{*}{Contextual and Scene Information}
  & Usage context: indoor vs.\ outdoor setting (outdoor garden furniture) \\
  & Usage context: lifestyle depiction (ideal for office use, perfect for outdoor camping) \\
  & Usage context: human interaction (worn by a model, held by a hand) \\
  & Usage context: environmental cues (in a kitchen, on a desk) \\
  & Completeness of scene: completeness of product display (full product shown, partially shown) \\
  & Completeness of scene: supporting objects (props, background elements) \\
  & Completeness of scene: scene cleanliness (cluttered, minimalistic background) \\

\midrule
\multirow{9}{*}{Visual Appearance and Presentation}
  & Image quality: blurriness \\
  & Image quality: lighting (bright, natural light, shadows) \\
  & Image quality: reflections (glare on metallic surfaces) \\
  & Background: plain background (white, black) \\
  & Background: styled background (lifestyle images with props) \\
  & Background: transparency (images with transparent backgrounds) \\
  & Perspective and angles: front / side / top / angled views \\
  & Perspective and angles: close-up shots of key features (zoomed-in details) \\
  & Perspective and angles: 360-degree or multi-angle views (implied by multiple images) \\

\midrule
\multirow{8}{*}{Functional Information}
  & Intended use: functionality (multi-purpose tool, designed for running) \\
  & Intended use: usability (easy to use, one-click operation) \\
  & Intended use: special features (waterproof, wireless connectivity) \\
  & Performance characteristics: capacity (16\,GB RAM, 1\,TB storage) \\
  & Performance characteristics: durability (scratch-resistant, long-lasting) \\
  & Performance characteristics: safety (child-safe material, non-toxic) \\
  & Compatibility: compatibility with other products (compatible with iOS and Android) \\
  & Compatibility: accessories included or sold separately (includes charging cable) \\

\midrule
\multirow{7}{*}{Brand and Marketing Information}
  & Branding: visible logo or brand name \\
  & Branding: trademark symbols (\textregistered, \texttrademark) \\
  & Branding: brand-specific design elements (signature patterns) \\
  & Branding: model / version identification (iPhone 14 Pro) \\
  & Promotional elements: sale indicators (50\% off tag) \\
  & Promotional elements: certifications or labels (FDA-approved, eco-friendly) \\
  & Promotional elements: awards and recognitions (Best product of the year) \\

\midrule
\multirow{4}{*}{Textual Elements in Image}
  & Embedded text recognition: product name or description \\
  & Embedded text recognition: price tags or discount labels \\
  & Embedded text recognition: usage instructions or warnings (handle with care) \\
  & Embedded text recognition: slogans or promotional text \\

\midrule
\multirow{5}{*}{Packaging and Accessories}
  & Packaging description: box, bag, or wrapper (premium gift box) \\
  & Packaging description: product labels on packaging (organic, recyclable) \\
  & Packaging description: packaging design (minimalistic, vintage style) \\
  & Accessories: listing of included accessories (comes with charger and earphones) \\
  & Accessories: identification of main product vs.\ accessories (distinguishing primary product from props) \\

\bottomrule
\end{tabular}
\end{table*}

\begin{table*}[t]
\centering
\small
\caption{Embodied-AI-domain caption-utility taxonomy in \textbf{CaptionQA}.}
\label{tab:taxonomy_embodied}
\setlength{\tabcolsep}{6pt}
\begin{tabular}{p{0.28\linewidth} p{0.65\linewidth}}
\toprule
\textbf{Level-1 top-level category} & \textbf{Level-2 subcategories (examples in parentheses)} \\
\midrule

\multirow{15}{*}{Perception}
  & Object recognition: object identification (e.g., cup, door handle) \\
  & Object recognition: object category (e.g., furniture, tools, appliances) \\
  & Attribute: color \\
  & Attribute: shape \\
  & Attribute: size \\
  & Attribute: material \\
  & Attribute: state \\
  & Attribute: orientation \\
  & Relationships: positional relationships (on top of, next to) \\
  & Relationships: containment (object inside a box) \\
  & Relationships: attachment (tool attached to a robotic arm) \\
  & Relationships: occlusion (partially visible object behind another object) \\
  & Interaction: contact (robot holding a bottle) \\
  & Interaction: manipulation (grasping, pushing, pulling) \\
  & Interaction: proximity (object within reach, object far from reach) \\

\midrule
\multirow{7}{*}{Spatial and Environment Context}
  & Spatial: proximity (near the edge of the table) \\
  & Spatial: distance estimation (approximately 2 meters away) \\
  & Spatial: perspective (view from a high angle, low-angle view) \\
  & Environmental description: indoor vs.\ outdoor (indoor kitchen, outdoor garden) \\
  & Environmental description: room type (living room, office, workshop) \\
  & Environmental description: surroundings (surrounded by shelves, in an open space) \\
  & Environmental description: surface properties (wooden floor, metal surface) \\

\midrule
\multirow{10}{*}{Activity and Task Context}
  & Task identification: navigation tasks (robot navigating a hallway) \\
  & Task identification: object manipulation tasks (picking up a tool) \\
  & Task identification: cleaning tasks (sweeping debris) \\
  & Task identification: inspection tasks (inspecting a pipe for damage) \\
  & Implied actions: action in progress (robot approaching a table) \\
  & Implied actions: action completed (door successfully opened) \\
  & Implied actions: task outcome (object successfully placed in the bin) \\
  & Human–robot interaction: human presence (person standing nearby) \\
  & Human–robot interaction: interaction type (handing an object to the robot) \\
  & Human–robot interaction: collaborative actions (robot assisting a person with a task) \\

\midrule
\multirow{5}{*}{Scene Dynamics}
  & Motion and kinematics: robot motion (robot moving forward, robot arm rotating) \\
  & Motion and kinematics: object motion (ball rolling on the floor) \\
  & Motion and kinematics: velocity estimation (object moving quickly, slow movement) \\
  & Temporal information: time-specific context (morning light coming through the window) \\
  & Temporal information: sequential actions (after opening the drawer, picking up the tool) \\

\midrule
\multirow{7}{*}{Sensor and Embodiment Information}
  & Sensor-specific features: camera type (RGB, depth, thermal) \\
  & Sensor-specific features: depth perception (distance to object measured by depth sensor) \\
  & Sensor-specific features: field of view (wide-angle view, narrow focus on object) \\
  & Sensor-specific features: sensor artifacts (glare on metallic surface, low-light noise) \\
  & Robot embodiment: robot components in frame (robot arm, gripper) \\
  & Robot embodiment: self-awareness (robot's shadow visible, robot base in view) \\
  & Robot embodiment: tool attachment (screwdriver attached to gripper) \\

\midrule
\multirow{7}{*}{Functional and Semantic Understanding}
  & Affordance recognition: affordances of objects (graspable handle, pourable bottle) \\
  & Affordance recognition: tool usability (wrench ready to tighten a bolt) \\
  & Affordance recognition: interaction potential (button pressable by robot finger) \\
  & Semantic completeness: completeness of scene description (all key objects and actions described) \\
  & Semantic completeness: avoidance of hallucination (no mention of non-existent objects or actions) \\
  & Contextual relevance: task-specific relevance (focusing on objects necessary for the task) \\
  & Contextual relevance: importance weighting (emphasizing key objects over background elements) \\

\bottomrule
\end{tabular}
\end{table*}

\section{Cost of Extending CaptionQA to New Domains}
\label{sec:cost-new-domains}

One of our design goals is that CaptionQA should be easy to extend beyond the four domains used in the main paper (Natural, Document, E-commerce, Embodied AI). In this section we clarify what needs to be done to add a new domain and how the computational cost scales in our reference implementation.

\paragraph{Practical steps for adding a new domain.}
Given a new domain $\mathcal{D}_{\text{new}}$, extending CaptionQA is a purely mechanical procedure driven by our released code:
\begin{enumerate}
    \item \textbf{Write a taxonomy.} Define a domain-specific taxonomy of information needs (analogous to Table 1-4 in the Supplementary), specifying which aspects (objects, layout, OCR, affordances, etc.) are important for downstream applications in $\mathcal{D}_{\text{new}}$. This is a one-time configuration file.
    \item \textbf{Collect images.} Curate a set of (around 100-150 images to balance evaluation time and thoroughness) $N_{\text{img}}^{\text{new}}$ representative images for $\mathcal{D}_{\text{new}}$.
    \item \textbf{Run the pipeline.} Invoke our end-to-end scripts, which automatically
    (i) generate taxonomy-grounded multiple-choice questions with three VLM agents (GPT 4o, o4-mini, GPT-5),
    (ii) apply the Qwen-based text-only filter and Qwen3 embedding deduplication, and
    (iii) perform dual-VLM visual verification before optional human spot-checking.
\end{enumerate}
No manual question authoring is required: once the taxonomy and image list are prepared, all remaining steps are controlled by code scripts.

\paragraph{API cost.}
For a new domain at the same scale as our Natural split (around $N_{\text{img}}^{\text{new}}\!\approx\!150$ images and $\sim 10{,}000$ final QA pairs), the main API usage comes from (i) taxonomy-guided question generation with three VLM agents (GPT-4o, o4-mini, GPT-5), and (ii) dual-VLM quality control with GPT-5 and Gemini 2.5 Pro. In our current implementation this corresponds to roughly $17.7$M total tokens across these APIs. Using current list prices for these models, this amounts to about \textbf{\$40–\$60} of one-time API cost to construct a new domain with $\sim 10$k finalized questions. Since this cost scales linearly with the number of images and the desired question density, smaller/larger domains incur proportionally smaller/larger API budgets.

\paragraph{GPU cost.}
All Qwen models in the pipeline (Qwen2.5-72B for text-only filtering and QA evaluation, and Qwen3-Embedding for deduplication) are run locally on a single AMD Instinct MI325. For a new domain with $\sim 10$k final QA pairs, and assuming a modest candidate-to-final ratio (e.g., $\sim 3$ candidates per retained question and 10 blind QA passes per candidate during text-only filtering), the total number of Qwen calls corresponds to roughly \textbf{3–4 MI325 GPU-hours} for construction. Once the domain is built, evaluating one captioning model on that domain (QA-on-caption scoring) takes less than \textbf{8 minutes} of MI325 time, with no additional API cost since Qwen runs locally. Thus, both construction-time and evaluation-time GPU costs are low and scale linearly with the number of questions.

\paragraph{Human cost.}
Human effort is deliberately kept minimal. Adding a new domain typically requires: (i) \textbf{2–4 hours} of expert time to design and refine the domain taxonomy (often by adapting and editing existing taxonomies), (ii) a short image curation pass to collect around 150 representative images, and (iii) \textbf{3–5 hours} of manual checking of questions that are flagged by the dual-VLM checker as potentially ambiguous, ungrounded, or too reasoning-heavy. In practice, this corresponds to roughly \textbf{one expert-day ($\sim$6–8 hours)} of human work per new domain, far less than what would be needed to author tens of thousands of multiple-choice questions from scratch.

\paragraph{Takeaway.}
The main conclusion is that extending CaptionQA to a new domain is entirely feasible:
once a taxonomy and image set are specified, the rest of the process reduces to running our public code on a single GPU.
The computational cost scales linearly with $N_{\text{img}}^{\text{new}}$ and $N_{\text{q}}$ and is dominated by Qwen inference on one AMD MI325,
placing extensions with a few hundred images per domain well within reach for typical academic and industrial users.

\section{Rational and Reliability of LLM as QA Reader}
Modern industrial systems increasingly rely on large language models not only as standalone chatbots, but as \emph{components} inside downstream pipelines: LLM-based embedding models for retrieval and recommendation, LLM-driven re-ranking in search and feeds, and LLM agentic pipelines that orchestrate tools and multi-step plans. In many of these settings, the LLM consumes text surrogates of visual content (captions, alt-text, OCR transcripts) rather than raw pixels. In other words, a caption is often fed directly into an LLM that must then make a decision, retrieve items, or answer questions. From this perspective, using an LLM to \emph{read} captions in CaptionQA is not a toy setup, but a practical abstraction of how captions are actually used in modern systems.

\paragraph{Different from generic ``LLM-as-a-judge'' settings.}
Our use of an LLM QA reader is conceptually different from generic ``LLM-as-a-judge'' approaches that ask a model to rate a caption on heuristic criteria such as fluency, correctness, or level of detail. In CaptionQA, the LLM is not asked to produce a direct quality score; instead, it is placed in a concrete downstream task: answer multiple-choice questions about an image \emph{using only the caption as input}. The LLM has no access to the image and must treat the caption as its sole evidence. This makes the LLM a stand-in for a real downstream consumer that must act based on textual surrogates, rather than a meta-critic with access to special instructions or reference answers. The quantity we measure is therefore not ``how good the caption looks to the LLM'', but \emph{whether the caption actually supports successful task completion when an LLM tries to use it.}

\paragraph{Why QA, and why an LLM QA reader?}
We deliberately choose QA as the interaction between captions and the downstream LLM for two reasons. First, QA is a natural interface for many real applications: agents answering user questions about documents or products, assistants reasoning over screenshots or forms, and recommender systems extracting attributes from item descriptions. In these settings, the LLM rarely ``rates'' captions in the abstract; instead, it must use captions to answer concrete questions such as \emph{``What color is the dress?'', ``Is the document signed?'', ``Which button should the robot press?''}. Our QA items are constructed exactly in this style. For example, in the E-commerce domain, we build a taxonomy around the information that real product recommendation and ranking systems need (category, style, material, pattern, fit, defects, packaging, etc.), and then generate multiple-choice questions that directly query these attributes. Similarly, the Document taxonomy targets fields and layout that drive automation pipelines, and the Embodied AI taxonomy focuses on object states and affordances relevant for agents. As a result, CaptionQA questions are not generic trivia, but explicit probes of domain-specific information needs.

Second, QA is a task where strong LLMs already exhibit high absolute performance and remarkable \emph{stability}. In our experiments (Section~3.5), the QA reader attains high accuracy on image-grounded QA and shows low variance under option shuffling and repeated sampling. This stability is crucial: if the QA model itself were noisy, it would be unclear whether errors come from the caption or from the reader. Our analysis shows that, once the reader is strong enough, the dominant source of failures is the information available in the caption, not randomness in the LLM. In this sense, QA-on-caption with an LLM reader is a practical proxy for downstream usage: it asks whether a caption contains the specific, taxonomy-grounded facts that an LLM-based system (embedding model, recommender, or agent) will actually need in order to make correct decisions.

\paragraph{Our assumption: measure utility from the downstream LLM's perspective.}
CaptionQA is built around a simple assumption: what ultimately matters is what a \emph{downstream LLM} can recover from the caption. If a caption causes the QA reader to answer incorrectly or become confused, this is evidence that the caption is missing or misleading with respect to that aspect of the image. If the reader still guesses the answer correctly by relying on world knowledge or priors, this also reflects something about caption utility: the caption may be underspecified (the LLM is filling in gaps) or the question may be solvable without the image. In both cases, we treat the combination of caption and QA reader as a black-box downstream system and measure whether it succeeds. This aligns more closely with real deployments, where practitioners care about \emph{end-to-end decisions} made from captions, not about any particular intrinsic caption metric.

\paragraph{Independence from a specific QA model.}
Our framework is agnostic to the choice of QA reader: any sufficiently strong LLM can in principle be plugged into the CaptionQA evaluation protocol. In the main experiments we adopt Qwen2.5-72B as the default reader, but the benchmark does not depend on this particular model. Swapping the QA reader corresponds to changing the downstream consumer---e.g., using a different embedding model, recommender, or agent backend---while keeping the questions and captions fixed. This makes CaptionQA a flexible tool: users who rely on different LLM stacks in practice can re-run QA-on-caption with their own reader, and directly assess how well a caption model serves \emph{their} downstream LLM-based system.

\section{Prompt Transition Analysis: Where Does Length Help?}
\label{sec:appendix_coverage}

We analyze accuracy changes across four prompt transitions to identify which categories benefit from longer or more structured prompts.

\subsection{Short to Simple: Identifying High-ROI vs. Low-ROI Categories}

\begin{figure}[htb!]
\centering
\includegraphics[width=\columnwidth]{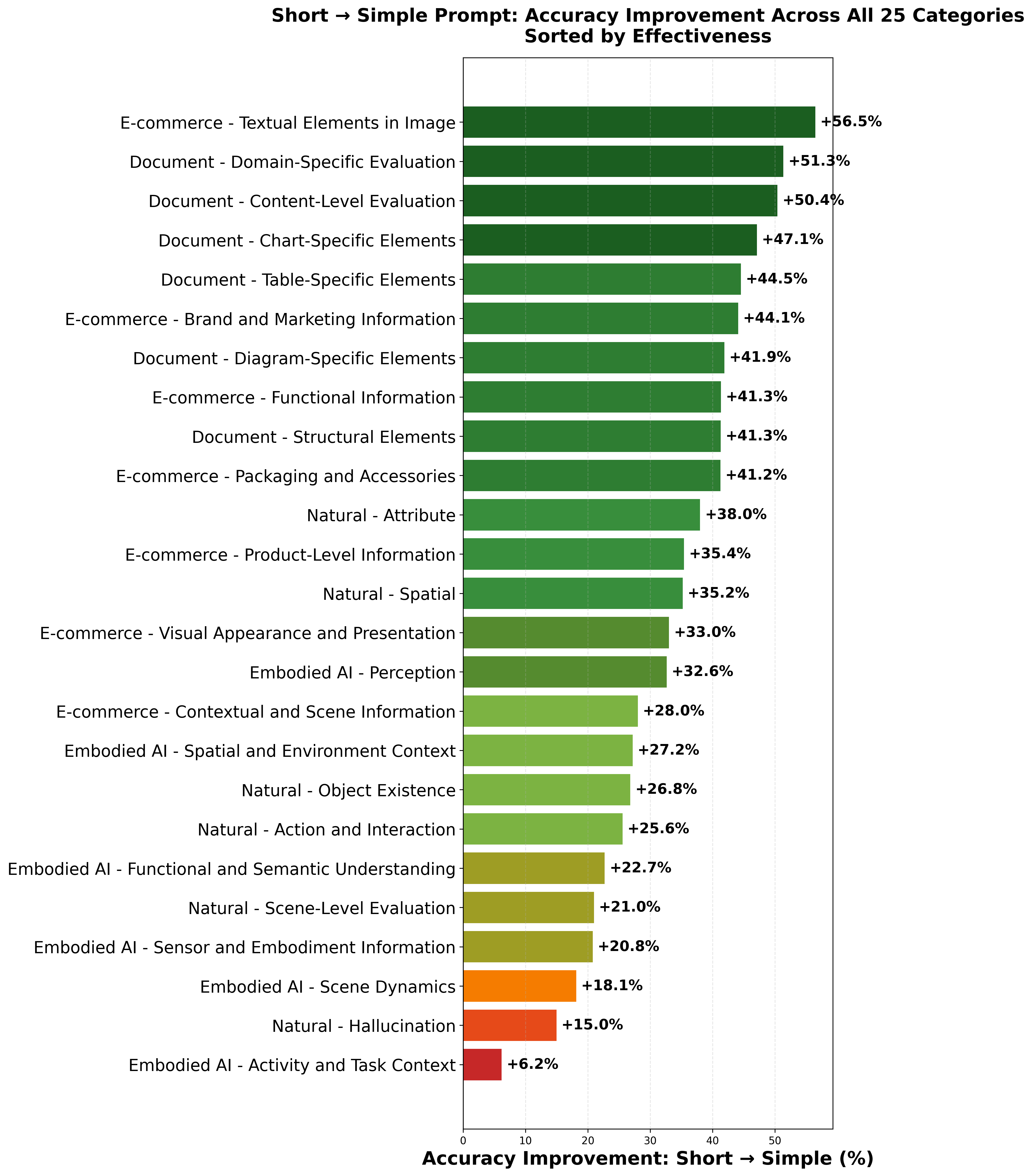}
\caption{\textbf{Short to Simple effectiveness across all 25 categories.} Categories sorted by improvement, color-coded from dark green (+47-56\%) to red (+6-21\%). Top categories: Document domain-specific evaluation and E-commerce textual elements. Bottom categories: Embodied AI activity context and Natural hallucination. Top categories gain up to 9$\times$ more than bottom categories.}
\label{fig:transition_short_simple}
\end{figure}

Figure~\ref{fig:transition_short_simple} shows all 25 categories sorted by improvement. Document domain-specific evaluation (+47-51\%) and E-commerce textual elements (+44-56\%) gain the most. Embodied AI categories (+6-33\%) and Natural hallucination (+15\%) gain the least. Top categories gain up to 9$\times$ more than bottom categories, showing clear domain-specific patterns. The Short to Simple transition captures the majority of gains (mean +33.8\%), while Simple to Long adds minimal value (mean +0.4\%, shown next).

\subsection{Simple to Long: Diminishing Marginal Returns}

\begin{figure}[htb!]
\centering
\includegraphics[width=\columnwidth]{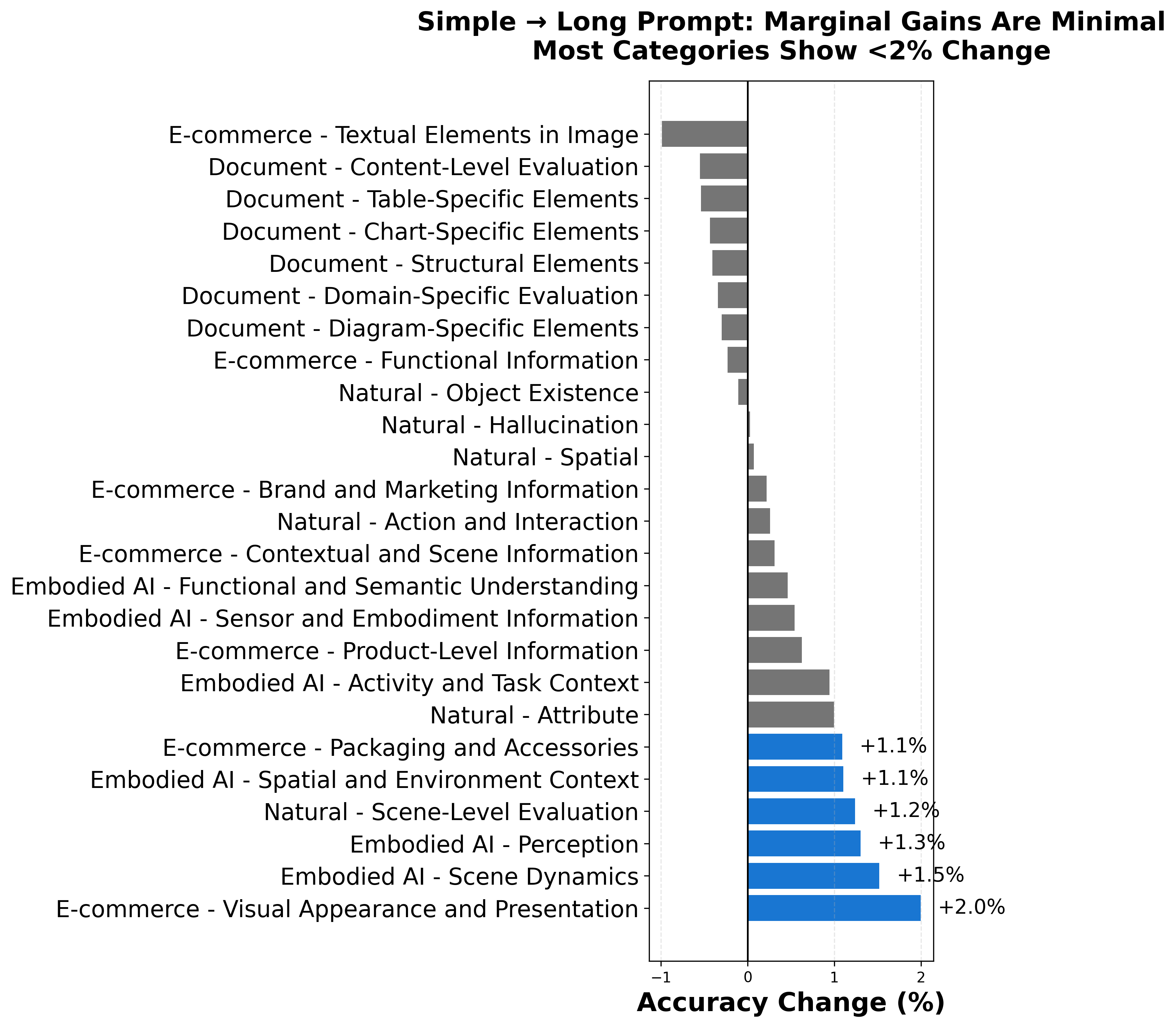}
\caption{\textbf{Marginal gains: Simple to Long.} All 25 categories show $<$2\% change (mean +0.35\%). Simple and Long achieve nearly identical scores (75.5\% vs 75.7\%) despite 1.4$\times$ length difference (355 vs 510 words). Additional length provides minimal gain.}
\label{fig:transition_simple_long}
\end{figure}

Figure~\ref{fig:transition_simple_long} shows that Simple to Long transitions are nearly flat. All categories show $<$2\% change, mean +0.35\%. Most gains occur in the Short to Simple transition (+33.8\% mean), while Simple to Long adds little value. Simple prompts achieve 99\% of Long's performance at 70\% of the length.

\subsection{Long to Taxonomy-Hinted: When Structure Backfires}

\begin{figure}[htb!]
\centering
\includegraphics[width=\columnwidth]{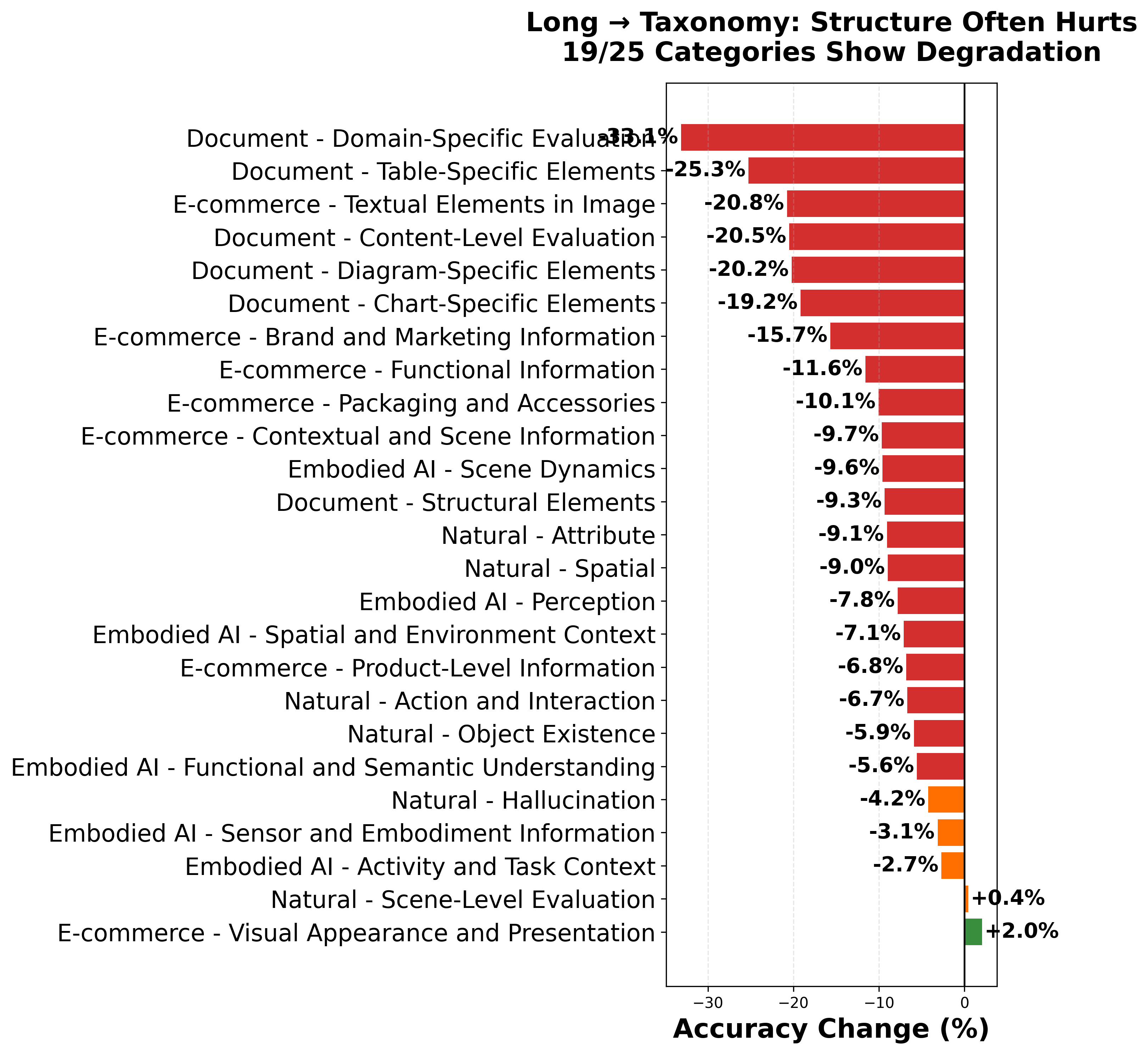}
\caption{\textbf{Taxonomy-Hinted prompts often degrade performance.} 23 of 25 categories show losses (mean -10.8\%), with 20 losing $>$5\%. Only 2 categories gain (Visual Appearance +2.0\%, Scene-Level Evaluation +0.4\%). Largest losses: Document Domain-Specific Evaluation (-33.1\%), Embodied AI Perception (-7.8\%), Document Structural Elements (-9.3\%).}
\label{fig:transition_long_taxonomy}
\end{figure}

Figure~\ref{fig:transition_long_taxonomy} shows that Taxonomy-Hinted prompts (which list all 69 subcategories) hurt 23/25 categories vs Long, with mean loss of -10.8\%. Largest losses occur in categories that already struggle: Document Domain-Specific Evaluation (-33.1\%), Embodied AI Perception (-7.8\%), Document Structural Elements (-9.3\%). When models cannot extract information, explicit category lists may pressure them to fabricate details.

The 2 categories that benefit from Taxonomy-Hinted--Scene-Level Evaluation (Natural) and Visual Appearance (E-commerce)--are both high-level judgments where structure may help organize outputs. Taxonomy-Hinted prompts may work for conceptual categories but not for fine-grained perceptual ones.

\subsection{Coverage-Accuracy Relationship Across Prompt Transitions}

We examine the relationship between \emph{coverage} (Cannot-Answer rate reduction) and \emph{accuracy} improvement across all four prompt transitions. Figures~\ref{fig:coverage_short_to_long}--\ref{fig:coverage_long_to_taxonomy} reveal distinct patterns for each transition.

\subsubsection{Short to Long: Strong Positive Correlation}

\begin{figure}[htb!]
\centering
\includegraphics[width=\columnwidth]{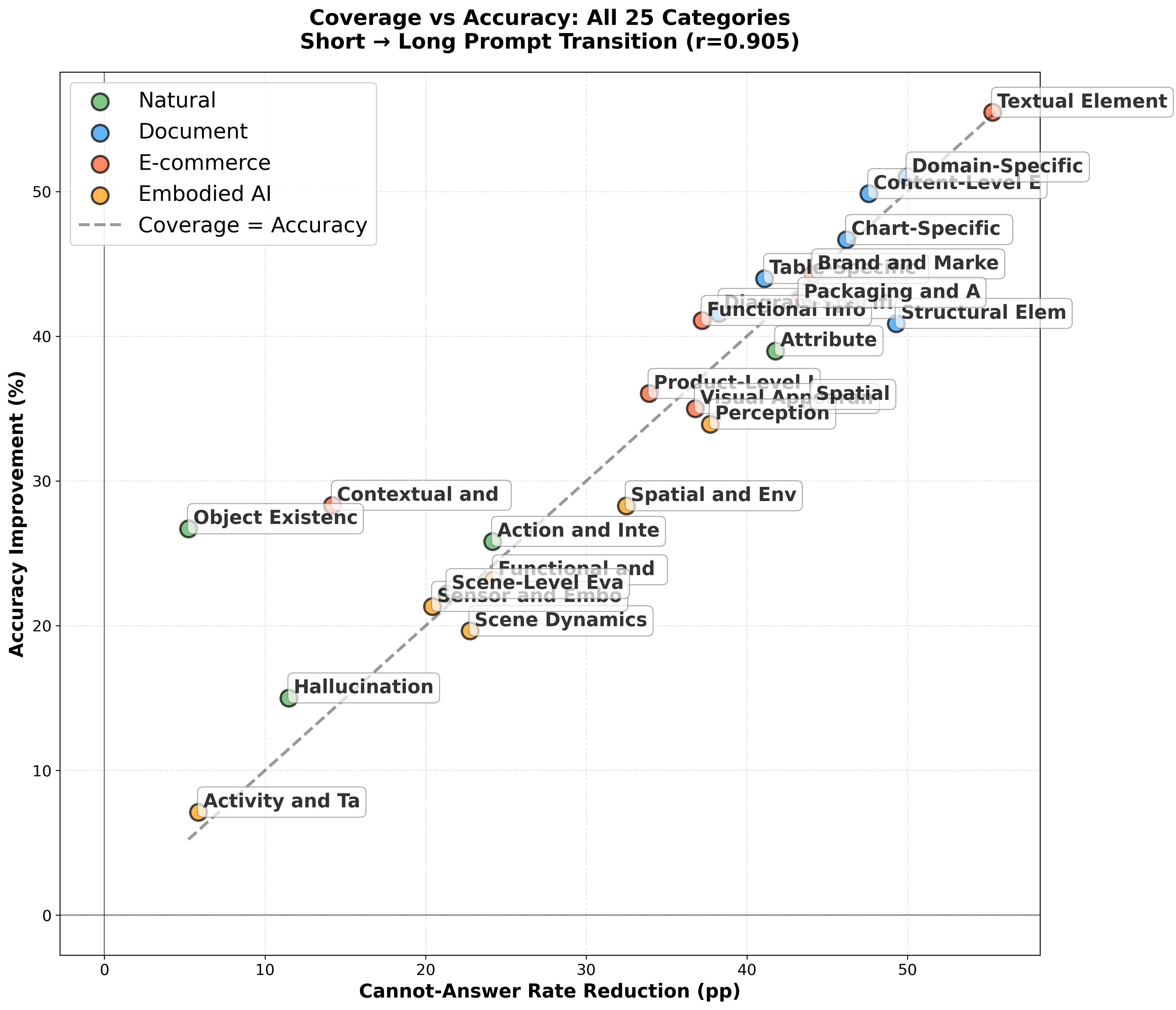}
\caption{\textbf{Coverage vs. Accuracy: Short to Long (r=0.905).} Most categories cluster near diagonal. Below diagonal: Natural Spatial (43.9\% coverage vs 35.3\% accuracy), Document Structural (49.3\% vs 40.9\%)—more coverage than accuracy. Above diagonal: Natural Object Existence (5.3\% coverage vs 26.7\% accuracy), E-commerce Contextual (14.4\% vs 28.3\%)—more accuracy than coverage.}
\label{fig:coverage_short_to_long}
\end{figure}

Figure~\ref{fig:coverage_short_to_long} shows strong correlation (r=0.905) between coverage and accuracy—longer captions generally improve both. \textbf{Below diagonal} (coverage $>$ accuracy): Long captions answer many more questions, but newly answerable questions have lower accuracy (79-84\%). Example: Natural Spatial gains 44\% coverage but only 35\% accuracy--adding substantial content but some is wrong. \textbf{Above diagonal} (accuracy $>$ coverage): More efficient captioning that improves accuracy on already-answerable questions without adding much new content. Example: Natural Object Existence gains 27\% accuracy but only 5\% coverage--fixing errors rather than expanding scope.

\subsubsection{Short to Simple: Captures Most of Long's Benefits}

\begin{figure}[htb!]
\centering
\includegraphics[width=\columnwidth]{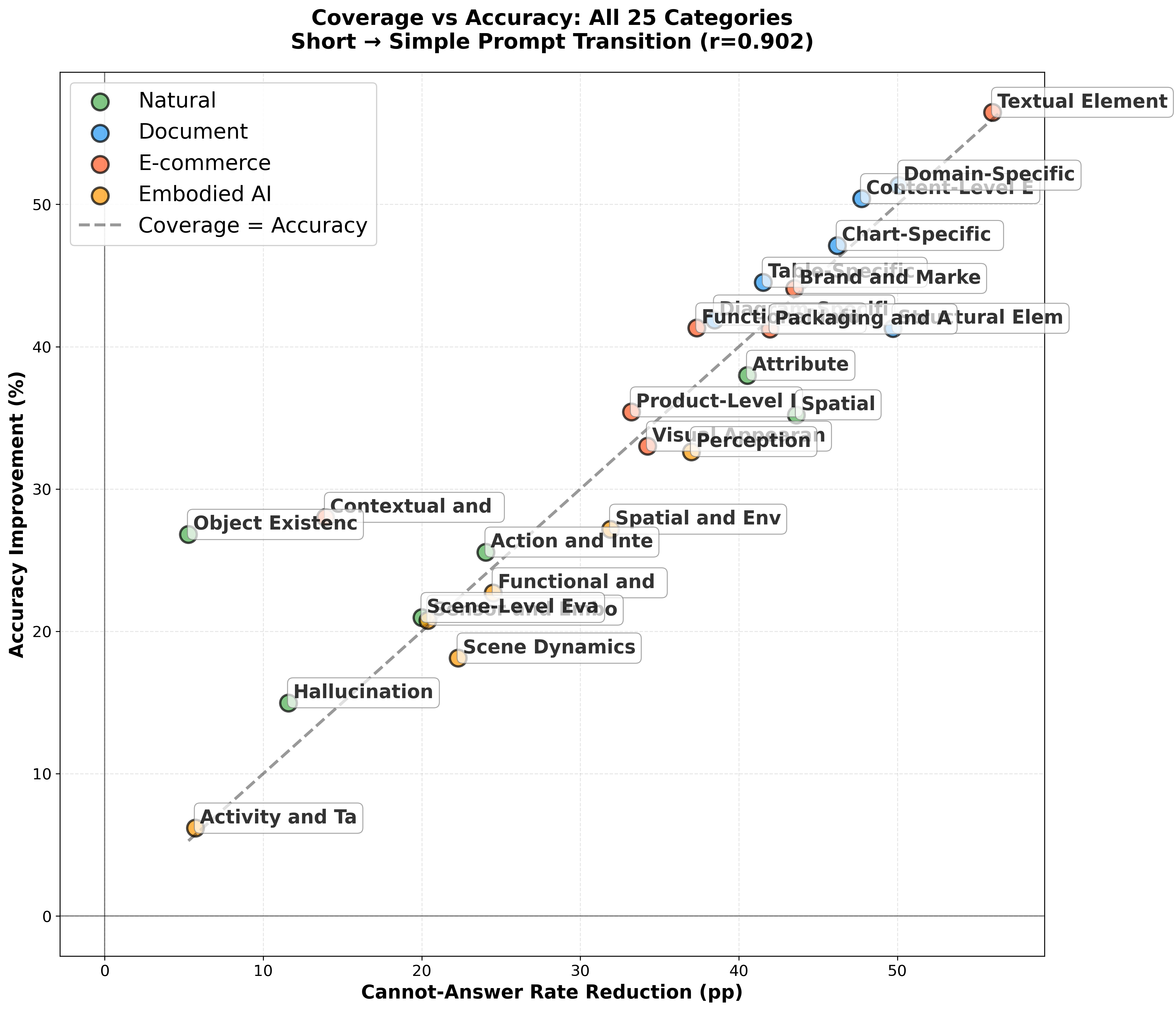}
\caption{\textbf{Coverage vs. Accuracy: Short to Simple (r=0.902).} Mean coverage gain 32.8\%, mean accuracy gain 33.8\%. Comparing to Short to Long (33.1\% coverage, 34.2\% accuracy), this transition achieves 99\% of Long's gains at 70\% of the length.}
\label{fig:coverage_short_to_simple}
\end{figure}

Figure~\ref{fig:coverage_short_to_simple} shows r=0.902 with mean gains of 32.8\% coverage and 33.8\% accuracy. Comparing to Short to Long (r=0.905, 33.1\% coverage, 34.2\% accuracy), Simple achieves 99\% of Long's benefits. This means the additional Simple to Long step adds minimal value (see next).

\subsubsection{Simple to Long: Minimal Changes}

\begin{figure}[htb!]
\centering
\includegraphics[width=\columnwidth]{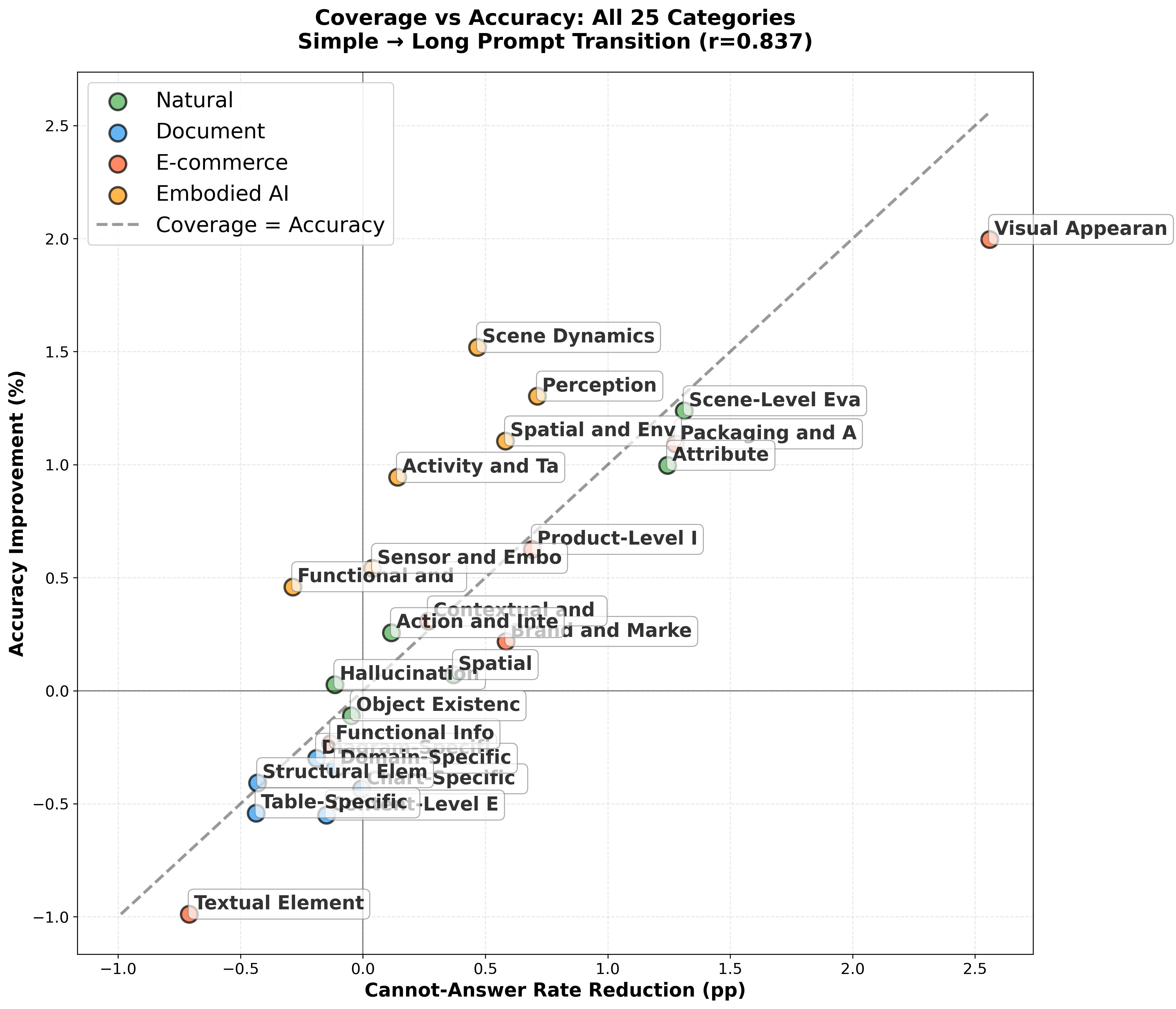}
\caption{\textbf{Coverage vs. Accuracy: Simple to Long (r=0.837).} Clustered near origin. Mean coverage change +0.3\%, mean accuracy change +0.4\%. E-commerce Visual Appearance is outlier with +2.6\% coverage and +2.0\% accuracy. Near-zero changes confirm diminishing returns.}
\label{fig:coverage_simple_to_long}
\end{figure}

Figure~\ref{fig:coverage_simple_to_long} shows minimal changes. Most categories cluster near (0,0), confirming Simple prompts achieve nearly all benefits of Long prompts. Both coverage and accuracy changes are $<$2\% for all categories, showing the Simple to Long step adds negligible value. E-commerce Visual Appearance is the only notable outlier.

\subsubsection{Long to Taxonomy-Hinted: Negative Correlation}

\begin{figure}[htb!]
\centering
\includegraphics[width=\columnwidth]{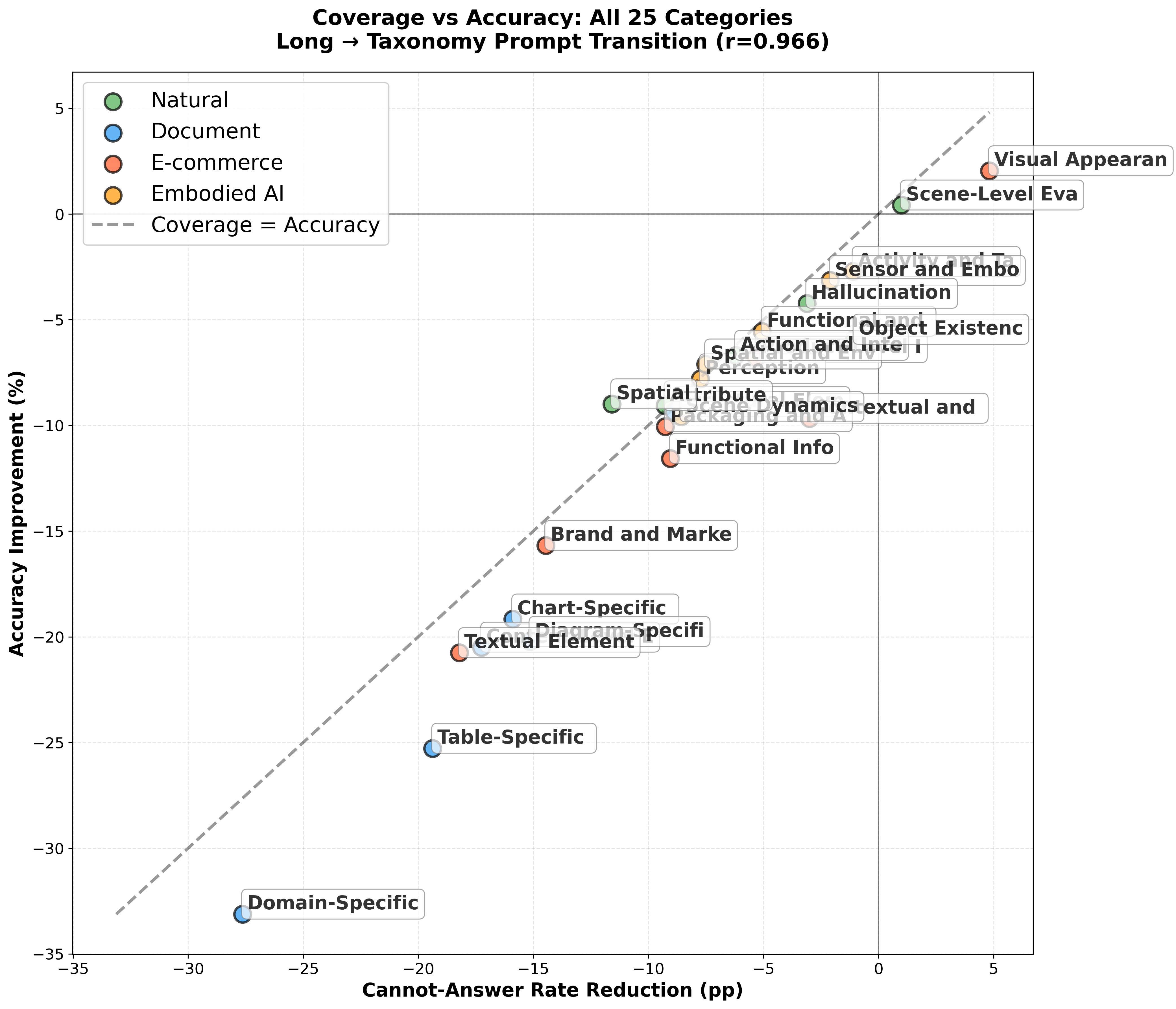}
\caption{\textbf{Coverage vs. Accuracy: Long to Taxonomy-Hinted (r=0.966).} Strong negative correlation. Mean coverage change -8.8\%, mean accuracy change -10.8\%. Document Domain-Specific is worst outlier (-27.6\% coverage, -33.1\% accuracy). Only 2 categories show gains. Bottom-left quadrant: Taxonomy-Hinted prompts add wrong content that reduces both coverage and accuracy.}
\label{fig:coverage_long_to_taxonomy}
\end{figure}

Figure~\ref{fig:coverage_long_to_taxonomy} shows both coverage and accuracy decrease. Taxonomy-Hinted prompts harm both metrics for 23/25 categories. Strong correlation (r=0.966) shows coverage and accuracy decline together. Structured prompts that force ungroundable content cause captioning models to add fabricated details, which are then detected as unreliable by QA models (increasing Cannot-Answer) while also reducing accuracy on answerable questions.

\subsection{Extreme Cases: Information Omission in Short Captions}

Categories where Short captions yield Cannot-Answer rates exceeding 70\%:
\begin{itemize}[leftmargin=*,noitemsep]
\item \textbf{Perception (Embodied AI)}: 73.7\% cannot-answer. Short captions omit fine-grained object attributes for manipulation tasks.
\item \textbf{Spatial (Natural)}: 74.2\% cannot-answer. Short captions exclude precise spatial relationships (distances, orientations).
\item \textbf{Textual Elements (E-commerce)}: 72.1\% cannot-answer. Short captions rarely transcribe product text (labels, specifications).
\end{itemize}

High Cannot-Answer rates indicate information is missing from the caption, not from the image. This is correct QA behavior—answering would require guessing.

\section{Category-Level Statistical Summary}
\label{sec:appendix_category_stats}

Table~\ref{tab:category_summary} shows statistics for all 25 top-level categories under the Simple prompt, aggregated across all models. For each category: mean score, standard deviation (across models), minimum and maximum scores (model variance), Cannot-Answer rate, and question count.

\begin{table*}[t]
\centering
\small
\caption{Per-category performance summary across all models (Simple prompt). Categories are sorted by mean score within each domain. High std indicates large inter-model variance; high Cannot rate indicates systematic omissions.}
\label{tab:category_summary}
\setlength{\tabcolsep}{4pt}
\begin{tabular}{llcccccc}
\toprule
\textbf{Domain} & \textbf{Category} & \textbf{Mean} & \textbf{Std} & \textbf{Min} & \textbf{Max} & \textbf{Cannot \%} & \textbf{\#Q} \\
\midrule
\multirow{6}{*}{Natural}
  & Scene-Level Evaluation & 82.3 & 8.1 & 65.2 & 91.4 & 14.7 & 1247 \\
  & Object Existence & 77.7 & 9.3 & 58.1 & 89.2 & 4.3 & 2215 \\
  & Action and Interaction & 76.9 & 8.7 & 61.3 & 88.5 & 22.4 & 1682 \\
  & Attribute & 69.0 & 10.2 & 48.9 & 84.7 & 32.5 & 3178 \\
  & Hallucination & 85.3 & 7.4 & 71.2 & 95.1 & 10.0 & 1913 \\
  & Spatial & 62.8 & 11.6 & 42.1 & 78.9 & 30.6 & 1210 \\
\midrule
\multirow{6}{*}{Document}
  & Domain-Specific Evaluation & 80.8 & 12.3 & 58.9 & 93.2 & 18.9 & 982 \\
  & Content-Level Evaluation & 79.0 & 11.8 & 59.3 & 91.7 & 20.6 & 2448 \\
  & Diagram-Specific Elements & 77.6 & 10.9 & 58.2 & 90.1 & 17.8 & 467 \\
  & Chart-Specific Elements & 75.5 & 10.5 & 57.9 & 88.3 & 21.0 & 891 \\
  & Table-Specific Elements & 74.6 & 10.2 & 56.8 & 87.4 & 17.3 & 1124 \\
  & Structural Elements & 64.7 & 9.8 & 47.3 & 79.2 & 23.7 & 1510 \\
\midrule
\multirow{7}{*}{E-commerce}
  & Contextual and Scene Info & 86.5 & 8.9 & 71.2 & 95.8 & 6.1 & 1041 \\
  & Product-Level Information & 85.4 & 9.2 & 69.8 & 94.3 & 17.4 & 1125 \\
  & Textual Elements in Image & 83.5 & 11.7 & 61.9 & 95.2 & 17.7 & 687 \\
  & Brand and Marketing Info & 83.4 & 10.1 & 66.7 & 93.5 & 20.8 & 1037 \\
  & Functional Information & 86.3 & 9.4 & 70.1 & 95.9 & 16.0 & 589 \\
  & Packaging and Accessories & 78.8 & 10.6 & 61.2 & 91.3 & 23.9 & 682 \\
  & Visual Appearance & 71.8 & 9.8 & 55.4 & 85.7 & 22.6 & 725 \\
\midrule
\multirow{6}{*}{Embodied AI}
  & Activity and Task Context & 82.5 & 9.7 & 66.8 & 93.2 & 18.1 & 1689 \\
  & Perception & 65.3 & 11.4 & 45.9 & 81.7 & 36.7 & 4294 \\
  & Sensor and Embodiment Info & 70.4 & 10.8 & 52.6 & 85.1 & 27.8 & 741 \\
  & Functional/Semantic Underst. & 78.4 & 10.2 & 61.7 & 90.3 & 23.9 & 1247 \\
  & Scene Dynamics & 70.1 & 11.6 & 51.3 & 86.4 & 24.1 & 478 \\
  & Spatial and Environment & 71.8 & 10.9 & 54.2 & 87.6 & 25.9 & 825 \\
\bottomrule
\end{tabular}
\end{table*}

Several insights emerge from Table~\ref{tab:category_summary}: (1) \textbf{Hallucination detection is robust} (85.3\% mean, 10.0\% Cannot), suggesting models effectively avoid generating false claims; (2) \textbf{Spatial reasoning shows high variance} (std=11.6 for Natural Spatial, 10.9 for Embodied Spatial), indicating inconsistent grounding across models; (3) \textbf{Cannot-Answer rates correlate with difficulty} (r=0.68), but not perfectly—Perception (Embodied AI) has both low score (65.3\%) AND high Cannot rate (36.7\%), indicating pervasive information absence.

\subsection{Domain-Specific Subcategory Analysis}

\begin{figure*}[htb!]
\centering
\includegraphics[width=0.95\textwidth]{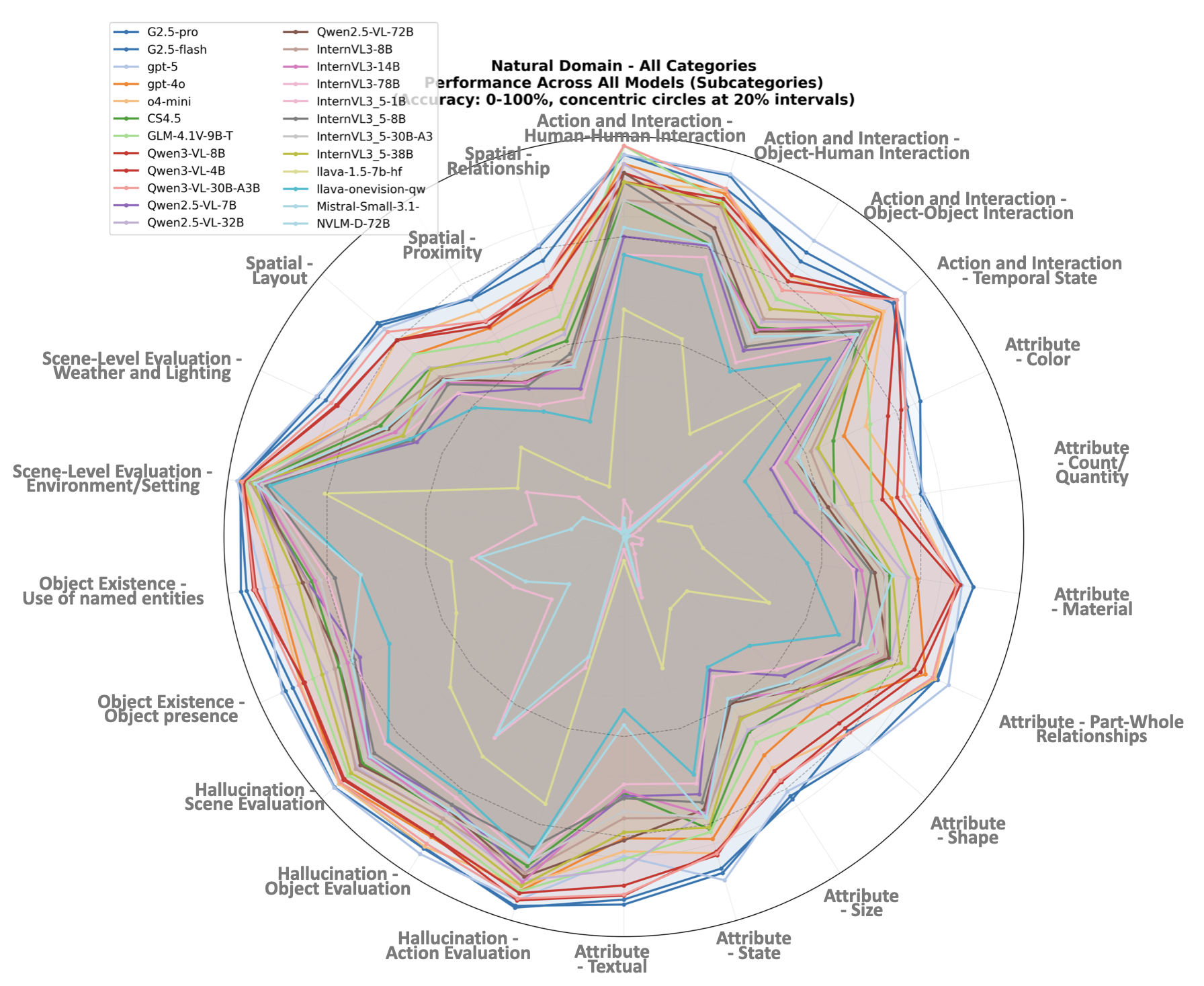}
\caption{\textbf{Natural domain: 22 subcategories.} Models perform best on scene-level evaluation (80-92\%) and object existence (75-90\%), but struggle with spatial reasoning (40-65\%) and fine-grained attributes (50-70\%). GPT-5 leads on most categories. Spatial subcategories (distance, orientation, relative position) show the largest performance gaps and highest variance.}

\label{fig:radar_natural}
\end{figure*}

\begin{figure*}[htb!]
\centering
\includegraphics[width=0.95\textwidth]{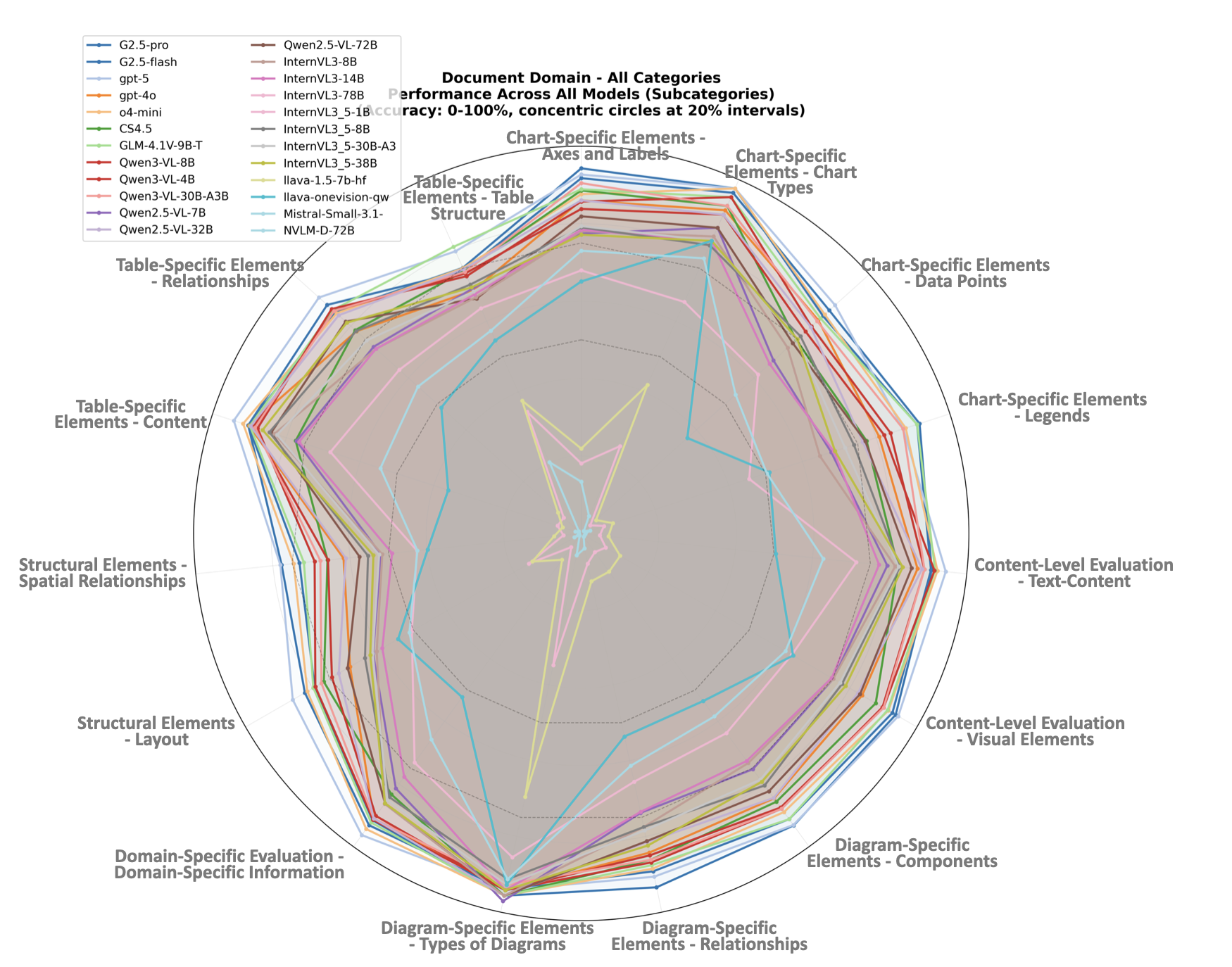}
\caption{\textbf{Document domain: 15 subcategories.} Models excel on high-level evaluation (80-93\%) but struggle with structural elements (50-75\%). Gemini models show relative strength on table/chart parsing. Variance is high across subcategories, with chart-specific elements (axis labels, legends) being particularly challenging.}

\label{fig:radar_document}
\end{figure*}

\begin{figure*}[htb!]
\centering
\includegraphics[width=0.95\textwidth]{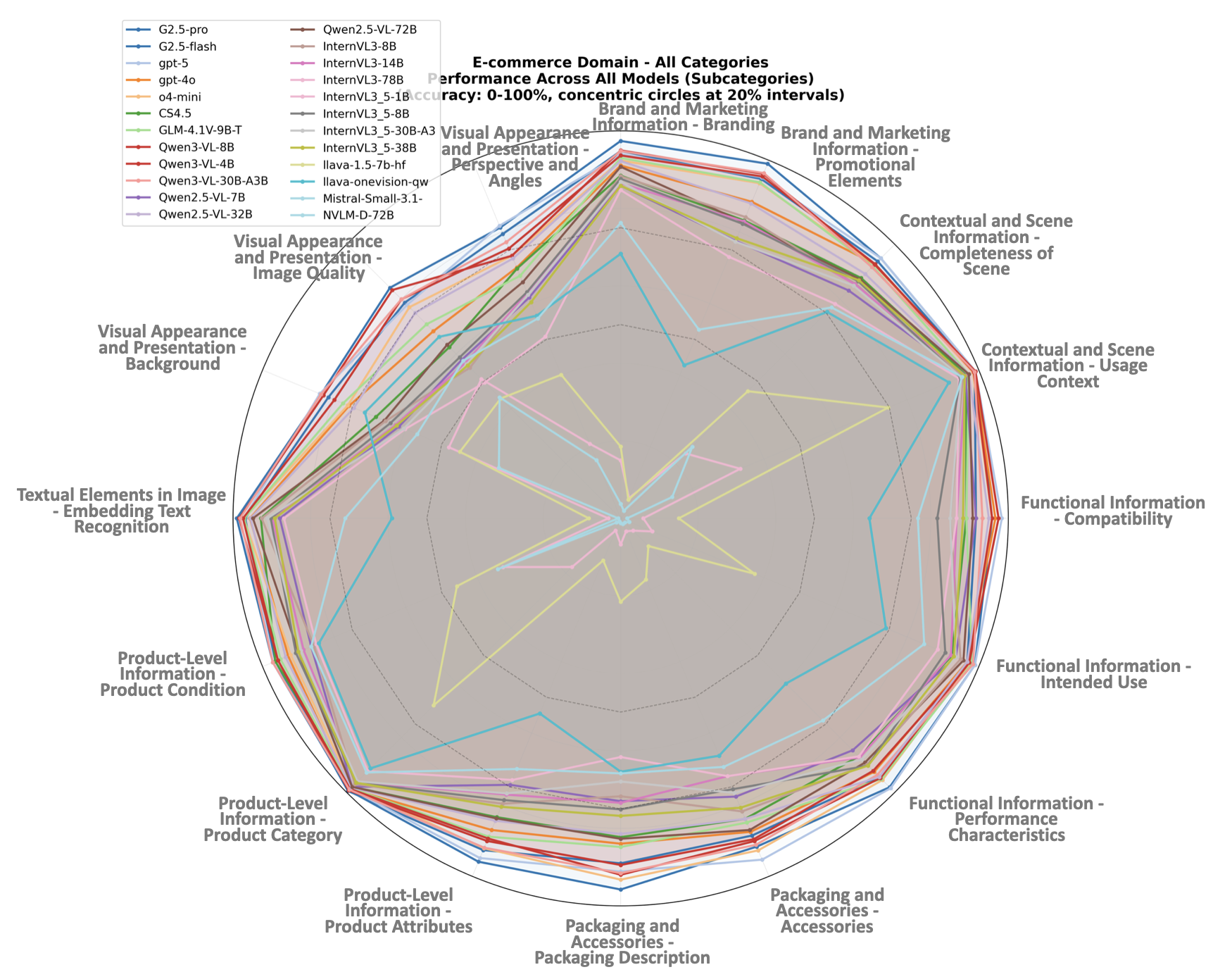}
\caption{\textbf{E-commerce domain: 16 subcategories.} Models achieve highest overall scores (70-96\%) across all domains. Contextual understanding (85-96\%) and product-level information (82-94\%) are strengths. Visual appearance details (color matching, style) are harder (60-75\%). Text extraction varies by text type.}

\label{fig:radar_ecommerce}
\end{figure*}

\begin{figure*}[htb!]
\centering
\includegraphics[width=0.95\textwidth]{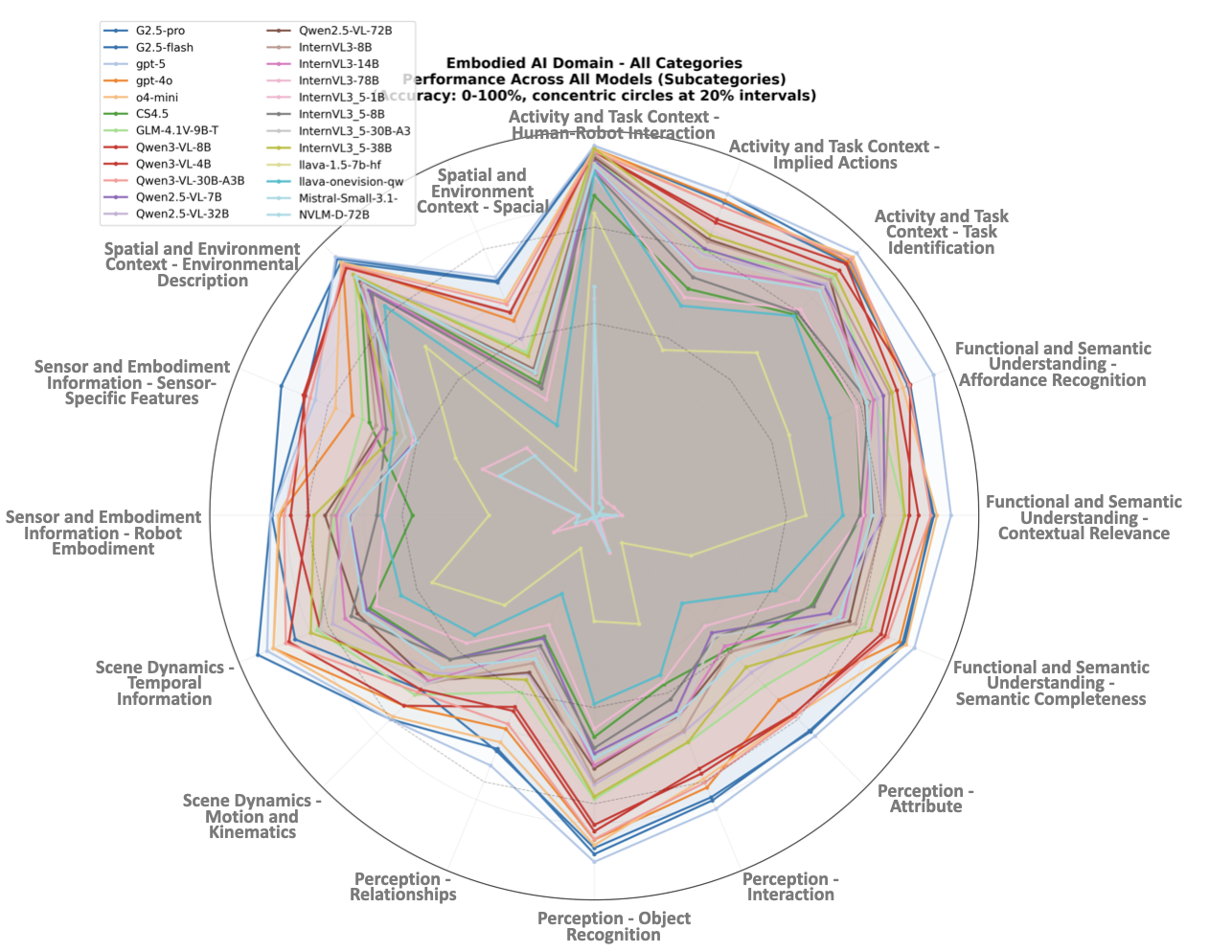}
\caption{\textbf{Embodied AI domain: 16 subcategories.} Most challenging domain overall (50-85\% range). Activity/task understanding (80-93\%) is relatively strong, but perception subcategories (object properties, affordances, manipulation) drop to 40-70\%. Sensor-specific information (depth, embodiment viewpoint) is systematically under-described.}
\label{fig:radar_embodied}
\end{figure*}

While Table~\ref{tab:category_summary} provides top-level statistics, Figures~\ref{fig:radar_natural}--\ref{fig:radar_embodied} break down performance across all 69 fine-grained subcategories, grouped by domain. Each radar chart shows all 24 evaluated models across all subcategories within that domain.

\subsubsection{Natural Domain}

Natural domain contains 22 subcategories spanning scene understanding, object recognition, attributes, actions, and spatial reasoning. Key findings: (1) \textbf{Scene-level categories are easiest}: Scene type classification (88\%), environment recognition (85\%), and overall scene description (82\%) achieve high scores across all models. (2) \textbf{Spatial subcategories are hardest}: Distance estimation (45\%), orientation (52\%), and relative positioning (58\%) show 20-30 point drops vs perceptual categories. (3) \textbf{Attribute granularity matters}: Coarse attributes (color, size: 75\%) outperform fine-grained ones (texture, material: 55\%).

\subsubsection{Document Domain}

Document domain contains 15 subcategories for charts, tables, diagrams, and text documents. Key findings: (1) \textbf{Content evaluation outperforms structure parsing}: Content-level evaluation (82\%) and domain-specific reasoning (80\%) are 15-20 points higher than structural parsing (60-65\%). (2) \textbf{Chart elements are inconsistent}: Axis labels (62\%), legends (58\%), and data point extraction (55\%) show high variance (std$>$12). (3) \textbf{Gemini advantage on tables}: Gemini 2.5 Pro leads by 5-8 points on table-specific subcategories (cell content, row/column understanding), suggesting better OCR integration.

\subsubsection{E-commerce Domain}

E-commerce domain contains 16 subcategories for product images. Key findings: (1) \textbf{Highest overall performance}: Mean across subcategories (81\%) exceeds other domains by 8-12 points. Product images may have cleaner backgrounds and clearer focal objects. (2) \textbf{Text extraction type-dependent}: Brand names (85\%) and product titles (82\%) are captured well, but fine print (68\%) and specifications (65\%) are often missed. (3) \textbf{Visual appearance is relative}: Color description (75\%) works but style matching (62\%) and material appearance (58\%) are subjective and harder to ground.

\subsubsection{Embodied AI Domain}

Embodied AI domain contains 16 subcategories for robotics and embodied perception. Key findings: (1) \textbf{Task context vs perception gap}: Activity recognition (83\%) and task-level reasoning (78\%) are 15-20 points higher than object properties (60\%), affordances (55\%), and manipulation planning (52\%). High-level semantics are easier than action-relevant details. (2) \textbf{Sensor information is omitted}: Depth cues (48\%), camera viewpoint (52\%), and embodiment context (55\%) have $>$30\% Cannot-Answer rates, indicating captions rarely describe sensor geometry. (3) \textbf{Dynamics are hard}: Temporal dynamics (58\%), motion prediction (53\%), and scene changes (50\%) show captioning models struggle with implicit temporal reasoning from static images.

\section{Detailed Model Performance Analysis}
\label{sec:appendix_model_performance}

\begin{figure*}[t]
\centering
\includegraphics[width=0.95\textwidth]{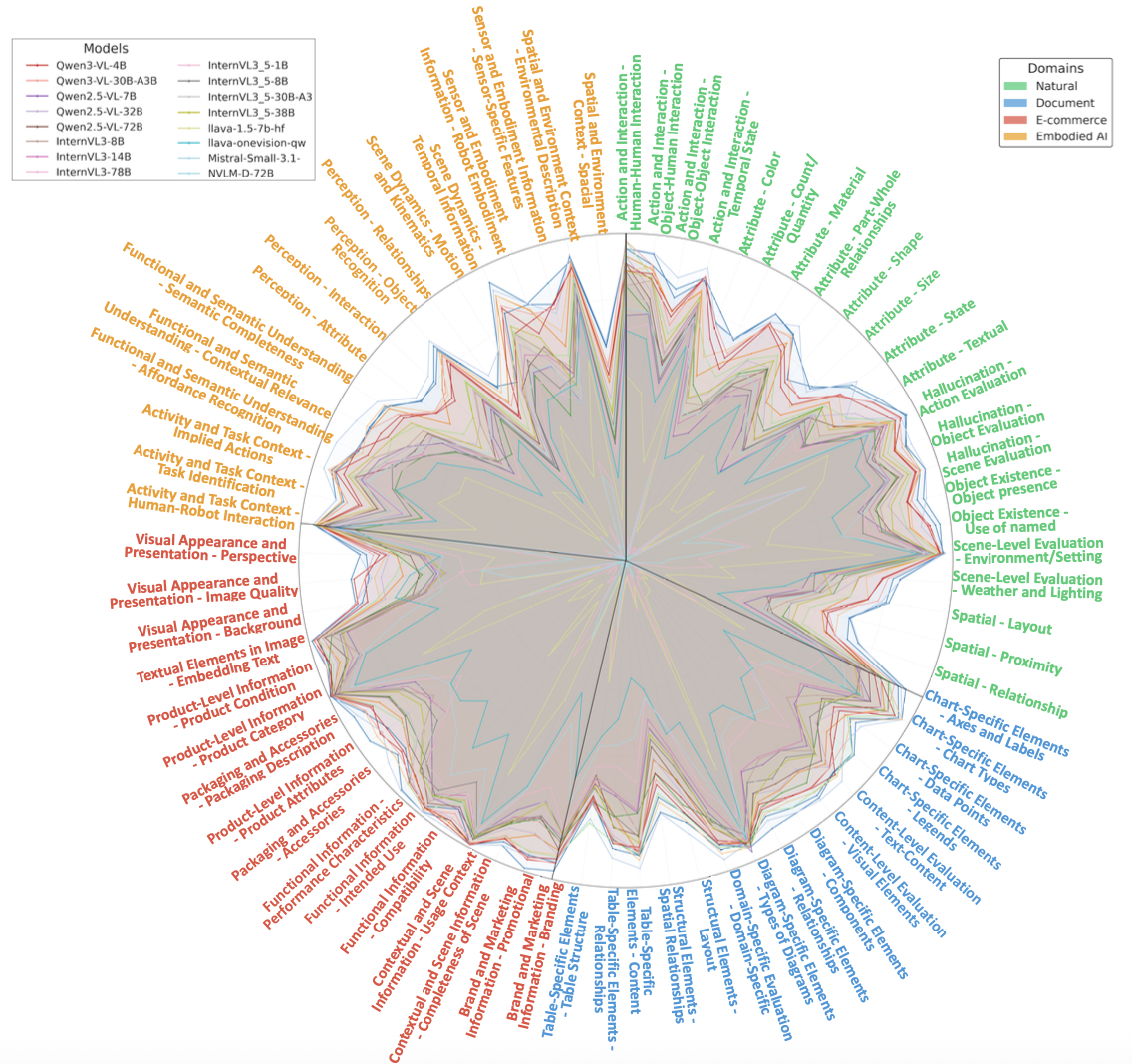}
\caption{\textbf{Comprehensive model performance across all 69 subcategories and 4 domains.}
Each colored line represents one of 24 evaluated models across 69 fine-grained subcategories.
The chart is divided into four domain sections (Natural, Document, E-commerce, Embodied AI) separated by black radial lines.
Concentric circles indicate accuracy levels from 0-100\% at 20\% intervals.
Top-performing models include\textbf{ GPT-5} and \textbf{Gemini 2.5 Flash} among proprietary models, and \textbf{Qwen3-VL 30B-A3B }and \textbf{GLM-4.1V 9B} among open-source models, though performance varies dramatically across categories within each model (30-95\% range), revealing that no single model excels uniformly across all task types.
Notably, categories requiring spatial reasoning (Embodied AI section) show consistently depressed performance across all models (40-70\%) compared to perceptual categories (70-90\%), indicating a systematic capability gap in current vision-language models.}
\label{fig:radar_comprehensive}
\end{figure*}

Figure~\ref{fig:radar_comprehensive} presents a unified view of all 24 evaluated models across all 69 subcategories and 4 domains. Each axis represents one subcategory, with axes colored by domain (Natural=green, Document=blue, E-commerce=red, Embodied AI=orange) and separated by black radial lines marking domain boundaries.

Three findings emerge from the comprehensive view. First, \textbf{model strengths are domain-specific}: GPT-5 leads on Natural and E-commerce categories, while Gemini models perform better on Document structural elements. Second, \textbf{spatial reasoning is hard for all models}: scores drop to 35-60\% on Embodied AI spatial categories vs 75-90\% on perceptual categories. Third, \textbf{category matters more than model}: within-model variance across categories (35-95\%) exceeds between-model variance on the same category (5-15 points), showing that \emph{what} to caption is harder than \emph{how well} to caption it.

\section{Question Difficulty Distribution}
\label{sec:difficulty}

To assess whether CaptionQA provides adequate discrimination across different capability levels, we analyze question difficulty based on the percentage of models that answer each question correctly.

\subsection{Difficulty Categorization}

We categorize questions into three difficulty levels based on the proportion of models (out of 24 total) that answer correctly:
\begin{itemize}[leftmargin=*,noitemsep,topsep=4pt]
    \item \textbf{Easy:} $\geq$80\% of models answer correctly
    \item \textbf{Medium:} 50-80\% of models answer correctly
    \item \textbf{Hard:} $<$50\% of models answer correctly
\end{itemize}

\subsection{Distribution Across Domains}

\begin{figure}[htb!]
\centering
\includegraphics[width=0.95\columnwidth]{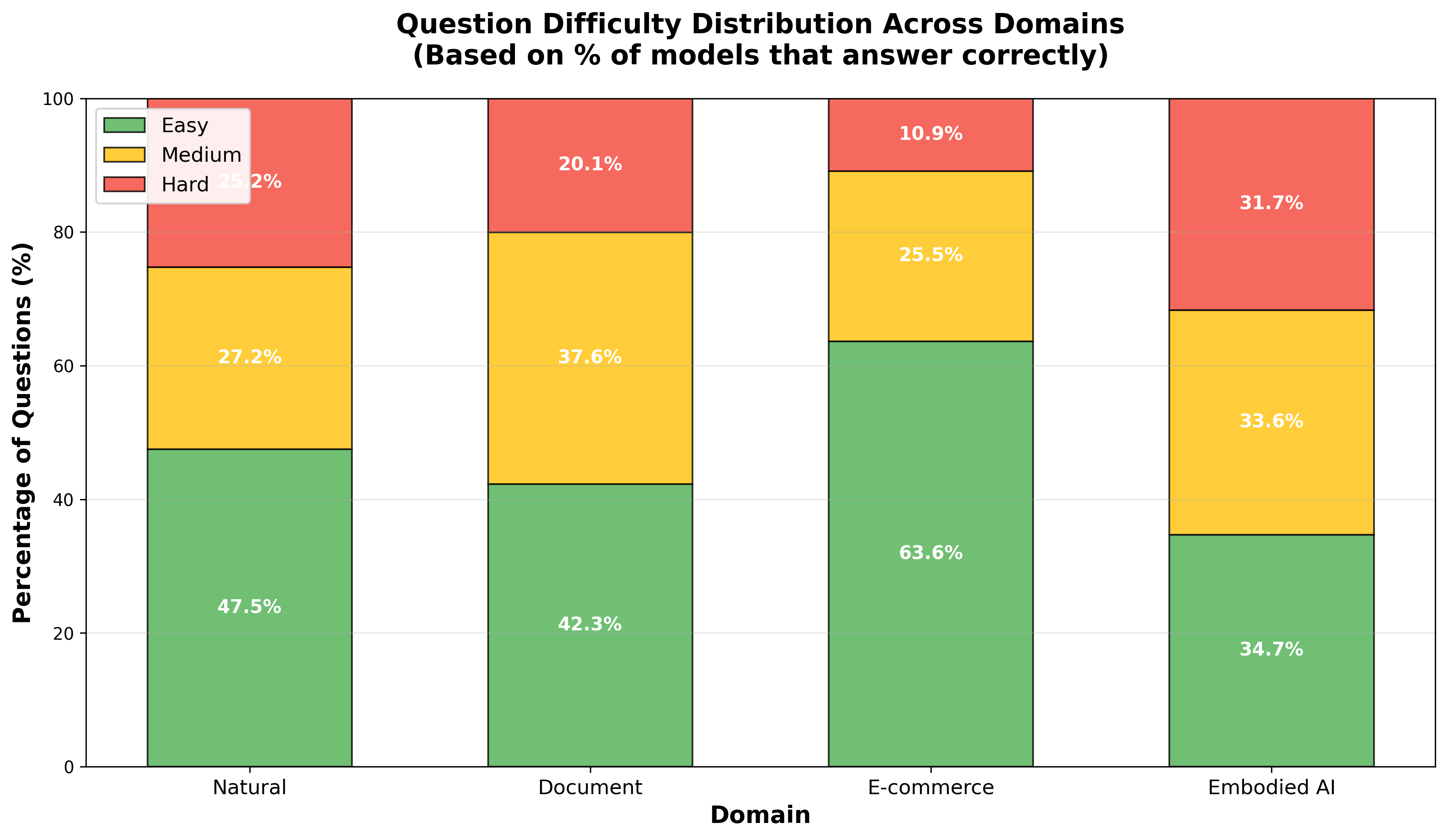}
\caption{\textbf{Question Difficulty Distribution Across Domains.} Each domain exhibits a different difficulty profile: E-commerce has more easy questions (64\%), while Embodied AI is the most challenging (32\% hard questions). This diversity ensures CaptionQA can discriminate between models at different capability levels.}
\label{fig:difficulty_dist}
\end{figure}

Figure~\ref{fig:difficulty_dist} shows the difficulty distribution for each domain.

\begin{table}[h]
\centering
\small
\caption{\textbf{Question Difficulty Distribution Summary.}}
\label{tab:difficulty_summary}
\begin{tabular}{l|ccc|c}
\toprule
\textbf{Domain} & \textbf{Easy} & \textbf{Medium} & \textbf{Hard} & \textbf{Total} \\
\midrule
Natural & 47.5\% & 27.2\% & 25.2\% & 10,445 \\
Document & 42.3\% & 37.6\% & 20.1\% & 7,417 \\
E-commerce & 63.6\% & 25.5\% & 10.9\% & 5,884 \\
Embodied AI & 34.7\% & 33.6\% & 31.7\% & 9,273 \\
\bottomrule
\end{tabular}
\end{table}

\subsection{Examples of Hardest and Easiest Questions}

\begin{figure*}[htb!]
\centering
\includegraphics[width=0.98\textwidth]{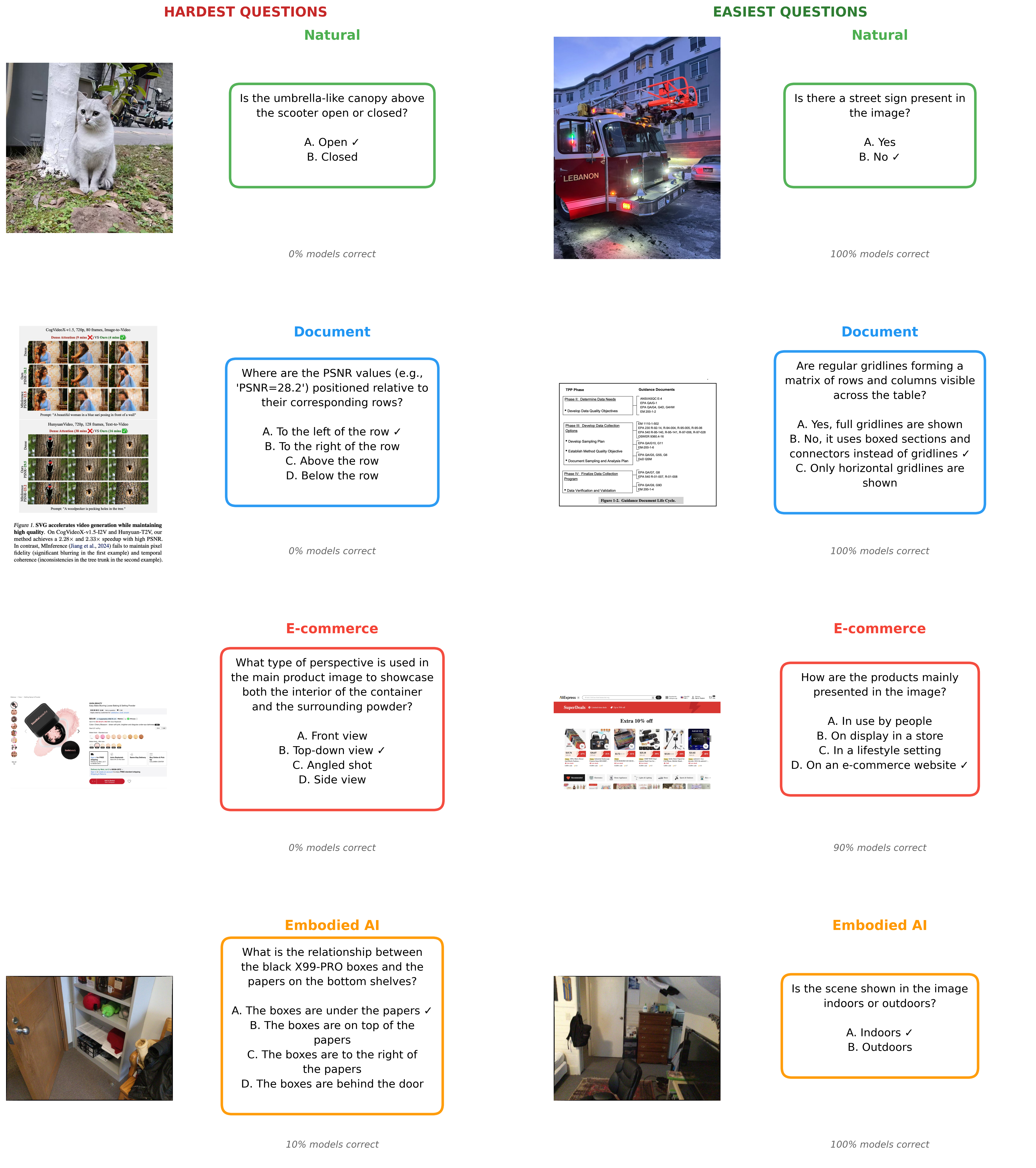}
\caption{\textbf{Example Questions with Images and Answer Choices: Hardest vs Easiest.} Left column shows hardest questions (0-10\% of models answered correctly), highlighting challenges in fine-grained spatial reasoning, technical detail recognition, and complex relational understanding. Right column shows easiest questions (90-100\% of models answered correctly), typically involving basic object presence or simple binary attributes. For each example, the image is shown on the left with the question text and multiple-choice options displayed together in a white box with domain-colored borders (green=Natural, blue=Document, red=E-commerce, orange=Embodied AI). The correct answer is marked with a checkmark (\checkmark). Note that different images are used for hardest vs easiest questions to avoid confounding factors.}
\label{fig:example_questions}
\end{figure*}

Figure~\ref{fig:example_questions} shows concrete examples of the hardest and easiest questions with their corresponding images, answer choices, and correct answers across all four domains.

\section{Full Results}
\label{sec:full-results}

We report the full CaptionQA results as shown in Table~\ref{tab:nat_long}--Table~\ref{tab:emb_tax} for all evaluated models, prompts, and domains in this section. These tables complement the main-paper summary (Table~3) by providing per-domain, per-prompt breakdowns, and by including all three metrics: Score, Acc, and Cannot. \emph{Models in the tables are ranked by score}.

\paragraph{Models covered.}
In total we evaluate \textbf{24} multimodal LLMs, spanning both open-source and proprietary systems and covering a wide range of scales (from 1B to 78B parameters for open-source models, and large API-only models on the proprietary side).

\textbf{Open-source VLMs.}
Our open-source pool includes 18 models from five major families:
\begin{itemize}
    \item \textbf{Qwen3-VL}: 4B, 8B, and 30B-A3B.
    \item \textbf{Qwen2.5-VL}: 7B, 32B, and 72B.
    \item \textbf{GLM-4.1V}: 9B.
    \item \textbf{InternVL family}: InternVL3.5 (1B, 8B, 30B-A3B, 38B) and InternVL3 (8B, 14B, 78B).
    \item \textbf{Other baselines}: NVLM-D-72B, LLaVA-OneVision-7B, LLaVA-1.5-7B, and Mistral-Small-24B.
\end{itemize}

\textbf{Proprietary VLMs.}
We further evaluate 6 proprietary models:
\begin{itemize}
    \item \textbf{OpenAI}: GPT-5, GPT-4o, and GPT-o4-mini.
    \item \textbf{Google}: Gemini 2.5 Pro and Gemini 2.5 Flash.
    \item \textbf{Anthropic}: Claude Sonnet 4.5.
\end{itemize}
These models represent the current generation of large, API-only VLMs commonly deployed in production systems.

\paragraph{Prompts.}
Every model is evaluated under the same set of \textbf{four caption prompts} described in Section 3.3 of the main paper:
\begin{itemize}
    \item \textbf{Long}: “Write a very long and detailed caption describing the given image as comprehensively as possible.”
    \item \textbf{Short}: “Write a very short caption for the given image.”
    \item \textbf{Simple}: “Describe this image in detail.” (our recommended default setting).
    \item \textbf{Taxonomy-Hinted}: We supply the domain taxonomy as a list of aspect prompts (Top-level $\rightarrow$ Subcategory) and ask the model to describe the image from those perspectives.
\end{itemize}
All four prompts are applied to all four domains for each captioning model, so every model is evaluated in 16 settings (4 domains $\times$ 4 prompts).

\paragraph{Domains and metrics.}
The tables are organized by domain (\textbf{Natural}, \textbf{Document}, \textbf{E-commerce}, and \textbf{Embodied AI}). Within each domain we report one table per prompt (Long, Short, Simple, Taxonomy-Hinted). Each table contains three metrics:
\begin{itemize}
    \item \textbf{Score} (\%): our utility-oriented multiple-choice score averaged over all questions.
    \item \textbf{Acc} (\%): CaptionQA accuracy, i.e., fraction of questions where the QA reader selects the correct option.
    \item \textbf{Cannot} (\%): fraction of questions where the QA reader selects “Cannot answer from the caption.”, indicating that the caption is judged insufficient for that question.
\end{itemize}
Rows are grouped into \emph{Open-Source VLMs} and \emph{Proprietary VLMs}, and sorted by Score.

\paragraph{Table layout.}
For clarity, we mirror the same layout across domains. For example, Table~\ref{tab:nat_long} reports all models on the \textbf{Natural} domain under the \textbf{Long} prompt; analogous tables report the other prompts on the Natural domain, and similarly for \textbf{Document}, \textbf{E-commerce}, and \textbf{Embodied AI}. Together, these tables provide a complete view of CaptionQA performance across \emph{all} model--prompt--domain combinations evaluated in this work.

\begin{table}[t]
\centering
\scriptsize
\setlength{\tabcolsep}{2pt}
\caption{Results on the \textbf{Natural} domain with the \textbf{Long} prompt. 
Score and Acc are reported in \%. Cannot is the percentage of questions where the model caption does not contain an answer.}
\label{tab:nat_long}
\begin{tabular}{l c c c c}
\toprule
\textbf{Model} & \textbf{Size} & \textbf{Score$\uparrow$} & \textbf{Acc$\uparrow$} & \textbf{Cannot$\downarrow$} \\
\midrule
\multicolumn{5}{c}{\textbf{Open-Source VLMs}} \\
\midrule
Qwen3-VL                & 8B       & 85.69 & 83.11 & 11.8 \\
Qwen3-VL                & 30B-A3B  & 85.62 & 83.18 & 11.2 \\
Qwen3-VL                & 4B       & 83.95 & 80.87 & 14.2 \\
GLM-4.1V                & 9B       & 82.92 & 79.07 & 17.6 \\
Qwen2.5-VL              & 32B      & 78.95 & 74.58 & 20.0 \\
InternVL3.5             & 38B      & 78.70 & 73.94 & 21.9 \\
Qwen2.5-VL              & 72B      & 77.29 & 72.10 & 23.9 \\
InternVL3.5             & 30B-A3B  & 74.33 & 68.06 & 29.0 \\
Qwen2.5-VL              & 7B       & 73.84 & 67.61 & 28.7 \\
InternVL3.5             & 8B       & 73.73 & 67.24 & 29.9 \\
NVLM-D                  & 72B      & 72.66 & 66.20 & 29.8 \\
InternVL3               & 14B      & 72.41 & 65.68 & 31.3 \\
InternVL3.5             & 1B       & 72.22 & 65.65 & 30.5 \\
LLaVA-OneVision         & 7B       & 70.20 & 62.94 & 33.6 \\
InternVL3               & 8B       & 66.39 & 57.10 & 42.9 \\
LLaVA-1.5               & 7B       & 53.06 & 39.14 & 64.7 \\
InternVL3               & 78B      & 38.66 & 19.44 & 89.6 \\
Mistral Small           & 24B      & 35.81 & 15.34 & 95.6 \\
\midrule
\multicolumn{5}{c}{\textbf{Proprietary VLMs}} \\
\midrule
GPT-5                   & --       & 90.34 & 88.89 &  6.7 \\
Gemini 2.5 Pro          & --       & 89.44 & 87.71 &  7.9 \\
Gemini 2.5 Flash        & --       & 89.28 & 87.57 &  7.8 \\
GPT-o4-mini             & --       & 85.99 & 83.62 & 11.0 \\
GPT-4o                  & --       & 84.71 & 81.60 & 14.3 \\
Claude Sonnet 4.5       & --       & 77.78 & 72.96 & 22.2 \\
\bottomrule
\end{tabular}
\end{table}

\begin{table}[t]
\centering
\scriptsize
\setlength{\tabcolsep}{2pt}
\caption{Results on the \textbf{Natural} domain with the \textbf{Simple} prompt. 
Score and Acc are reported in \%. Cannot is the percentage of questions where the model caption does not contain an answer.}
\label{tab:nat_simple}
\begin{tabular}{l c c c c}
\toprule
\textbf{Model} & \textbf{Size} & \textbf{Score$\uparrow$} & \textbf{Acc$\uparrow$} & \textbf{Cannot$\downarrow$} \\
\midrule
\multicolumn{5}{c}{\textbf{Open-Source VLMs}} \\
\midrule
Qwen3-VL                & 30B-A3B  & 86.14 & 83.68 & 11.3 \\
Qwen3-VL                & 8B       & 85.25 & 82.46 & 12.7 \\
Qwen3-VL                & 4B       & 84.73 & 81.72 & 13.7 \\
GLM-4.1V                & 9B       & 81.67 & 77.48 & 19.2 \\
Qwen2.5-VL              & 32B      & 78.35 & 73.60 & 21.8 \\
InternVL3.5             & 38B      & 78.26 & 72.95 & 24.4 \\
Qwen2.5-VL              & 72B      & 75.26 & 69.34 & 27.4 \\
InternVL3.5             & 30B-A3B  & 74.58 & 68.43 & 28.5 \\
InternVL3-14B           & 14B      & 74.16 & 67.71 & 29.8 \\
NVLM-D                  & 72B      & 73.13 & 66.74 & 29.6 \\
InternVL3.5             & 8B       & 72.97 & 66.43 & 30.2 \\
Qwen2.5-VL              & 7B       & 71.64 & 64.58 & 32.6 \\
InternVL3.5             & 1B       & 70.82 & 63.55 & 33.6 \\
LLaVA-OneVision         & 7B       & 66.56 & 57.91 & 40.1 \\
LLaVA-1.5               & 7B       & 52.51 & 38.53 & 65.0 \\
InternVL3               & 78B      & 38.86 & 19.72 & 89.3 \\
Mistral Small           & 24B      & 35.91 & 15.26 & 96.5 \\
\midrule
\multicolumn{5}{c}{\textbf{Proprietary VLMs}} \\
\midrule
Gemini 2.5 Flash        & --       & 88.95 & 87.10 &  8.4 \\
GPT-5                   & --       & 88.78 & 86.67 &  9.7 \\
Gemini 2.5 Pro          & --       & 87.89 & 85.72 & 10.0 \\
GPT-4o                  & --       & 82.69 & 78.84 & 17.7 \\
GPT-o4-mini             & --       & 84.66 & 81.66 & 13.8 \\
Claude Sonnet 4.5       & --       & 76.56 & 71.12 & 25.0 \\
\bottomrule
\end{tabular}
\end{table}

\begin{table}[t]
\centering
\scriptsize
\setlength{\tabcolsep}{2pt}
\caption{Results on the \textbf{Natural} domain with the \textbf{Short} prompt. 
Score and Acc are reported in \%. Cannot is the percentage of questions where the model caption does not contain an answer.}
\label{tab:nat_short}
\begin{tabular}{l c c c c}
\toprule
\textbf{Model} & \textbf{Size} & \textbf{Score$\uparrow$} & \textbf{Acc$\uparrow$} & \textbf{Cannot$\downarrow$} \\
\midrule
\multicolumn{5}{c}{\textbf{Open-Source VLMs}} \\
\midrule
GLM-4.1V                & 9B       & 62.43 & 50.58 & 54.8 \\
Qwen3-VL                & 4B       & 60.52 & 48.63 & 55.0 \\
Qwen3-VL                & 30B-A3B  & 59.48 & 47.23 & 56.7 \\
Qwen3-VL                & 8B       & 57.38 & 44.00 & 62.0 \\
InternVL3.5             & 38B      & 56.52 & 43.21 & 61.7 \\
InternVL3               & 8B       & 56.50 & 43.13 & 61.9 \\
InternVL3.5             & 1B       & 55.28 & 41.55 & 63.7 \\
InternVL3.5             & 30B-A3B  & 55.16 & 41.28 & 64.4 \\
InternVL3-14B           & 14B      & 55.13 & 41.12 & 64.9 \\
InternVL3.5             & 8B       & 54.89 & 40.94 & 64.7 \\
Qwen2.5-VL              & 7B       & 54.77 & 40.73 & 65.1 \\
Qwen2.5-VL              & 32B      & 54.74 & 40.68 & 65.2 \\
Qwen2.5-VL              & 72B      & 54.70 & 40.42 & 66.1 \\
NVLM-D                  & 72B      & 51.29 & 36.02 & 70.8 \\
LLaVA-OneVision         & 7B       & 51.01 & 35.58 & 71.6 \\
LLaVA-1.5               & 7B       & 49.05 & 32.92 & 74.9 \\
InternVL3               & 78B      & 38.24 & 18.48 & 92.2 \\
Mistral Small           & 24B      & 36.21 & 15.50 & 96.7 \\
\midrule
\multicolumn{5}{c}{\textbf{Proprietary VLMs}} \\
\midrule
Claude Sonnet 4.5       & --       & 56.77 & 43.38 & 62.0 \\
GPT-5                   & --       & 55.65 & 41.62 & 64.6 \\
Gemini 2.5 Pro          & --       & 55.33 & 41.17 & 65.6 \\
Gemini 2.5 Flash        & --       & 55.13 & 40.67 & 67.0 \\
GPT-o4-mini             & --       & 54.76 & 40.61 & 65.6 \\
GPT-4o                  & --       & 54.19 & 39.67 & 67.4 \\
\bottomrule
\end{tabular}
\end{table}

\begin{table}[t]
\centering
\scriptsize
\setlength{\tabcolsep}{2pt}
\caption{Results on the \textbf{Natural} domain with the \textbf{Taxonomy-Hinted} prompt. Score and Acc are reported in \%. Cannot is the percentage of questions where the model caption does not contain an answer.}
\label{tab:nat_tax_default}
\begin{tabular}{l c c c c}
\toprule
\textbf{Model} & \textbf{Size} & \textbf{Score$\uparrow$} & \textbf{Acc$\uparrow$} & \textbf{Cannot$\downarrow$} \\
\midrule
\multicolumn{5}{c}{\textbf{Open-Source VLMs}} \\
\midrule
Qwen3-VL                & 8B       & 78.99 & 73.78 & 23.9 \\
Qwen3-VL                & 4B       & 76.30 & 70.27 & 27.7 \\
GLM-4.1V                & 9B       & 75.87 & 69.79 & 27.9 \\
Qwen2.5-VL              & 32B      & 74.64 & 68.51 & 28.2 \\
Qwen3-VL                & 30B-A3B  & 74.46 & 68.29 &  7.2 \\
Qwen2.5-VL              & 72B      & 72.93 & 66.25 & 30.6 \\
NVLM-D                  & 72B      & 72.63 & 66.25 & 29.4 \\
InternVL3               & 8B       & 71.57 & 64.10 & 34.4 \\
InternVL3.5             & 38B      & 69.30 & 61.24 & 37.2 \\
Qwen2.5-VL              & 7B       & 68.46 & 59.66 & 40.7 \\
InternVL3-14B           & 14B      & 68.27 & 59.60 & 40.1 \\
InternVL3.5             & 8B       & 65.77 & 56.29 & 43.9 \\
InternVL3.5             & 30B-A3B  & 65.66 & 56.12 & 44.1 \\
InternVL3.5             & 1B       & 65.58 & 56.13 & 43.7 \\
LLaVA-OneVision         & 7B       & 63.24 & 53.14 & 46.8 \\
LLaVA-1.5               & 7B       & 51.24 & 36.58 & 68.1 \\
InternVL3               & 78B      & 38.39 & 19.33 & 88.9 \\
Mistral Small           & 24B      & 35.95 & 15.71 & 94.5 \\
\midrule
\multicolumn{5}{c}{\textbf{Proprietary VLMs}} \\
\midrule
Gemini 2.5 Flash        & --       & 87.55 & 85.11 & 11.0 \\
Gemini 2.5 Pro          & --       & 87.47 & 85.01 & 11.2 \\
GPT-5                   & --       & 86.87 & 83.79 & 14.1 \\
GPT-o4-mini             & --       & 82.50 & 78.57 & 18.0 \\
GPT-4o                  & --       & 78.30 & 72.82 & 25.1 \\
Claude Sonnet 4.5       & --       & 77.36 & 72.14 & 24.1 \\
\bottomrule
\end{tabular}
\end{table}

\begin{table}[t]
\centering
\scriptsize
\setlength{\tabcolsep}{2pt}
\caption{Results on the \textbf{Document} domain with the \textbf{Long} prompt. 
Score and Acc are reported in \%. Cannot is the percentage of questions where the model caption does not contain an answer.}
\label{tab:doc_long}
\begin{tabular}{l c c c c}
\toprule
\textbf{Model} & \textbf{Size} & \textbf{Score$\uparrow$} & \textbf{Acc$\uparrow$} & \textbf{Cannot$\downarrow$} \\
\midrule
\multicolumn{5}{c}{\textbf{Open-Source VLMs}} \\
\midrule
GLM-4.1V                & 9B  & 88.34 & 87.00 &  4.6 \\
Qwen3-VL                & 8B  & 86.63 & 85.19 &  4.9 \\
Qwen3-VL           & 30B-A3B & 86.05 & 84.46 &  5.5 \\
Qwen3-VL                & 4B  & 84.63 & 82.72 &  6.6 \\
Qwen2.5-VL             & 32B & 82.14 & 79.54 &  9.0 \\
Qwen2.5-VL             & 72B & 80.36 & 77.11 & 11.2 \\
InternVL3.5            & 38B & 78.78 & 74.92 & 13.4 \\
InternVL3.5             & 8B  & 77.29 & 73.21 & 14.1 \\
InternVL3.5        & 30B-A3B & 76.45 & 71.86 & 15.9 \\
Qwen2.5-VL              & 7B  & 76.35 & 72.08 & 14.8 \\
InternVL3               & 8B  & 72.95 & 67.71 & 18.1 \\
InternVL3              & 14B & 71.88 & 66.35 & 19.1 \\
InternVL3.5             & 1B  & 67.50 & 60.33 & 24.7 \\
NVLM-D                 & 72B & 65.39 & 57.73 & 26.5 \\
LLaVA-OneVision         & 7B  & 61.93 & 52.41 & 32.9 \\
LLaVA-1.5               & 7B  & 36.22 & 12.05 & 83.7 \\
InternVL3              & 78B & 34.29 &  9.01 & 87.5 \\
Mistral Small          & 24B & 30.75 &  2.56 & 97.6 \\
\midrule
\multicolumn{5}{c}{\textbf{Proprietary VLMs}} \\
\midrule
GPT-5                      & --  & 90.01 & 89.11 &  3.1 \\
Gemini 2.5 Pro             & --  & 88.67 & 87.60 &  3.7 \\
GPT-o4-mini                & --  & 88.38 & 87.15 &  4.2 \\
Gemini 2.5 Flash           & --  & 86.90 & 85.34 &  5.4 \\
Claude Sonnet 4.5          & --  & 85.08 & 82.70 &  8.2 \\
GPT-4o                     & --  & 82.18 & 79.34 &  9.8 \\
\bottomrule
\end{tabular}
\end{table}

\begin{table}[t]
\centering
\scriptsize
\setlength{\tabcolsep}{2pt}
\caption{Results on the \textbf{Document} domain with the \textbf{Simple} prompt. 
Score and Acc are reported in \%. Cannot is the percentage of questions where the model caption does not contain an answer.}
\label{tab:doc_simple}
\begin{tabular}{l c c c c}
\toprule
\textbf{Model} & \textbf{Size} & \textbf{Score$\uparrow$} & \textbf{Acc$\uparrow$} & \textbf{Cannot$\downarrow$} \\
\midrule
\multicolumn{5}{c}{\textbf{Open-Source VLMs}} \\
\midrule
GLM-4.1V           & 9B       & 87.86 & 86.36 &  5.1 \\
Qwen3-VL           & 30B-A3B  & 85.89 & 84.21 &  5.8 \\
Qwen3-VL           & 8B       & 85.85 & 84.10 &  6.0 \\
Qwen3-VL           & 4B       & 84.99 & 83.02 &  6.8 \\
Qwen2.5-VL         & 32B      & 82.67 & 80.08 &  8.9 \\
Qwen2.5-VL         & 72B      & 80.56 & 77.42 & 10.8 \\
InternVL3.5        & 38B      & 78.91 & 74.92 & 13.8 \\
InternVL3.5        & 8B       & 78.56 & 74.72 & 13.3 \\
InternVL3.5        & 30B-A3B  & 77.72 & 73.64 & 14.1 \\
InternVL3          & 8B       & 75.83 & 71.66 & 14.4 \\
Qwen2.5-VL         & 7B       & 75.85 & 71.07 & 16.5 \\
InternVL3          & 14B      & 74.17 & 69.55 & 16.1 \\
InternVL3.5        & 1B       & 68.08 & 61.30 & 23.4 \\
NVLM-D             & 72B      & 65.25 & 57.27 & 27.6 \\
LLaVA-OneVision    & 7B       & 61.45 & 52.11 & 32.3 \\
LLaVA-1.5          & 7B       & 36.48 & 12.22 & 84.0 \\
InternVL3          & 78B      & 34.19 &  9.03 & 87.2 \\
Mistral Small      & 24B      & 30.81 &  2.56 & 97.9 \\
\midrule
\multicolumn{5}{c}{\textbf{Proprietary VLMs}} \\
\midrule
GPT-5              & --       & 90.81 & 90.09 &  2.5 \\
Gemini 2.5 Flash   & --       & 88.97 & 87.75 &  4.1 \\
Gemini 2.5 Pro     & --       & 88.66 & 87.42 &  4.3 \\
GPT-o4-mini        & --       & 88.14 & 86.84 &  4.5 \\
Claude Sonnet 4.5  & --       & 83.09 & 80.18 & 10.0 \\
GPT-4o             & --       & 82.55 & 79.87 &  9.3 \\
\bottomrule
\end{tabular}
\end{table}

\begin{table}[t]
\centering
\scriptsize
\setlength{\tabcolsep}{2pt}
\caption{Results on the \textbf{Document} domain with the \textbf{Short} prompt. 
Score and Acc are reported in \%. Cannot is the percentage of questions where the model caption does not contain an answer.}
\label{tab:doc_short}
\begin{tabular}{l c c c c}
\toprule
\textbf{Model} & \textbf{Size} & \textbf{Score$\uparrow$} & \textbf{Acc$\uparrow$} & \textbf{Cannot$\downarrow$} \\
\midrule
\multicolumn{5}{c}{\textbf{Open-Source VLMs}} \\
\midrule
GLM-4.1V           & 9B       & 55.69 & 40.47 & 52.7 \\
Qwen3-VL           & 30B-A3B  & 48.90 & 32.04 & 58.5 \\
Qwen3-VL           & 4B       & 48.12 & 31.28 & 58.4 \\
Qwen2.5-VL         & 32B      & 45.32 & 26.20 & 66.2 \\
InternVL3.5        & 8B       & 45.26 & 26.01 & 66.7 \\
InternVL3.5        & 30B-A3B  & 44.76 & 25.60 & 66.4 \\
InternVL3          & 8B       & 44.58 & 25.13 & 67.4 \\
InternVL3          & 14B      & 44.57 & 25.54 & 66.0 \\
InternVL3.5        & 38B      & 44.24 & 24.50 & 68.4 \\
InternVL3.5        & 1B       & 43.69 & 23.61 & 69.6 \\
Qwen2.5-VL         & 7B       & 43.32 & 23.07 & 70.1 \\
Qwen3-VL           & 8B       & 43.27 & 22.98 & 70.3 \\
Qwen2.5-VL         & 72B      & 42.93 & 22.33 & 71.4 \\
NVLM-D             & 72B      & 40.99 & 19.26 & 75.3 \\
LLaVA-OneVision    & 7B       & 38.28 & 14.70 & 81.7 \\
LLaVA-1.5          & 7B       & 35.33 & 10.12 & 87.3 \\
InternVL3          & 78B      & 33.70 &  7.48 & 90.8 \\
Mistral Small      & 24B      & 30.78 &  2.26 & 98.8 \\
\midrule
\multicolumn{5}{c}{\textbf{Proprietary VLMs}} \\
\midrule
GPT-o4-mini        & --       & 46.40 & 27.99 & 63.7 \\
Gemini 2.5 Pro     & --       & 46.25 & 27.73 & 64.1 \\
Claude Sonnet 4.5  & --       & 46.04 & 27.64 & 63.7 \\
GPT-5              & --       & 44.85 & 25.79 & 66.0 \\
Gemini 2.5 Flash   & --       & 44.09 & 24.51 & 67.8 \\
GPT-4o             & --       & 42.84 & 22.29 & 71.2 \\
\bottomrule
\end{tabular}
\end{table}

\begin{table}[t]
\centering
\scriptsize
\setlength{\tabcolsep}{2pt}
\caption{Results on the \textbf{Document} domain with the \textbf{Taxonomy-Hinted} prompt.
Score and Acc are reported in \%. Cannot is the percentage of questions where the model caption does not contain an answer.}
\label{tab:doc_tax_default}
\begin{tabular}{l c c c c}
\toprule
\textbf{Model} & \textbf{Size} & \textbf{Score$\uparrow$} & \textbf{Acc$\uparrow$} & \textbf{Cannot$\downarrow$} \\
\midrule
\multicolumn{5}{c}{\textbf{Open-Source VLMs}} \\
\midrule
GLM-4.1V                & 9B       & 72.40 & 67.10 & 18.3 \\
Qwen2.5-VL              & 32B      & 71.44 & 65.77 & 19.5 \\
Qwen3-VL                & 8B       & 70.82 & 64.83 & 20.7 \\
Qwen3-VL                & 4B       & 66.63 & 58.82 & 27.1 \\
Qwen3-VL                & 30B-A3B  & 62.49 & 52.26 & 35.3 \\
Qwen2.5-VL              & 72B      & 62.48 & 53.66 & 30.5 \\
NVLM-D                  & 72B      & 59.61 & 49.29 & 35.7 \\
Qwen2.5-VL              & 7B       & 57.05 & 44.95 & 41.8 \\
LLaVA-OneVision         & 7B       & 56.57 & 44.48 & 41.8 \\
InternVL3.5             & 1B       & 56.37 & 44.04 & 42.7 \\
InternVL3               & 8B       & 55.05 & 42.42 & 43.6 \\
InternVL3               & 14B      & 54.52 & 41.36 & 45.5 \\
InternVL3.5             & 8B       & 52.60 & 38.08 & 50.1 \\
InternVL3.5             & 38B      & 51.94 & 37.38 & 50.4 \\
InternVL3.5             & 30B-A3B  & 51.83 & 37.53 & 49.5 \\
LLaVA-1.5               & 7B       & 35.65 & 10.94 & 85.5 \\
InternVL3               & 78B      & 33.41 &  7.29 & 90.4 \\
Mistral Small           & 24B      & 30.60 &  2.39 & 97.8 \\
\midrule
\multicolumn{5}{c}{\textbf{Proprietary VLMs}} \\
\midrule
Claude Sonnet 4.5       & --       & 83.23 & 80.90 &  8.0 \\
Gemini 2.5 Flash        & --       & 81.89 & 79.36 &  8.7 \\
Gemini 2.5 Pro          & --       & 81.34 & 78.71 &  9.0 \\
GPT-o4-mini             & --       & 75.08 & 69.62 & 18.8 \\
GPT-5                   & --       & 72.36 & 65.39 & 24.1 \\
GPT-4o                  & --       & 63.48 & 54.58 & 30.8 \\
\bottomrule
\end{tabular}
\end{table}

\begin{table}[t]
\centering
\scriptsize
\setlength{\tabcolsep}{2pt}
\caption{Results on the \textbf{E-commerce} domain with the \textbf{Long} prompt. 
Score and Acc are reported in \%. Cannot is the percentage of questions where the model caption does not contain an answer.}
\label{tab:ecom_long}
\begin{tabular}{l c c c c}
\toprule
\textbf{Model} & \textbf{Size} & \textbf{Score$\uparrow$} & \textbf{Acc$\uparrow$} & \textbf{Cannot$\downarrow$} \\
\midrule
\multicolumn{5}{c}{\textbf{Open-Source VLMs}} \\
\midrule
Qwen3-VL                & 30B-A3B  & 94.27 & 93.34 &  3.6 \\
Qwen3-VL                & 8B       & 93.95 & 92.86 &  4.2 \\
Qwen3-VL                & 4B       & 93.62 & 92.54 &  4.2 \\
GLM-4.1V                & 9B       & 93.03 & 91.45 &  6.1 \\
Qwen2.5-VL              & 32B      & 90.76 & 88.97 &  6.9 \\
Qwen2.5-VL              & 72B      & 89.85 & 87.58 &  8.8 \\
InternVL3.5             & 38B      & 87.53 & 84.20 & 12.8 \\
Qwen2.5-VL              & 7B       & 87.21 & 84.13 & 11.8 \\
InternVL3.5             & 30B-A3B  & 85.78 & 81.87 & 15.1 \\
InternVL3.5             & 8B       & 85.77 & 82.03 & 14.5 \\
InternVL3               & 8B       & 84.53 & 80.50 & 15.6 \\
InternVL3               & 14B      & 84.05 & 79.77 & 16.6 \\
InternVL3.5             & 1B       & 80.12 & 74.29 & 22.5 \\
NVLM-D                  & 72B      & 79.05 & 73.07 & 23.1 \\
LLaVA-OneVision         & 7B       & 77.24 & 70.51 & 26.1 \\
LLaVA-1.5               & 7B       & 48.39 & 30.43 & 69.6 \\
InternVL3               & 78B      & 38.46 & 16.51 & 85.1 \\
Mistral Small           & 24B      & 34.59 & 10.75 & 92.5 \\
\midrule
\multicolumn{5}{c}{\textbf{Proprietary VLMs}} \\
\midrule
GPT-5                   & --       & 96.11 & 95.62 &  1.9 \\
Gemini 2.5 Pro          & --       & 95.60 & 94.94 &  2.6 \\
Gemini 2.5 Flash        & --       & 95.55 & 94.94 &  2.4 \\
GPT-o4-mini             & --       & 94.22 & 93.32 &  3.5 \\
GPT-4o                  & --       & 91.14 & 89.07 &  8.0 \\
Claude Sonnet 4.5       & --       & 91.11 & 88.94 &  8.4 \\
\bottomrule
\end{tabular}
\end{table}

\begin{table}[t]
\centering
\scriptsize
\setlength{\tabcolsep}{2pt}
\caption{Results on the \textbf{E-commerce} domain with the \textbf{Simple} prompt. 
Score and Acc are reported in \%. Cannot is the percentage of questions where the model caption does not contain an answer.}
\label{tab:ecom_simple}
\begin{tabular}{l c c c c}
\toprule
\textbf{Model} & \textbf{Size} & \textbf{Score$\uparrow$} & \textbf{Acc$\uparrow$} & \textbf{Cannot$\downarrow$} \\
\midrule
\multicolumn{5}{c}{\textbf{Open-Source VLMs}} \\
\midrule
Qwen3-VL                & 30B-A3B  & 93.90 & 92.85 &  4.1 \\
Qwen3-VL                & 4B       & 93.77 & 92.63 &  4.4 \\
Qwen3-VL                & 8B       & 93.35 & 92.05 &  5.1 \\
GLM-4.1V                & 9B       & 92.04 & 90.18 &  7.1 \\
Qwen2.5-VL              & 32B      & 90.81 & 88.75 &  7.9 \\
Qwen2.5-VL              & 72B      & 89.07 & 86.20 & 11.0 \\
Qwen2.5-VL              & 7B       & 85.38 & 81.31 & 15.7 \\
InternVL3               & 8B       & 87.01 & 83.59 & 13.2 \\
InternVL3-14B           & 14B      & 86.17 & 82.35 & 14.8 \\
InternVL3.5             & 38B      & 86.47 & 82.69 & 14.6 \\
InternVL3.5             & 8B       & 86.60 & 82.98 & 13.9 \\
InternVL3.5             & 30B-A3B  & 85.79 & 81.80 & 15.4 \\
InternVL3.5             & 1B       & 82.69 & 77.76 & 19.0 \\
NVLM-D                  & 72B      & 78.46 & 72.43 & 23.4 \\
LLaVA-OneVision         & 7B       & 75.09 & 67.38 & 30.0 \\
LLaVA-1.5               & 7B       & 49.00 & 31.12 & 69.2 \\
InternVL3               & 78B      & 38.47 & 16.58 & 84.9 \\
Mistral Small           & 24B      & 34.52 & 10.31 & 93.8 \\
\midrule
\multicolumn{5}{c}{\textbf{Proprietary VLMs}} \\
\midrule
Gemini 2.5 Flash        & --       & 95.73 & 95.09 &  2.5 \\
GPT-5                   & --       & 94.73 & 93.78 &  3.6 \\
Gemini 2.5 Pro          & --       & 93.91 & 92.69 &  4.7 \\
GPT-o4-mini             & --       & 93.18 & 91.67 &  5.8 \\
GPT-4o                  & --       & 91.40 & 89.28 &  8.2 \\
Claude Sonnet 4.5       & --       & 88.86 & 85.91 & 11.4 \\
\bottomrule
\end{tabular}
\end{table}

\begin{table}[t]
\centering
\scriptsize
\setlength{\tabcolsep}{2pt}
\caption{Results on the \textbf{E-commerce} domain with the \textbf{Short} prompt. 
Score and Acc are reported in \%. Cannot is the percentage of questions where the model caption does not contain an answer.}
\label{tab:ecom_short}
\begin{tabular}{l c c c c}
\toprule
\textbf{Model} & \textbf{Size} & \textbf{Score$\uparrow$} & \textbf{Acc$\uparrow$} & \textbf{Cannot$\downarrow$} \\
\midrule
\multicolumn{5}{c}{\textbf{Open-Source VLMs}} \\
\midrule
GLM-4.1V                & 9B       & 64.05 & 50.97 & 50.6 \\
Qwen3-VL                & 4B       & 61.69 & 48.32 & 51.6 \\
Qwen3-VL                & 30B-A3B  & 61.23 & 47.60 & 52.6 \\
InternVL3.5             & 38B      & 59.16 & 44.75 & 55.7 \\
Qwen2.5-VL              & 32B      & 57.89 & 43.17 & 56.9 \\
Qwen3-VL                & 8B       & 57.66 & 42.71 & 57.8 \\
InternVL3.5             & 30B-A3B  & 57.58 & 42.32 & 59.0 \\
Qwen2.5-VL              & 72B      & 57.06 & 41.68 & 59.4 \\
InternVL3               & 8B       & 56.99 & 41.62 & 59.4 \\
Qwen2.5-VL              & 7B       & 56.88 & 41.39 & 59.8 \\
InternVL3.5             & 8B       & 56.67 & 41.17 & 59.9 \\
InternVL3-14B           & 14B      & 56.27 & 40.60 & 60.5 \\
InternVL3.5             & 1B       & 55.23 & 39.26 & 61.7 \\
NVLM-D                  & 72B      & 50.88 & 33.28 & 68.1 \\
LLaVA-OneVision         & 7B       & 49.80 & 31.28 & 71.6 \\
LLaVA-1.5               & 7B       & 43.91 & 23.38 & 79.4 \\
InternVL3               & 78B      & 35.32 & 11.79 & 91.1 \\
Mistral Small           & 24B      & 33.03 &  8.17 & 96.3 \\
\midrule
\multicolumn{5}{c}{\textbf{Proprietary VLMs}} \\
\midrule
Claude Sonnet 4.5       & --       & 59.31 & 44.90 & 55.7 \\
Gemini 2.5 Pro          & --       & 58.98 & 44.02 & 57.8 \\
GPT-5                   & --       & 57.36 & 41.98 & 59.4 \\
GPT-o4-mini             & --       & 56.84 & 41.40 & 59.6 \\
Gemini 2.5 Flash        & --       & 56.77 & 41.03 & 60.8 \\
GPT-4o                  & --       & 55.48 & 39.29 & 62.5 \\
\bottomrule
\end{tabular}
\end{table}

\begin{table}[t]
\centering
\scriptsize
\setlength{\tabcolsep}{2pt}
\caption{Results on the \textbf{E-commerce} domain with the \textbf{Taxonomy-Hinted} prompt.
Score and Acc are reported in \%. Cannot is the percentage of questions where the model caption does not contain an answer.}
\label{tab:ecom_taxo}
\begin{tabular}{l c c c c}
\toprule
\textbf{Model} & \textbf{Size} & \textbf{Score$\uparrow$} & \textbf{Acc$\uparrow$} & \textbf{Cannot$\downarrow$} \\
\midrule
\multicolumn{5}{c}{\textbf{Open-Source VLMs}} \\
\midrule
Qwen3-VL                & 8B      & 84.19 & 80.29 & 15.1 \\
GLM-4.1V                & 9B      & 83.87 & 79.92 & 15.4 \\
Qwen2.5-VL              & 72B     & 83.80 & 79.95 & 15.0 \\
Qwen2.5-VL              & 32B     & 83.55 & 79.68 & 15.1 \\
Qwen3-VL                & 4B      & 82.98 & 78.83 & 16.1 \\
Qwen3-VL           & 30B-A3B   & 76.27 & 68.72 & 29.3 \\
NVLM-D                  & 72B     & 76.12 & 69.54 & 25.6 \\
InternVL3               & 8B      & 75.82 & 69.57 & 24.3 \\
Qwen2.5-VL              & 7B      & 75.21 & 68.71 & 25.3 \\
InternVL3               & 14B     & 74.83 & 67.84 & 27.2 \\
InternVL3.5             & 38B     & 73.46 & 65.65 & 30.4 \\
InternVL3.5        & 30B-A3B   & 73.27 & 65.34 & 30.8 \\
InternVL3.5             & 8B      & 71.70 & 63.49 & 31.9 \\
LLaVA-OneVision         & 7B      & 69.78 & 60.24 & 36.9 \\
InternVL3.5             & 1B      & 68.54 & 59.16 & 36.4 \\
LLaVA-1.5               & 7B      & 47.99 & 29.56 & 71.4 \\
InternVL3               & 78B     & 39.11 & 18.01 & 81.9 \\
Mistral Small           & 24B     & 35.87 & 12.98 & 88.8 \\
\midrule
\multicolumn{5}{c}{\textbf{Proprietary VLMs}} \\
\midrule
Gemini 2.5 Flash        & --      & 93.41 & 92.37 &  4.1 \\
Gemini 2.5 Pro          & --      & 91.50 & 89.79 &  6.7 \\
Claude Sonnet 4.5       & --      & 89.74 & 87.71 &  7.9 \\
GPT-5                   & --      & 88.63 & 85.96 & 10.4 \\
GPT-o4-mini             & --      & 86.89 & 83.74 & 12.3 \\
GPT-4o                  & --      & 81.51 & 76.73 & 18.6 \\
\bottomrule
\end{tabular}
\end{table}

\begin{table}[t]
\centering
\scriptsize
\setlength{\tabcolsep}{2pt}
\caption{Results on the \textbf{Embodied AI} domain with the \textbf{Long} prompt. 
Score and Acc are reported in \%. Cannot is the percentage of questions where the model caption does not contain an answer.}
\label{tab:emb_long}
\begin{tabular}{l c c c c}
\toprule
\textbf{Model} & \textbf{Size} & \textbf{Score$\uparrow$} & \textbf{Acc$\uparrow$} & \textbf{Cannot$\downarrow$} \\
\midrule
\multicolumn{5}{c}{\textbf{Open-Source VLMs}} \\
\midrule
Qwen3-VL           & 30B-A3B & 82.55 & 78.79 & 13.2 \\
Qwen3-VL           & 8B      & 81.06 & 76.87 & 14.7 \\
Qwen3-VL           & 4B      & 80.35 & 75.67 & 16.4 \\
GLM-4.1V           & 9B      & 79.01 & 73.68 & 18.7 \\
InternVL3.5        & 38B     & 75.41 & 68.84 & 23.1 \\
Qwen2.5-VL         & 72B     & 73.77 & 66.72 & 24.7 \\
Qwen2.5-VL         & 32B     & 73.73 & 67.20 & 22.9 \\
InternVL3.5        & 30B-A3B & 70.02 & 61.28 & 30.6 \\
Qwen2.5-VL         & 7B      & 68.74 & 59.41 & 32.8 \\
InternVL3.5        & 8B      & 68.41 & 58.89 & 33.5 \\
NVLM-D             & 72B     & 67.15 & 56.78 & 36.5 \\
InternVL3          & 14B     & 66.89 & 56.71 & 35.7 \\
LLaVA-OneVision    & 7B      & 65.67 & 55.16 & 36.9 \\
InternVL3          & 8B      & 65.05 & 53.55 & 40.4 \\
InternVL3.5        & 1B      & 64.68 & 53.97 & 37.7 \\
LLaVA-1.5          & 7B      & 50.01 & 32.29 & 62.4 \\
Mistral Small      & 24B     & 39.24 & 15.05 & 85.2 \\
InternVL3          & 78B     & 33.84 &  7.34 & 93.5 \\
\midrule
\multicolumn{5}{c}{\textbf{Proprietary VLMs}} \\
\midrule
Gemini 2.5 Flash   & --      & 86.97 & 84.27 &  9.4 \\
Gemini 2.5 Pro     & --      & 86.78 & 84.14 &  9.3 \\
GPT-o4-mini        & --      & 83.33 & 80.34 & 10.5 \\
GPT-4o             & --      & 83.21 & 79.50 & 13.0 \\
GPT-5              & --      & 82.83 & 80.41 &  1.6 \\
Claude Sonnet 4.5  & --      & 69.90 & 61.12 & 30.9 \\
\bottomrule
\end{tabular}
\end{table}

\begin{table}[t]
\centering
\scriptsize
\setlength{\tabcolsep}{2pt}
\caption{Results on the \textbf{Embodied AI} domain with the \textbf{Simple} prompt. 
Score and Acc are reported in \%. Cannot is the percentage of questions where the model caption does not contain an answer.}
\label{tab:emb_simple}
\begin{tabular}{l c c c c}
\toprule
\textbf{Model} & \textbf{Size} & \textbf{Score$\uparrow$} & \textbf{Acc$\uparrow$} & \textbf{Cannot$\downarrow$} \\
\midrule
\multicolumn{5}{c}{\textbf{Open-Source VLMs}} \\
\midrule
Qwen3-VL              & 30B-A3B & 82.15 & 78.13 & 14.1 \\
Qwen3-VL              & 4B      & 80.56 & 75.93 & 16.2 \\
Qwen3-VL              & 8B      & 80.37 & 75.70 & 16.3 \\
GLM-4.1V              & 9B      & 75.56 & 68.71 & 24.1 \\
InternVL3.5           & 38B     & 74.68 & 67.65 & 24.7 \\
Qwen2.5-VL            & 32B     & 72.98 & 65.85 & 25.1 \\
InternVL3             & 8B      & 72.07 & 63.97 & 28.4 \\
Qwen2.5-VL            & 72B     & 71.60 & 63.36 & 28.9 \\
NVLM-D                & 72B     & 70.31 & 61.75 & 30.1 \\
InternVL3             & 14B     & 69.75 & 61.00 & 30.7 \\
InternVL3.5           & 30B-A3B & 69.75 & 60.75 & 31.6 \\
Qwen2.5-VL            & 7B      & 68.36 & 58.83 & 33.5 \\
InternVL3.5           & 8B      & 67.24 & 57.25 & 35.1 \\
InternVL3.5           & 1B      & 64.46 & 53.73 & 37.7 \\
LLaVA-OneVision       & 7B      & 61.01 & 48.39 & 44.4 \\
LLaVA-1.5             & 7B      & 49.84 & 31.73 & 63.8 \\
Mistral Small         & 24B     & 33.78 &  6.34 & 96.5 \\
InternVL3             & 78B     & 34.32 &  7.88 & 93.2 \\
\midrule
\multicolumn{5}{c}{\textbf{Proprietary VLMs}} \\
\midrule
GPT-5                 & --      & 86.82 & 84.21 &  9.1 \\
Gemini 2.5 Pro        & --      & 85.45 & 82.22 & 11.3 \\
Gemini 2.5 Flash      & --      & 84.89 & 81.63 & 11.4 \\
GPT-o4-mini           & --      & 82.94 & 79.21 & 13.1 \\
GPT-4o                & --      & 81.61 & 77.11 & 15.7 \\
Claude Sonnet 4.5     & --      & 67.27 & 57.08 & 35.8 \\
\bottomrule
\end{tabular}
\end{table}

\begin{table}[t]
\centering
\scriptsize
\setlength{\tabcolsep}{2pt}
\caption{Results on the \textbf{Embodied AI} domain with the \textbf{Short} prompt. 
Score and Acc are reported in \%. Cannot is the percentage of questions where the model caption does not contain an answer.}
\label{tab:emb_short}
\begin{tabular}{l c c c c}
\toprule
\textbf{Model} & \textbf{Size} & \textbf{Score$\uparrow$} & \textbf{Acc$\uparrow$} & \textbf{Cannot$\downarrow$} \\
\midrule
\multicolumn{5}{c}{\textbf{Open-Source VLMs}} \\
\midrule
GLM-4.1V              & 9B      & 61.29 & 46.72 & 51.1 \\
Qwen3-VL              & 4B      & 59.63 & 44.33 & 53.7 \\
Qwen3-VL              & 30B-A3B & 57.56 & 41.23 & 57.4 \\
InternVL3.5           & 38B     & 57.16 & 40.78 & 57.5 \\
Qwen3-VL              & 8B      & 56.24 & 39.24 & 59.7 \\
InternVL3             & 8B      & 55.17 & 37.88 & 60.8 \\
Qwen2.5-VL            & 72B     & 54.83 & 37.22 & 61.8 \\
InternVL3.5           & 30B-A3B & 54.61 & 37.33 & 60.7 \\
InternVL3             & 14B     & 53.77 & 35.68 & 63.5 \\
Qwen2.5-VL            & 32B     & 53.58 & 35.66 & 63.0 \\
Qwen2.5-VL            & 7B      & 53.27 & 35.12 & 63.8 \\
InternVL3.5           & 1B      & 52.66 & 34.73 & 63.0 \\
InternVL3.5           & 8B      & 52.12 & 33.59 & 65.1 \\
NVLM-D                & 72B     & 51.53 & 32.97 & 65.2 \\
LLaVA-OneVision       & 7B      & 49.41 & 29.63 & 69.6 \\
LLaVA-1.5             & 7B      & 46.75 & 26.02 & 73.0 \\
InternVL3             & 78B     & 34.03 &  7.36 & 94.1 \\
Mistral Small         & 24B     & 33.95 &  6.59 & 96.3 \\
\midrule
\multicolumn{5}{c}{\textbf{Proprietary VLMs}} \\
\midrule
GPT-o4-mini           & --      & 57.76 & 41.39 & 57.5 \\
GPT-5                 & --      & 57.73 & 41.36 & 57.5 \\
Gemini 2.5 Pro        & --      & 57.40 & 40.76 & 57.9 \\
Gemini 2.5 Flash      & --      & 55.78 & 38.43 & 61.0 \\
GPT-4o                & --      & 54.47 & 36.50 & 63.1 \\
Claude Sonnet 4.5     & --      & 53.16 & 34.97 & 64.0 \\
\bottomrule
\end{tabular}
\end{table}

\begin{table}[t]
\centering
\scriptsize
\setlength{\tabcolsep}{2pt}
\caption{Results on the \textbf{Embodied AI} domain with the \textbf{Taxonomy-Hinted} prompt.
Score and Acc are reported in \%. Cannot is the percentage of questions where the model caption does not contain an answer.}
\label{tab:emb_tax}
\begin{tabular}{l c c c c}
\toprule
\textbf{Model} & \textbf{Size} & \textbf{Score$\uparrow$} & \textbf{Acc$\uparrow$} & \textbf{Cannot$\downarrow$} \\
\midrule
\multicolumn{5}{c}{\textbf{Open-Source VLMs}} \\
\midrule
Qwen3-VL                & 8B      & 74.85 & 67.32 & 26.3 \\
GLM-4.1V                & 9B      & 72.59 & 64.33 & 28.9 \\
Qwen2.5-VL              & 32B     & 70.02 & 61.21 & 30.8 \\
Qwen3-VL                & 30B-A3B & 70.70 & 61.67 & 31.7 \\
Qwen2.5-VL              & 72B     & 69.66 & 60.21 & 33.2 \\
Qwen3-VL                & 4B      & 69.55 & 59.76 & 34.2 \\
InternVL3               & 8B      & 67.46 & 57.27 & 35.7 \\
InternVL3.5             & 38B     & 66.35 & 55.41 & 38.4 \\
InternVL3.5             & 30B-A3B & 65.15 & 53.84 & 39.5 \\
NVLM-D                  & 72B     & 64.42 & 53.18 & 39.4 \\
Qwen2.5-VL              & 7B      & 63.44 & 51.65 & 41.3 \\
InternVL3               & 14B     & 62.86 & 50.87 & 42.0 \\
LLaVA-OneVision         & 7B      & 62.03 & 49.32 & 44.7 \\
InternVL3.5             & 8B      & 61.62 & 48.96 & 44.4 \\
InternVL3.5             & 1B      & 60.61 & 47.88 & 44.7 \\
LLaVA-1.5               & 7B      & 46.92 & 27.44 & 68.6 \\
InternVL3               & 78B     & 34.40 &  8.13 & 92.7 \\
Mistral Small           & 24B     & 34.17 &  7.22 & 95.0 \\
\midrule
\multicolumn{5}{c}{\textbf{Proprietary VLMs}} \\
\midrule
Gemini 2.5 Pro          & --      & 85.35 & 81.86 & 12.1 \\
GPT-5                   & --      & 85.10 & 81.37 & 13.1 \\
Gemini 2.5 Flash        & --      & 82.81 & 78.42 & 15.4 \\
GPT-o4-mini             & --      & 80.08 & 74.69 & 18.9 \\
GPT-4o                  & --      & 76.27 & 69.30 & 24.4 \\
Claude Sonnet 4.5       & --      & 70.44 & 62.23 & 28.8 \\
\bottomrule
\end{tabular}
\end{table}


\end{document}